%% file: acl_latex.tex
\pdfoutput=1
\documentclass[11pt]{article}

\usepackage[final]{acl}

\usepackage{times}
\usepackage{latexsym}

\usepackage[T1]{fontenc}
\usepackage[utf8]{inputenc}

\usepackage{microtype}

\usepackage{inconsolata}

\usepackage{booktabs}
\usepackage{graphicx}    % For including graphics
\usepackage{caption}     % For customizing captions
\usepackage{subcaption}  % For creating subfigures
\usepackage{adjustbox} 
\usepackage{multirow}
\usepackage{placeins}

\usepackage{changepage}

\usepackage{pgf}       % needed for \pgfmathsetmacro

\usepackage{tcolorbox}
\tcbuselibrary{skins} 

\usepackage{longtable} % For full-page tables
\usepackage{array}     % For flexible column definitions

\usepackage{pgfmath}
\usepackage{makecell}
\usepackage{fontawesome5}
\usepackage{hyperref}
\usepackage{epigraph}
\definecolor{teal100}{RGB}{0, 90, 90}  
\definecolor{teal90}{RGB}{0, 110, 110} 
\definecolor{teal80}{RGB}{0, 150, 150}  
\definecolor{teal70}{RGB}{0, 220, 220}

\definecolor{burntorange}{RGB}{204, 85, 0}

\newcommand{\fancyepigraph}[2]{%
\vspace{-5pt}
  \begin{center}
    \begin{tikzpicture}
      \node [opacity=0.1, text=black, font=\fontsize{11em}{11em}\selectfont] at (0,0) {``};
      \node [align=center, text width=0.97\linewidth] {
        {\small  \itshape #1}\\[0em]
        {\footnotesize --- #2}
      };
    \end{tikzpicture}
  \end{center}
}

\newcommand{\colorTPR}[1]{%
  \pgfmathsetmacro{\val}{int(#1 + 0.5)}
  \ifnum\val=100
    \textcolor{teal100}{\textbf{#1}}%
  \else
    \ifnum\val>89
      \textcolor{teal80}{\textbf{#1}}%
    \else
      \ifnum\val>79
        \textcolor{burntorange}{\textbf{#1}}%
      \else
        \ifnum\val>69
          \textcolor{burntorange}{\textbf{#1}}%
        \else
          \textcolor{purple}{\textbf{#1}}%
        \fi
      \fi
    \fi
  \fi
}

\newcommand{\colorFPR}[1]{%
  \pgfmathsetmacro{\val}{int(#1 + 0.5)}% 
  \ifnum\val=0
    \textcolor{teal100}{\textbf{#1}}% 
  \else
    \ifnum\val<6
      \textcolor{teal80}{\textbf{#1}}% %
    \else
      \ifnum\val<11
        \textcolor{burntorange}{\textbf{#1}}% 
      \else
        \textcolor{purple}{\textbf{#1}}%
      \fi
    \fi
  \fi
}

\newcommand{\tprfpr}[2]{%
  \makebox[3em][r]{\colorTPR{#1}} \,\footnotesize(\colorFPR{#2})%
}

\definecolor{lightred}{HTML}{e99090}

\newcommand{\name}{\textsc{\small Human Detectors}}

\newcommand{\claude}{\textsc{\small Claude-3.5-Sonnet}}
\newcommand{\gptpro}{\textsc{\small o1-Pro}}
\newcommand{\oone}{\textsc{\small o1}}
\newcommand{\gpt}{\textsc{\small GPT-4o}}
\newcommand{\gptnov}{\textsc{\small GPT-4o-2024-11-20}}
\newcommand{\gptaug}{\textsc{\small GPT-4o-2024-08-06}}

\title{People who frequently use ChatGPT for writing tasks\\ are accurate and robust detectors of AI-generated text}

\author{
  Jenna Russell\textsuperscript{1}\quad
  Marzena Karpinska\textsuperscript{2}\quad
  Mohit Iyyer\textsuperscript{1,3}\\
  \textsuperscript{1}University of Maryland, College Park \quad
  \textsuperscript{2}Microsoft \quad \textsuperscript{3}UMass Amherst\\
  \texttt{\{jennarus,miyyer\}@umd.edu}, \texttt{mkarpinska@microsoft.com}
}

\begin{document}
\maketitle
\input{sections/0-abstract}
\input{sections/1-intro}

\input{sections/2-data_and_methodology}
\input{sections/3-human_detectors}

\input{sections/33-explanations}

\input{sections/4-explainable-text-detection}
\input{sections/5-related-work}
\input{sections/6-conclusion}

\input{sections/7-limitations}

\input{sections/8-ethical_considerations}

\input{sections/9-ack}

\bibliography{custom}

\appendix
\input{sections/appendix}

\end{document}

%% file: sections/0-abstract.tex
\begin{abstract}

In this paper, we study how well \emph{humans} can detect text generated by commercial LLMs (\gpt, \claude, \gptpro). We hire annotators to read 300 non-fiction English articles, label them as either human-written or AI-generated, and provide paragraph-length explanations for their decisions. Our experiments show that annotators who frequently use LLMs for writing tasks excel at detecting AI-generated text, even without any specialized training or feedback. In fact, the majority vote among five such ``expert'' annotators misclassifies only 1 of 300 articles, significantly outperforming most commercial and open-source detectors we evaluated even in the presence of evasion tactics like paraphrasing and humanization.
Qualitative analysis of the experts’ free-form explanations shows that while they rely heavily on specific lexical clues,
they also pick up on more complex phenomena within the text 
that are challenging to assess for automatic detectors. We release our annotated dataset and code to spur future research into both human and automated detection of AI-generated text.

    \faGithub[regular]  {\scriptsize \url{https://github.com/jenna-russell/human_detectors}}

\end{abstract}

%% file: sections/1-intro.tex
\section{Introduction}
\label{sec:intro}

AI-generated text is rampant in today's world,
spurring research into automatic detectors to help combat academic plagiarism \cite{zhu_embracing_2024} and fake content creation \cite{gameiro_llm_2024}. Unfortunately, automatic detectors suffer from low detection rates, poor robustness to evasion attempts, and limited explainability to end users \cite{sadasivan_can_2024, ji_detecting_2024}.\footnote{This paper focuses only on \emph{post-hoc} detection of AI-generated text, which, unlike watermarking~\cite{pmlr-v202-kirchenbauer23a} requires no cooperation from LLM providers.}

\begin{figure}[tbp]
\centering
\includegraphics[width=\linewidth]{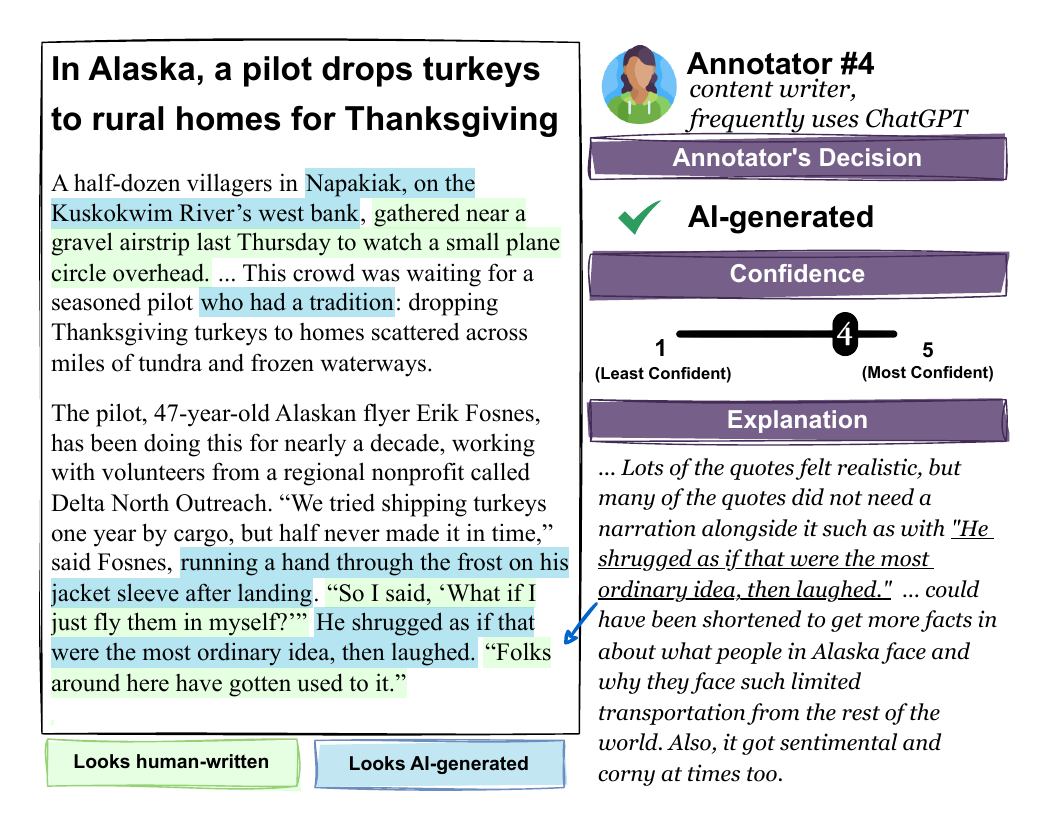}
\caption{A human expert's annotations of an article generated by OpenAI's \gptpro\ with humanization. The expert provides a judgment on whether the text is written by a human or AI, a confidence score, and an explanation (including both free-form text and highlighted spans) of their decision. 
}
\vspace{-3mm}
\label{fig:main}
\end{figure}

In this paper, we instead study how well \emph{humans} can detect AI-generated text. Unlike prior work on this topic, which was mainly conducted in the pre-ChatGPT era \cite{ippolito_automatic_2020, brown_language_2020, clark_all_2021}, we focus on text generated by modern LLMs (\gpt, \claude, \gptpro) and in the presence of evasion attempts 
(paraphrasing, humanization). We hire human annotators to read non-fiction English articles, label them as written by either a human or by AI, and provide a paragraph-length explanation of their decision-making process (\autoref{fig:main}). Overall, we collect \textbf{1790} annotations on \textbf{300} unique articles at a total cost of \textbf{\$4.9K USD}, which allows us to compare humans to automatic detectors and analyze what kind of clues they rely on. 

\paragraph{Experience matters:}
Unsurprisingly, annotators who rarely or never use LLMs are  poor detectors of AI-generated text; in fact, they overestimate their own ability to perform the task by providing high confidence scores for their decisions. We identify a subset of five high-performing annotators who frequently use LLMs for writing tasks (e.g., editing, copywriting, creative writing). The majority vote of this subset of ``expert'' annotators fails to predict the correct label on only \textbf{one} out of 300 articles. We emphasize that we do not provide any training to our annotators: they are given no feedback on which articles they labeled incorrectly.

\paragraph{Human experts outperform automatic detectors:} The majority vote of our five expert annotators
outperforms almost every commercial and open-source detector we tested on these 300 articles, with only the commercial Pangram model~\cite{emi_technical_2024} matching their near-perfect detection accuracy. In our most complex configuration, which requires detecting  articles generated and humanized by \gptpro, the expert majority vote is perfect (true positive rate of 100\%), while open-source methods such as Binoculars (6.7\%) and Fast-DetectGPT (23.3\%) struggle. We conclude that hiring expert human annotators to perform detection is a viable strategy, particularly in high-stakes settings where explainability is critical. 

\paragraph{What do expert annotators focus on?}
An analysis of our expert's explanations reveals that usage of ``AI vocabulary'' (e.g., \emph{vibrant}, \emph{crucial}, \emph{significantly}) form the most common giveaways. Close behind are formulaic sentence and document structures (e.g., optimistically vague 
conclusions) and originality (how creative or engaging an article is).
We observe that neither paraphrasing nor humanization effectively removes all of these signatures; that said, these evasion tactics and defenses for them are still underexplored in the research community. 

\paragraph{Can LLMs mimic human detectors?}
We attempt to automatically replicate the decision-making process of our human experts by providing a candidate text to an LLM, along with a guidebook compiled from human expert explanations, and instructing the model to use the guidebook to judge whether the text is written by an AI or by a human. This simple prompting strategy is competitive with existing detectors on easier configurations but struggles to detect humanized articles.
Moving forward, we believe LLMs can be used in conjunction with high-performing detectors such as Pangram to offer an explanation of the judgment to end users.

%% file: sections/2-data_and_methodology.tex
\section{How good are humans at detecting AI-generated text?}
\label{sec:data_methodology}
 
How well do \emph{human} detectors stack up against their automatic counterparts? What features do humans find most predictive of AI writing? To answer these questions, we conducted five experiments that ask humans to judge increasingly more deceptive AI-generated texts.  

\paragraph{Task setup:}
Each experiment consists of \textbf{60} articles, 30 human-written and 30 LLM-generated. Each human-written article has a corresponding AI-generated counterpart with the same title and subtitle to control for high-level content and topics.
Articles are shown to annotators 
in a randomized order. 
For each, they provide (1) a \textbf{binary label} (human-written or AI-generated), (2) a \textbf{confidence rating} on a scale from 1 (least confident) to 5 (most confident), (3) \textbf{highlighted spans} used as clues, and (4) a paragraph-length \textbf{explanation}. We evaluate with \emph{True Positive Rate (TPR)}, which measures the percentage of AI-generated articles that are successfully detected, as well as \emph{False Positive Rate (FPR)}, which measures the percentage of human-written articles marked as AI-generated.\footnote{Prior work in automatic detection reports either TPR at a fixed low FPR such as 1\% \cite{krishna_paraphrasing_2023, hans_spotting_2024, dugan_raid_2024} or AUROC \cite{mitchell_detectgpt_2023, hu_radar_2023, bao_fast-detectgpt_2024}, which we cannot easily do with human annotators.}

\paragraph{Article selection:}
 We restrict our study to English nonfiction articles of fewer than 1K words.\footnote{Human-written articles in our experiments come from eight different American publications: Associated Press, Discover Magazine, National Geographic, New York Times, Reader's Digest, Scientific American, Smithsonian Magazine, and Wall Street Journal.} While we cannot make claims about human detection performance in other domains (e.g., scientific papers or social media posts),\footnote{We report results on \S\ref{subsec:experiment1} for fictional stories in \S\ref{app_subsec:story_generation}.} we focus on such articles because (1) malicious LLM users generate large volumes of fake articles, making it practically impactful; (2) such articles do not require specialized knowledge to understand or judge; and (3) many previous studies also ask humans to judge AI-generated text in this domain~\cite{ippolito_automatic_2020, clark_all_2021, puccetti_ai_2024}. 

\paragraph{Generating paired AI articles:}
For each human-written article, we generate a corresponding AI article by prompting an LLM with its title, subtitle, desired length, and publication source. This ensures that AI-generated articles are directly comparable in topic and length,\footnote{We include a word count target to ensure comparable lengths between article pairs, reported in \autoref{tab:article_lengths}.} 
% eliminating confounds from content variation. 
By constructing these human-AI article pairs with only authorship as the varying factor, we create minimal pairs that allow direct comparisons \cite{gardner-etal-2020-evaluating, warstadt-etal-2020-blimp-benchmark, karpinska-etal-2022-demetr, karpinska-etal-2024-one}. 
For experiments 3 and 5, we modify the prompt to include perturbations designed to evade detectors 
(paraphrasing and humanization). 
We employ a within-subjects design for our experiments, where each annotator judges both the human-written and AI-generated articles. This reduces variability from individual differences and requires fewer annotators \cite{Allen2017-withinsubject}. 
We further minimize bias by randomizing article order and ensuring annotators are unaware of the pairing.\footnote{We set \texttt{temperature=0} for Experiments 1-3 and use ChatGPT (Pro) interface for Experiments 4-5 as the API for \oone\ was not yet available.} 

% An example article generation prompt is below:
% \begin{quote}
% \footnotesize
% \texttt{You are given the following title and subtitle of a general article from the Science section of New York Times and asked to write a corresponding article of around 750 words. Include quotations from relevant experts and make sure the article is concise and easily understandable to a lay audience. \\\\ 
% Title: The Science That Makes Baseball Mud ‘Magical’ \\\\
% Subtitle: Scientists dug up the real dirt on the substance applied to all the baseballs used in the major leagues. 
% \\ \\
% Article: }
% \end{quote}

\paragraph{Annotator details:} 
We recruited annotators via the Upwork freelancing platform, all of whom self-identified as native English speakers.\footnote{See \autoref{tab:survey_data} for survey details.} 
All annotators were required to read the project guidelines and sign a consent form prior to the task, provided in \S\ref{app_sec:human_evaluation}. 
Additionally, we surveyed each annotator to determine their level of education, profession, English dialect, and familiarity with LLMs, including how they use these models (see \autoref{tab:survey_data} for details). 
They were compensated \$2 per candidate text, resulting in an hourly rate of about \$15 - \$30 USD. Ultimately, we collect 1740 annotations from 9 annotators at a total cost of \$4.9K USD. 

\paragraph{Justification of sample size:}
Most experiments in this paper are conducted by five expert annotators (Experiment 1 includes four nonexpert annotators).\footnote{See \S\ref{app_subsec:finding_expert_annotators} for details on expert recruitment.} 
While a larger sample size would strengthen
population-level claims; prohibitive costs\footnote{It cost us \$865 USD per annotator for five experiments.}
limit us to the current sample size. 
We employ a within-subjects design for our experiments, where each annotator judges both the human-written and AI-generated articles, reducing individual variability and requiring fewer annotators \cite{Allen2017-withinsubject}. 
Our set of five expert annotators provide an insightful view of the human \emph{upper bound} at AI-generated text detection without additional training. Furthermore, collecting detailed explanations of these annotators' decision-making process, while expensive, enables fine-grained qualitative analysis and ensures that annotators actually read the texts.\footnote{Collecting explanations mitigates issues raised by ~\citet{karpinska-etal-2021-perils}, whose Mechanical Turk annotators ``read'' long-form texts in a matter of mere seconds before evaluation.}

%% file: sections/3-human_detectors.tex
\subsection{\colorbox{SkyBlue!70}{\faFlask\ Experiment 1:} What kinds of annotators can reliably detect AI-generated articles?}
\label{subsec:experiment1}

\begin{footnotesize}
      \hspace{2.6em} \fcolorbox{SkyBlue!70}{SkyBlue!30}{\faCogs \hspace{0.1cm} \textbf{LLM:} \gpt\ (\texttt{\scriptsize{2024-08-06}})} \\
      \hspace*{2.6em} \fcolorbox{SkyBlue!70}{SkyBlue!30}{\faGhost \hspace{0.25cm} \textbf{Evasion tactic:} none}\vspace{0.2cm}
    \end{footnotesize}

Is there a population of humans that is highly accurate at our detection task, and if so, are there any commonalities between them? 

We first ask five annotators with varying backgrounds and levels of expertise with LLMs to attempt a batch of 60 articles. 
% (half generated by \gpt\ {\small\texttt{2024-08-06}} and half human-written as described above). 
We observe both ends of the spectrum in terms of performance: annotators who rarely or never use LLMs for writing tasks performed near random, while one annotator who uses LLMs daily to edit LLM-generated text performed almost perfectly. More details on our initial study can be found in \S\ref{app_sec:pilot_study}. Inspired by this result, we recruit four more annotators with similar backgrounds as our top performer, who all of whom perform similarly well on the same batch of articles.  

\begin{table}[t!]
    \centering
    \footnotesize
    \resizebox{\columnwidth}{!}{%
    \begin{tabular}{lcc}
        \toprule
        \faChartLine\ \textsc{Metric} & \faUser\ \textsc{Nonexperts} & \faUserTie\ \textsc{Experts} \\
        \midrule
        Avg. TPR  & 56.7 & 92.7 \\
        Avg. FPR & 51.7 & 4.0 \\
        Avg. Confidence & 4.03 & 4.39 \\
        \bottomrule
    \end{tabular}
    }
    \caption{On average, nonexperts perform similar to random chance at detecting AI-generated text, while experts are highly accurate.}
    \label{tab:expert_vs_non_expert}
\end{table}

\paragraph{Annotators with limited LLM experience are unreliable detectors:}
The four annotators who self-report either not using LLMs at all, or using LLMs only for non-writing tasks, detect AI-generated text at a similar rate to random chance, achieving an average TPR of 56.7\% and FPR of 52.5\%  (\autoref{tab:expert_vs_non_expert}).\footnote{We report  average TPR and FPR instead of majority vote since there were only four \textit{nonexpert} annotators. For the remainder of the paper, we report the majority vote of the five \textit{expert} annotators (i.e., at least three out of five agreed).} 
Despite their poor performance, these annotators reported high confidence (avg. 4.03) in their decisions, suggesting they may be overestimating their detection abilities. We term this population \emph{nonexperts} for the rest of this paper. 

\paragraph{Annotators who frequently use LLMs for writing tasks are highly accurate:}
In contrast, the five annotators who have significant experience with using LLMs for writing-related tasks are able to detect AI-generated text very reliably, achieving a TPR of 92.7\%.\footnote{We compare expert vs. nonexpert clues in \S\ref{app_subsec:nonexpert_comments}.} The average FPR for this population of annotators was 3.3\%, meaning that they rarely mistake human-written text as AI-generated.  The majority vote out of these five annotators correctly determined authorship of all 60 articles; the ``\gpt'' column of \autoref{tab:main_results} contains more details. We term this population \emph{experts} for the rest of this paper; all subsequent experiments rely on only these five expert annotators. 

\paragraph{What do expert annotators see that nonexperts don't?}
To understand why experts far outperformed nonexperts at detecting AI-generated text, we analyze the comments each annotator provided in their explanations. Overall, nonexperts often mistakenly fixate on certain linguistic properties  compared to experts. One example is vocabulary choice, where nonexperts take the inclusion of any ``fancy'' or otherwise low-frequency word types as signs of AI-generated text; in contrast, experts are much more familiar with exact words and phrases overused by AI (e.g., \emph{testament}, \emph{crucial}).\footnote{A complete list of ``AI vocab'' found in the detection guide (\autoref{tab:detection-guide}) is detailed in \autoref{tab:ai_vocab}.} Nonexperts also believe that human authors are more likely than AI to form grammatically-correct sentences, while experts realize the opposite is true: humans actually make more grammatical errors. Finally, nonexperts attribute any text written in a neutral tone to AI, 
resulting in many false positives.

\begin{table*}[ht!]
\small
\centering
\begin{adjustbox}{max width=\textwidth}
\begin{tabular}{lllllll}
\toprule
\multirow{2}{*}{\textsc{Detection Method}} 
& \multicolumn{5}{c}{\textsc{Generation Method}} \\ 
\cmidrule(lr){2-6} 
& \makecell{\textsc{GPT-4o}\\ \footnotesize TPR\% (\scriptsize FPR\%)}
& \makecell{\textsc{Claude}\\ \footnotesize TPR\% (\scriptsize FPR\%)}
& \makecell{\textsc{GPT-4o {\small\textsc{paraphrased}}}\\ \footnotesize TPR\% (\scriptsize FPR\%)}
& \makecell{\textsc{o1-Pro}\\ \footnotesize TPR\% (\scriptsize FPR\%)}
& \makecell{\textsc{o1-Pro {\small\textsc{humanized}}}\\ \footnotesize TPR\% (\scriptsize FPR\%)}
& \makecell{\textsc{OVERALL}\\ \footnotesize TPR\% (\scriptsize FPR\%)} \\
\midrule
\multicolumn{6}{l}{\textbf{(A) Expert human detectors}}\vspace{0.1cm}\\
\textsc{\faUsers\ Expert Majority Vote}
 & \tprfpr{100}{0}
 & \tprfpr{100}{0}
  & \tprfpr{100}{0}
 & \tprfpr{96.7}{0}
 & \tprfpr{100}{0}
  & \tprfpr{99.3}{0}\vspace{0.1cm}
 \\

\hspace{0.5cm}\textsc{\faUser\ Annotator 1}
 & \tprfpr{96.7}{3.3} & \tprfpr{100}{0}
 & \tprfpr{100}{0}
 & \tprfpr{96.7}{6.7}
 & \tprfpr{90.0}{23.3} 
 & \tprfpr{96.7}{6.7} 
 \\
\hspace{0.5cm}\textsc{\faUser\ Annotator 2}
 & \tprfpr{96.7}{0}
  & \tprfpr{80.0}{30}
 & \tprfpr{86.7}{10}
 & \tprfpr{90.0}{10}
 & \tprfpr{86.7}{10} 
 & \tprfpr{88.0}{12} 
 \\
\hspace{0.5cm}\textsc{\faUser\ Annotator 3}
 & \tprfpr{86.7}{6.7}
 & \tprfpr{100}{0}
  & \tprfpr{93.3}{0}
 & \tprfpr{16.7}{0}
 & \tprfpr{0}{3.3} 
  & \tprfpr{59.3}{2} 
 \\
\hspace{0.5cm}\textsc{\faUser\ Annotator 4}
 & \tprfpr{90.0}{6.7}
 & \tprfpr{96.7}{13.3}
  & \tprfpr{100}{10}
 & \tprfpr{100}{0}
 & \tprfpr{100}{0} 
  & \tprfpr{97.3}{6} 
  \\
\hspace{0.5cm}\textsc{\faUser\ Annotator 5}
 & \tprfpr{93.3}{0}
 & \tprfpr{93.3}{6.7}
  & \tprfpr{93.3}{0}
 & \tprfpr{93.3}{0}
 & \tprfpr{93.3}{0} 
  & \tprfpr{93.3}{1.3} 
 \\
\midrule
\multicolumn{6}{l}{\textbf{(B) Automatic detectors}}\vspace{0.1cm}\\
\textsc{\faLock\ \href{https://www.pangram.com/blog/humanizers-announcement}{Pangram Humanizers}}
 & \tprfpr{100}{0}
 & \tprfpr{100}{3.3}
  & \tprfpr{100}{0}
 & \tprfpr{100}{0}
 & \tprfpr{96.7}{10} 
 & \tprfpr{99.3}{2.7} 
 
 \\

\textsc{\faLock\ \href{https://www.pangram.com/}{Pangram}}
 & \tprfpr{100}{0}
 & \tprfpr{100}{3.3}
  & \tprfpr{100}{0}
 & \tprfpr{100}{0}
 & \tprfpr{90.0}{6.7} 
 & \tprfpr{98.0}{2} 
 
 \\
\textsc{\faLock\ \href{https://gptzero.me/}{GPTZero}}
 & \tprfpr{100}{0}
 & \tprfpr{96.7}{0}
  & \tprfpr{100}{0}
 & \tprfpr{76.7}{0}
 & \tprfpr{46.7}{3.3} 
  & \tprfpr{85.3}{0.7} 
 \\

\textsc{\faUnlock\ \href{https://github.com/baoguangsheng/fast-detect-gpt}{Fast-DetectGPT} (FPR=0.05)}
 & \tprfpr{100}{0}
 & \tprfpr{96.7}{3.3}
  & \tprfpr{56.7}{3.3}
 & \tprfpr{86.7}{0}
 & \tprfpr{23.3}{3.3}
  & \tprfpr{80.0}{7.2} 
 \\
\textsc{\faUnlock\ \href{https://github.com/ahans30/Binoculars}{Binoculars} (Accuracy)}
 & \tprfpr{100}{0}
 & \tprfpr{93.3}{0}
  & \tprfpr{60.0}{6.7}
 & \tprfpr{73.3}{0}
 & \tprfpr{6.67}{0} 
 & \tprfpr{66.7}{1.3} 
 \\
\textsc{\faUnlock\ \href{https://github.com/ahans30/Binoculars}{Binoculars} (Low FPR)}
 & \tprfpr{96.7}{0}
 & \tprfpr{80.0}{0}
  & \tprfpr{13.3}{0}
 & \tprfpr{10.0}{0}
 & \tprfpr{0}{0} 
  & \tprfpr{40.0}{0} 
 \\
 \textsc{\faUnlock\ \href{https://github.com/IBM/RADAR}{RADAR} (FPR=0.05)}
 & \tprfpr{66.7}{0}
 & \tprfpr{0}{0}
 & \tprfpr{10}{3.3}
 & \tprfpr{0}{3.3}
 & \tprfpr{0}{3.3}
 & \tprfpr{15.3}{2}
 \\
 \midrule
\multicolumn{6}{l}{\textbf{(C) Prompt-based detectors}}\vspace{0.2cm}\\
\multicolumn{6}{l}{\hspace{0.5cm}Detector LLM: \textbf{\gptnov}}\vspace{0.1cm}\\

\textsc{\faCogs\ Zero-shot }
 & \tprfpr{100}{10} 
 & \tprfpr{93.3}{10}
  & \tprfpr{100}{6.7}
 & \tprfpr{56.7}{3.3}
 & \tprfpr{6.7}{3.3} 
  & \tprfpr{71.3}{6.7} 
 \\

 \textsc{\faCogs\ Zero-shot + CoT}
 & \tprfpr{63.3}{3.3} 
 & \tprfpr{33.3}{0}
  & \tprfpr{96.7}{6.7}
 & \tprfpr{16.7}{0}
 & \tprfpr{0}{0} 
 & \tprfpr{42.0}{2.0}
 \\

 \textsc{\faCogs\ Zero-shot + Guide}
 & \tprfpr{100}{10}
 & \tprfpr{96.7}{10}
  & \tprfpr{100}{13.3}
 & \tprfpr{80}{6.7}
 & \tprfpr{3.3}{3.3} 
  & \tprfpr{76.0}{8.7} 
 \\
 
\textsc{\faCogs\ Zero-shot + CoT + Guide}
 & \tprfpr{100}{10}
 & \tprfpr{100}{13.3}
  & \tprfpr{100}{16.7}
 & \tprfpr{86.7}{6.7}
 & \tprfpr{3.3}{3.3} 
  & \tprfpr{78.0}{10.7} \vspace{0.2cm}
 \\

\multicolumn{6}{l}{\hspace{0.5cm}Detector LLM: \textbf{\oone\footnotesize-2024-12-17}}\vspace{0.1cm}\\

  \textsc{\faCogs\ Zero-shot}
 & \tprfpr{93.3}{3.3}
 & \tprfpr{66.6}{6.7}
 & \tprfpr{96.7}{6.7}
 & \tprfpr{40.0}{3.3}
 & \tprfpr{20.0}{6.7}
 & \tprfpr{42.2}{5.6}
 \\

   \textsc{\faCogs\ Zero-shot + Cot}
 & \tprfpr{83.3}{6.7}
 & \tprfpr{53.3}{3.3}
 & \tprfpr{96.7}{3.3}
 & \tprfpr{20}{3.3}
 & \tprfpr{16.7}{3.3}
 & \tprfpr{54}{4}
 \\

   \textsc{\faCogs\ Zero-shot + Guide}
 & \tprfpr{93.3}{0}
 & \tprfpr{30.0}{0}
 & \tprfpr{96.7}{0}
 & \tprfpr{13.3}{0}
 & \tprfpr{0}{0}
 & \tprfpr{36.7}{0}
 \\

 \textsc{\faCogs\ Zero-shot + CoT + Guide}
 & \tprfpr{86.7}{0}
 & \tprfpr{43.3}{0}
  & \tprfpr{90.0}{0}
 & \tprfpr{6.7}{0}
 & \tprfpr{0}{0} 
  & \tprfpr{53.3}{0.6} 
 \\

\bottomrule
\end{tabular}
\end{adjustbox}

\caption{%
Performance of expert humans (top), existing automatic detectors (middle), and our prompt-based detectors (bottom), where each cell displays \textbf{TPR {\small(FPR)}}.
Colors indicate performance bins where \textbf{\textcolor{teal100}{darkest teal}} is best, \textbf{\textcolor{burntorange}{orange}} is middling, and  \textbf{\textcolor{purple}{purple}} is worst. We further mark closed-source (\faLock) and open-weights (\faUnlock) detectors. The majority vote of expert humans ties Pangram  Humanizers for highest overall TPR (99.3) without any false positives, while substantially outperforming all other detectors. While the majority vote is extremely reliable, individual annotator performance varies, especially on \gptpro\ articles with and without humanization. Prompt-based detectors are unable to match the performance of either expert humans or closed-source detectors.
}

\label{tab:main_results}
\end{table*}

\subsection{{\colorbox{SkyBlue!70}{\faFlask\ Experiment 2:}} Can experts detect articles generated by a different LLM?}
\label{subsec:experiment2}

\begin{footnotesize}
  \hspace{2em}%
  \fcolorbox{SkyBlue!70}{SkyBlue!30}{%
    \faCogs\ \textbf{LLM:} \claude
  } \\
  \hspace*{2em}%
  \fcolorbox{SkyBlue!70}{SkyBlue!30}{%
    \faGhost\ \textbf{Evasion tactic:} none
  }\vspace{0.2cm}
\end{footnotesize}

While ChatGPT is the most widely-used LLM (both in general and by our expert annotators), competitors such as Anthropic's Claude \cite {anthropicclaude} also have a wide user base. Are our experts overfitting to artifacts specific to \gpt, or do they detect patterns that generalize to other LLMs? We address this question in Experiment 2 by evaluating our experts on a second batch of 60 articles, this time 
with 30 new human-written articles and 30 corresponding articles 
generated via \claude\ \cite{anthropic_claude_3_addendum} using the same instructions as in Experiment 1. 

\paragraph{Experts reliably detect articles generated by Claude:}
Despite the change in model, experts are reliable at detecting AI-generated content. TPR is almost unchanged from Experiment 1  and the expert majority vote is again 100\% accurate (``\claude'' column of \autoref{tab:main_results}), with two annotators achieving a perfect score individually; however, we find one annotator performs worse on Claude-generated texts than those of \gpt.\footnote{This annotator's performance is detailed in \S\ref{app_subsec:individual_commentary}.} 
We note that one expert (Annotator 2) achieved higher performance in the experiments where an OpenAI model was used, with their TPR falling to 80\% in this experiment. 
Annotator 2 did not have any prior experience using Claude, although three other annotators who were also unfamiliar with Claude achieved TPRs of at least 96.7\%. However, Annotator 2 focuses more heavily on the existence of ``AI vocab'' in the article than the other annotators, and they were not familiar with the distribution of words that are overused by Claude.

\subsection{{\colorbox{SkyBlue!70}{\faFlask\ Experiment 3:}} How robust are experts to paraphrasing attacks?}
\label{subsec:experiment3}
\begin{footnotesize}
      \hspace{2.6em} \fcolorbox{SkyBlue!70}{SkyBlue!30}{\faCogs \hspace{0.1cm} \textbf{LLM:} \gpt\ (\texttt{\scriptsize{2024-08-06}})} \\
      \hspace*{2.6em} \fcolorbox{SkyBlue!70}{SkyBlue!30}{\faGhost \hspace{0.25cm} \textbf{Evasion tactic:} paraphrasing}\vspace{0.2cm}
\end{footnotesize}

Users of LLMs who hope to evade detection often resort to simple post-hoc modifications of LLM outputs such as \emph{paraphrasing}. These tactics significantly drop detection rates of automatic methods~\cite{krishna_paraphrasing_2023, sadasivan_can_2024}, but how do they affect our human experts? We evaluate the five expert annotators on a third set of 60 articles, where the 30 AI-generated articles go through an additional round of paraphrasing,\footnote{Details on paraphrasing setup are in \S\ref{app_subsec:ai_text_generation}.} 
% and we show that 
showing
they maintain high TPR and low FPR.

\paragraph{Experts are robust to paraphrased AI-generated articles:} 
Overall, TPR and FPR remain almost unchanged in this experiment compared to the first two, suggesting that paraphrasing is not an effective attack on expert human detectors (see ``\textsc{gpt-4o}  {\small\textsc{paraphrased}}'' column in~\autoref{tab:main_results}). The majority vote among the five experts again correctly detects the labels of all 60 articles. 

\subsection{{\colorbox{SkyBlue!70}{\faFlask\ Experiment 4:}} Can experts keep up with advances in LLM reasoning capabilities?}
\label{subsec:experiment4}

\begin{footnotesize}
      \hspace{2.6em} \fcolorbox{SkyBlue!70}{SkyBlue!30}{\faIcon{cogs} \hspace{0.1cm} \textbf{LLM:} \gptpro} \\
      \hspace*{2.6em} \fcolorbox{SkyBlue!70}{SkyBlue!30}{\faIcon{ghost}  \hspace{0.25cm} \textbf{Evasion tactic:} none}\vspace{0.2cm}
\end{footnotesize}

While we were in the middle of conducting experiments, OpenAI released their \oone\ model~\cite{openai_openai_2024}, ushering in a new paradigm of test-time scaling. This release offered us a unique opportunity: since none of our experts had been previously exposed to a model with such advanced reasoning capabilities \cite{zhong_evaluation_2024}, how well would they do at detecting its output? 
Interestingly, our experts remain reliable detectors of articles generated by \gptpro\ (using the same prompt as in Experiments 1 and 2),\footnote{\href{https://openai.com/index/introducing-chatgpt-pro/}{\gptpro} is available through the ChatGPT Pro subscription. OpenAI states \gptpro\ produces more reliably accurate and comprehensive responses than \oone.}
although their comments show that they often perform a more detailed analysis of articles to make their judgments.

\paragraph{Four out of five experts are robust to \gptpro:} For four of five experts, detection ability of \oone-generated content remains largely consistent with prior experiments, with a majority vote TPR of 96.7\% and FPR of 0\% (see ``\gptpro'' column of \autoref{tab:main_results}).\footnote{One AI-generated article was misclassified by the expert majority vote. Analyzing their explanations, the three annotators most focused on vocabulary and sentence structure did not find enough clues in those categories to judge the article as AI-generated. The full text of the article is in \autoref{tab:incorrect_expert_article}.} 
Average confidence dropped to 4.21 out of 5, compared to average confidence of 4.39, 4.38, and 4.48 from Experiments 1,2, \& 3 respectively (see \autoref{fig:confidence_score_distribution} for details). 
% The drop in confidence, 
Verbal feedback from experts, along with the drop in confidence,
demonstrates the increased challenge posed by \oone-generated articles.
However, their steady aggregate performance highlights that even new model paradigms are still detectable by human experts. 
 
\paragraph{Experts provide more nuanced explanations when detecting \gptpro:}
In prior experiments, expert explanations focused primarily on whether or not a candidate text possesses characteristics of AI. However, experts shift focus with \gptpro\ by 
more 
frequently commenting on identifying characteristics that make text sound ``human''. 
For instance, experts frequently point to how humans repeatedly write the word \emph{says}, while AI tries to use more descriptive synonyms like \emph{notes} and \emph{explains}. 
Annotator 3, the only one consistently fooled by \gptpro\ outputs, relied too much on signs of informality (e.g., contractions, slang usage, usage of \emph{just} and \emph{actually}) as a sign of human writing, with 66.7\% of their explanations relating to formality (\autoref{fig:heatmap_ann3}). 
% TODO: Move the annotator 3 comments to her analysis in appendix

\subsection{{\colorbox{SkyBlue!70}{\faFlask\ Experiment 5:}} Are experts robust to humanization?}
\label{subsec:experiment5}

{\nolinenumbers
  \begin{footnotesize}
    \hspace{2.6em} \fcolorbox{SkyBlue!70}{SkyBlue!30}{\faCogs \hspace{0.1cm} \textbf{LLM:} \gptpro} \\
    \hspace*{2.6em} \fcolorbox{SkyBlue!70}{SkyBlue!30}{\faGhost \hspace{0.25cm} \textbf{Evasion tactic:} humanization}\vspace{0.2cm}
  \end{footnotesize}
}

Many users attempt to evade detectors via \emph{humanization} methods, which explicitly attempt to make AI-generated text more human-like \cite{wang_raft_2024,wang_humanizing_2024}.
Are such methods effective at evading our human experts? Since no well-established humanization methods exist, we first create our own humanizer by prompting \gptpro\ with a detailed set of instructions derived from expert explanations from Experiments 1-4.  Despite considerably degrading performance of many automatic detectors, humanization does not fool our experts in aggregate (``\gptpro\ {\small\textsc{humanized}}'' column of \autoref{tab:main_results}).

\paragraph{Implementing a prompt-based humanizer:} 
We modify the article generation prompt used in prior experiments to include instructions on specific AI signatures to avoid. 
To obtain these instructions, we pay our experts \$45 each to provide us with a list of clues that they look for during detection. We then 
then manually organize these disparate 
clues into a ``guidebook'' with different sections (e.g., vocabulary, tone) that each contain explanations and examples of how AI writing differs from human writing (see \autoref{tab:detection-guide} for a truncated version of the guidebook). Then, we prompt \gptpro\ with the guidebook and an instruction to generate an article that would \emph{evade} a detector that was following the guidebook (see~\autoref{tab:evader_prompt} for prompt).\footnote{We only use \gptpro\ for this experiment as the previous experiments show it was more difficult to detect than \gpt\ and \claude.} 
More details on the development of our humanization method can be found in \S\ref{app_subsec:evader}. 

\begin{figure}[htbp]
\centering
\includegraphics[width=1\linewidth]{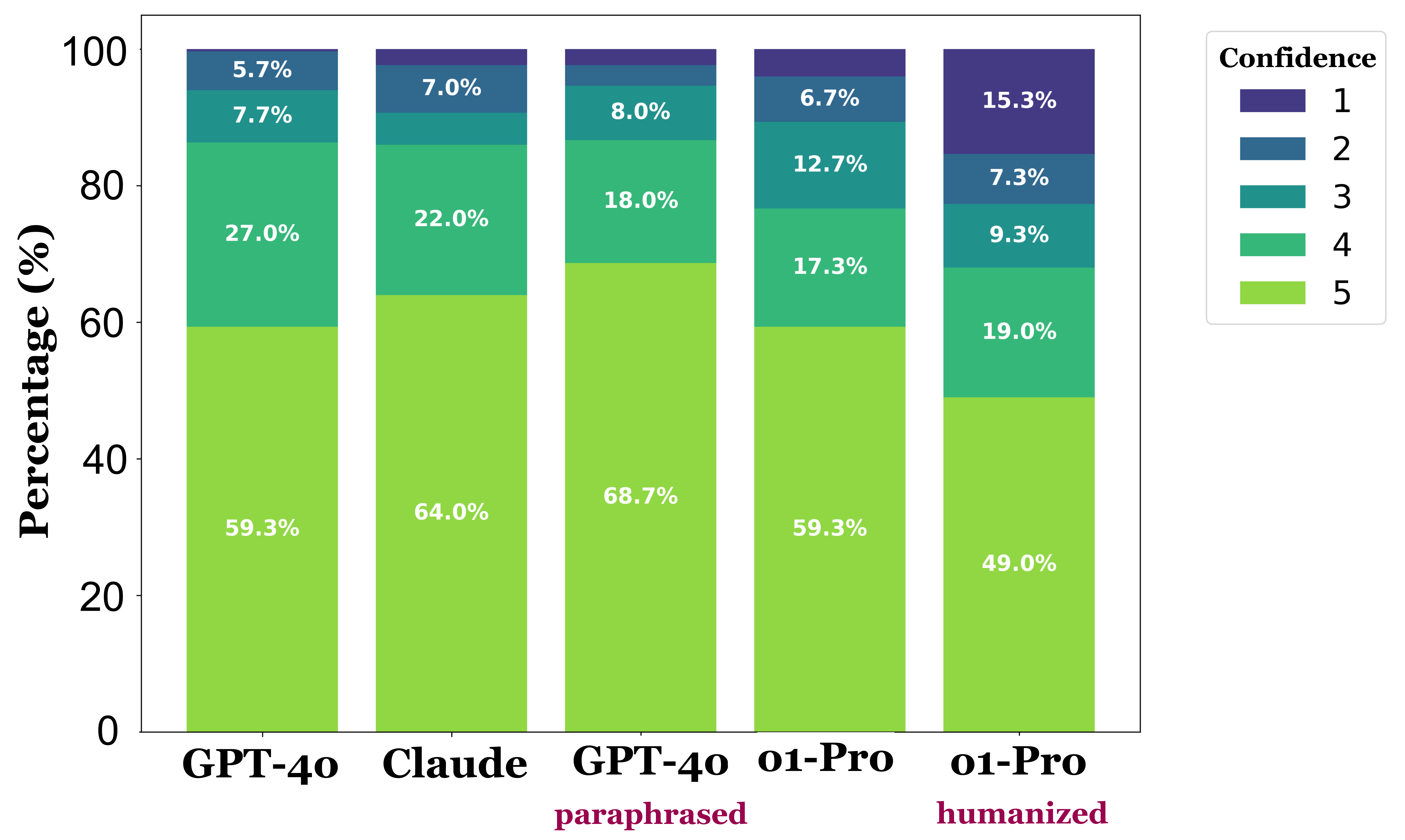}
\caption{Expert confidence in their decisions drops when judging humanized articles generated by \gptpro.}
\vspace{-5pt}
\label{fig:confidence_score_distribution}
\end{figure}

\paragraph{Experts remain robust to humanized articles:}
Despite our best efforts to generate articles that our experts would find undetectable, most of their detection rates remain largely unchanged from prior experiments, and the expert majority vote is again perfect on all 60 articles. 
Annotator 3, who struggled to detect non-humanized \gptpro\ articles in Experiment 4, performed remarkably poorly on this batch, achieving a TPR of zero by marking almost every article as human-written. 
While detection rates remained steady, overall confidence dropped, with 15.1\% of annotations having a confidence of 1, showing the increased difficulty presented by the humanized articles (\autoref{fig:confidence_score_distribution}). 

\paragraph{AI clues still remain after humanization: }
The clues used by experts remained consistent from prior experiments, showing that humanization does not completely remove the ``AI signature'' from texts. 
% TODO move to appendix
For example, while humanization increases variety in AI-generated names, often even including people relevant to the article who exist in real life, our experts found that humanized articles use titles for these people (e.g., Dr., Prof.)  much more frequently than in human-written articles. 
% Overall, 
Our results highlight the need for increased public research on humanization, as more advanced methods can challenge both human and automated detectors while offering better training data for more robust detection systems. 

% \begin{quote}
%     \small
%     \textcolor{violet}{\textit{Most of the word choices here are a weird mixture of casual and formal, such as with "a period of pristine professionalism" and "might subconsciously question the competence of a person" But more importantly, it's attempt to try and use metaphors and similes to describe the topic are... weird, they don't make sense. Also- why is the topic all about the Slack platform and people already in the workforce? It loses focus on that and devolves into weird attempts to provide detail, but in the wrong place.}}
%     \begin{flushright} 
%     \end{flushright}
% \end{quote}

%% file: sections/33-explanations.tex
\section{Analysis of expert performance} 
\label{sec:explainability}

We first provide in-depth comparisons between experts and automatic detectors.\footnote{We release our dataset of annotations under under the MIT License.} Next, we perform a fine-grained coding of expert explanations into different categories (e.g., vocabulary, originality, formality), which allows us to examine what they focus on during the different experiments. Finally, we analyze differences between annotators, 
% and what they focus on when they are incorrect, 
highlighting implications for the future training of human annotators for AI-generated text detection. 

\begin{table*}[!t]
\centering
\scriptsize
\scalebox{0.88}{
\begin{tabular}{p{0.1\textwidth}p{0.06\textwidth}p{0.34\textwidth}p{0.53\textwidth}}
\toprule
\multicolumn{1}{c}{\bf\textsc{Category}} & \multicolumn{1}{c}{\bf\textsc{Freq}} & \multicolumn{1}{c}{\bf\textsc{Definition}} & \multicolumn{1}{c}{\bf\textsc{Example Explanations}} \\
\midrule
\textsc{Vocabulary} & 53.1\% & LLMs use specific words and phrases more often than human writers, which often results in repetitive, unnatural, or overly complex wording. & \vspace{-2mm} \textbf{\color{teal}{Human}}: \textit{"Furthermore, I very much doubt AI would have used adventurous adjectives like `chunky', `musky' or `thin' to describe food. Nor would it have used verbs like `blitzing' or `bolstering'."} \newline \textbf{\color{purple}{AI (\textsc{o1-Humanized})}}: \emph{"Odd word choices: wheat that `stores' a lineage; genes that are `honed.'"}\\

\midrule
\textsc{Sentence} \textsc{Structure}  & 35.9\% & AI-generated sentences follow predictable patterns (e.g., high frequency of “not only … but also …”, or consistently listing three items), while human-written sentences vary more in terms of length. & \vspace{-2mm} \textbf{\color{teal}{Human}}: \textit{"Short choppy sentences and paragraphs."} \newline \textbf{\color{purple}{AI (\textsc{o1-Pro})}}: \textit{"One pattern I've been noticing with AI, and I think I've stated this before, is the comparison of `it's not just this, it's this' and I'm seeing it here, along with listings of specifically three ideas."}\\

\midrule
\textsc{Grammar \& Punctuation}  & 24.8\% & AI-generated text is usually grammatically perfect (also avoiding dashes and ellipses), while human-written text often contains minor errors. & \vspace{-2mm} \textbf{\color{teal}{Human}}: \textit{"There's a lot of variety in the article's grammar use, with dashes, brackets, quotes intermixed with sentences, and short spurts of comma sections throughout."} \newline \textbf{\color{purple}{AI (\textsc{GPT-4o-Para})}}:\textit{"there's nothing off about the grammar or syntax in this piece..."}\\

\midrule
\textsc{Originality} & 23.7\% & AI-generated writing is generally straightforward, “safe,” and lacking in surprises or humor, leaving annotators bored or disengaged. & \vspace{-2mm} \textbf{\color{teal}{Human}}: \textit{"it's offset by some great analogies and creative phrasing that works well to convey the topic, such as with "amateur sleuths", "catnip for a certain type of Reddit user." }\newline \textbf{\color{purple}{AI (\textsc{o1-Pro})}}: \textit{"What happens when AI tries to be creative? Penguins "stand on their own flippers"."}\\

\midrule
\textsc{Quotes}  & 22.3\% & AI-generated quotes sound overly formal, lack the varied nuances of real conversation, and often mirror the article’s main text too closely in style. & \vspace{-2mm}  \textbf{\color{teal}{Human}}: \textit{" The quotes being short snippets also makes me think they're real, as the writer had to find a way to fit them into the text, rather than them just perfectly stating either side's views."} \newline \textbf{\color{purple}{AI (\textsc{GPT-4o})}}: \textit{"The quotes also feel fake, every expert speaks the same way and it's too homogenous with the text."}\\

\midrule
\textsc{Clarity} & 19.5\% & AI-generated text often lacks concise flow by over-explaining or including irrelevant details, effectively “telling” rather than “showing”. & \vspace{-2mm} \textbf{\color{teal}Human}: \textit{"Words like "meander" are used, but are used sparingly to create better flow of ideas, and its writing style is simplified in the best way possible."} \newline \textbf{\color{purple}AI (\textsc{Claude-3.5-Sonnet})}: \textit{"The sentences are condensed to provide the best possible precision with its word choice, but the article lacks flow and clarity."}\\

\bottomrule
\end{tabular}}
\caption{Truncated taxonomy of clues used by experts to explain their detection decisions (see \autoref{tab:explanation_category_definitions_full} for full version). For each category, we report the frequency of explanations that mention that category (regardless of correctness) and provide sample explanations for human-written and AI-generated articles. 
While vocabulary is the most common clue, complex phenomena like originality, quotes and clarity are also distinguishing features.
}
\label{tab:explanation_category_definitions}
\end{table*}

\paragraph{Human experts vs. automatic detectors:}
We compare human experts with five AI detectors, including two closed-source classifiers, Pangram \cite{emi_technical_2024, masrour_damage_2025} and GPTZero \cite{tian2023gptzero}, and three open-source methods: Binoculars \cite{hans_spotting_2024}, Fast-DetectGPT \cite{bao_fast-detectgpt_2024}, and RADAR \cite{hu_radar_2023}.\footnote{We report TPR and FPR of automatic detectors using recommended thresholds whenever possible. For Fast-DetectGPT, we use a threshold calibrated to an FPR of 5\% on a held-out set of 40 human-written articles. Thresholds used are found in \S\ref{app_subsec:thresholds}. We evaluate Pangram in both its base and humanized modes, ignoring the ``Possibly AI'' label.}
\autoref{tab:main_results} shows that only Pangram Humanizers (average TPR of 99.3\% with FPR of 2.7\% for base model) matches the human expert majority vote, and it also
outperforms 
each expert individually, 
faltering  just slightly on humanized \gptpro\ articles. GPTZero struggles significantly on \gptpro\ with and without humanization. Open-source detectors degrade in the presence of paraphrasing and underperform both closed detectors.
We again note that our experts are \emph{untrained} at this detection task, and they could likely improve their individual performance if provided with feedback.

\begin{figure*}[t!]
\centering
\includegraphics[width=0.9\textwidth]{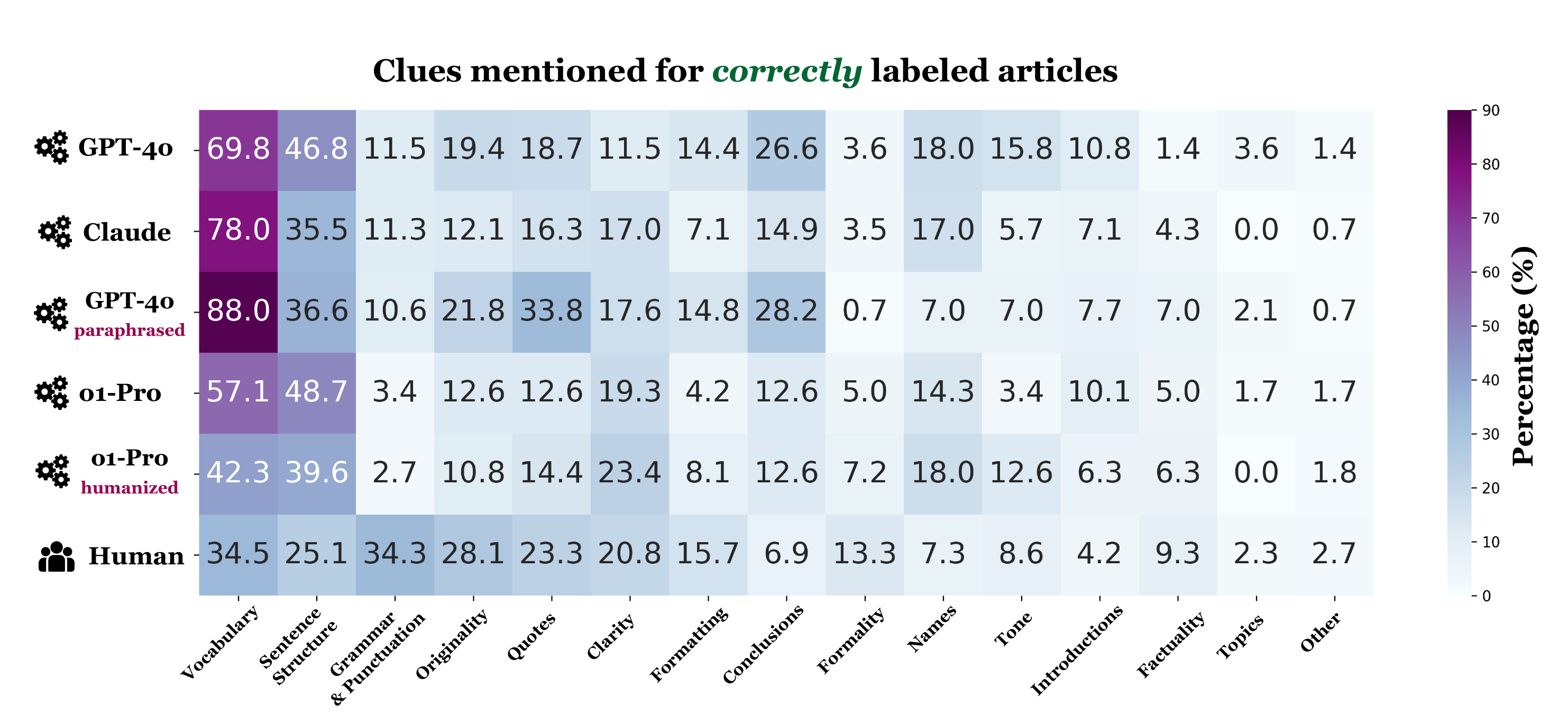}
\caption{A heatmap displaying the frequency with which annotators mentioned specific categories in their explanations when they were correct. Interestingly, vocabulary becomes a less frequent clue for \gptpro-generated articles, especially with humanization. A heat map of corresponding incorrect explanation is displayed in \autoref{fig:heatmaps_incorrect}.
Details of each category can be found in \autoref{tab:explanation_category_definitions}.}
\label{fig:heatmaps}
\end{figure*}

\paragraph{Coding expert explanations:}
In contrast to opaque automatic detectors, we can easily solicit explanations from human annotators. We use \textsc{GPT-4o} to code these explanations into a schema (\autoref{tab:explanation_category_definitions}) developed by the authors after careful manual analysis. Details of the coding process used to label explanations can be found in \S\ref{app_sec:comment_analysis}. Vocabulary-related clues are mentioned in the majority (53.1\%) of all explanations, while sentence structure (35.9\%), grammar (24.8\%), and originality (23.7\%) are also common. We note that many of these categories (e.g., originality, factuality, tone) are much more difficult to assess automatically than others (e.g., vocabulary), and these may currently be areas where humans have an advantage over automatic detectors.

\paragraph{When experts correctly detect AI-generated articles, what clues do they use?}

\autoref{fig:heatmaps} shows the frequency that clue categories are mentioned in explanations for which the expert makes the \emph{correct} decision. We observe several 
shifts as the complexity of the article generation setup increases: for example, 57.1\% of correct explanations about \gptpro\ articles mention vocabulary, compared to only 42.3\% for the humanized \gptpro\ articles prompted to avoid ``AI vocab''. 
Somewhat counterintuitively, explanations about paraphrased articles note AI vocab in 88\% of explanations, compared to only 69.8\% of non-paraphrased GPT-4o articles.
Quotations are mentioned in a surprisingly high 33.8\% of explanations about paraphrased articles: a close reading reveals that experts flagged quotes that were always in the same format and style (e.g., only placed at the end of each paragraph). 
Finally, the frequency of originality clues highlights the gulf in creativity between humans and LLMs \cite{tuhin_creative2024}.

\paragraph{Annotators don't always focus on the same clues:}
Our experts focus on different properties of the text to arrive at their decisions, as described further in \S\ref{app_subsec:individual_commentary}. Annotator 1 is the only one to pick up on ``AI names'', while
in fact, 63.3\% of \gpt\ and 70\% of \claude\ articles include either the name Emily or Sarah.
\footnote{\gptpro\ favors names of real people instead of fictional names when generating articles.} 
Annotators 2 and 3 emphasize grammaticality or the tendency of LLMs to always list examples in groupings of three. 
Annotator 4 focuses on article flow, analyzing the specificity of detail and motivation behind an article, 
and Annotator 5 examines how natural any quotations sound. This diversity explains why an ensemble of expert annotators performs so well and 
it also suggests room for training so that all experts are at least aware of most distinguishing features. 
implies additional training can improve individual experts.

\paragraph{When experts are incorrect, what clues lead them astray?}
\autoref{fig:heatmaps_incorrect} shows 
clue frequencies in
\emph{incorrect} 
expert explanations (see \S\ref{app_subsec:individual_commentary} for more).
For example, \gptpro's frequent use of contractions and colloquialisms fooled Annotator 3: they mention formality in \emph{none} of their explanations for Experiments 2 and 3, compared to 66.7\% of \gptpro\ and 83.3\% of humanized \gptpro\ explanations (\autoref{fig:heatmap_ann3}). 
Studying expert false positives is also insightful (\autoref{fig:heatmaps}): 31\% of explanations here mention vocabulary, typically when human-written content contains ``AI vocab'' like \emph{delve} and \emph{crucial}. 
% More prominently, 50\% of false positives focus on sentence structure, when humans write stylistically-similar sentences to those preferred by LLMs. 
% (e.g., contiguous blocks of sentences with similar lengths, multiple lists of three items)

\paragraph{When is it worth it to use human detection?}
Human experts deliver detailed, freeform explanations and confidence ratings - features that automatic detectors generally lack, as they usually provide only a scalar score or binary label, limiting their trustworthiness in critical settings \cite{lipton2018mythos}. For example, before accusing a student of plagiarism or flagging a high-stakes legal document, someone may want to point to specific properties of the text that raised suspicion, or ensure that a false positive won’t have severe consequences. In our study, the majority vote of five expert annotators—without any specialized training—misclassified only 1 out of 300 articles, on par with the most accurate commercial detector (Pangram), even under adversarial paraphrasing and humanization. However, manual review cannot scale to high-volume, low-stakes tasks such as online forum moderation, where automated systems remain the only practical option.

%% file: sections/4-explainable-text-detection.tex
\section{Can LLMs be prompted to mimic human expert detectors?}
\label{sec:explainable_detector}

We have established that human experts are more accurate, robust, and interpretable  than automatic detectors. Can we prompt LLMs to mimic the decision-making process of human experts? 
In this section, we develop a prompt-based detector that also generates explanations,
using the guidebook from \S\ref{subsec:experiment5}. While the approach 
promisingly outperforms Binoculars and RADAR, 
it lags human experts and closed-source detectors like Pangram.

\paragraph{Implementing a prompt-based detector:} 
We prompt both \gpt\ and \oone\ to decide whether a candidate text is human-written or AI-generated and explain its decision based on the criteria in the guidebook (see \autoref{tab:prompt_template_explainable_detector} for prompt template).\footnote{We set \texttt{temperature=0} for \gpt\ and reasoning efforts to `medium' for \oone\ (note that OpenAI has fixed \texttt{temperature=1} for \oone). Further experiments can be found in \autoref{tab:additional results}.}   While we can control this detector's behavior via its prompt (e.g., with and without chain-of-thought), it produces the label text instead of a scalar score, making it impossible to set an FPR threshold.

\paragraph{Prompt-based detection shows promise but struggles with humanization and high FPRs:}
For \gpt, the best detector configuration is achieved with both CoT and the guidebook (average TPR 78\%), while \oone\ is best with CoT but without the guidebook (54\%). We observe that \oone\ is more conservative than \gpt, demonstrating lower FPRs on average at the cost of lower TPR. 
Our best configuration with \gptnov\ performs comparably to Binoculars and Fast-DetectGPT, although 
at a much higher cost. We hypothesize that the gap between prompt-based detection and humans can be reduced by fine-tuning.

%% file: sections/5-related-work.tex
\section{Related work}
\label{sec:related_work}

\paragraph{Human detection of AI-generated text: }
Research conducted prior to ChatGPT's release concludes that while na\"ive annotators do not reliably detect AI-generated texts \cite{ippolito_automatic_2020, brown_language_2020, clark_all_2021, karpinska-etal-2021-perils}, some annotators 
% manage to 
perform very well \cite{ippolito_automatic_2020, dugan_real_2023}. Other research explores detection in~\cite{porter_ai-generated_2024} and other languages \cite{puccetti_ai_2024, wang2025humanliketextlikedhumans}. 
\citet{dugan_roft_2020} and \citet{dugan_real_2023} address a more complex task of identifying the boundary between human and AI-generated text.

\paragraph{Automatic detection: }
Successful automatic detection methods rely on either perplexity \cite{mitchell_detectgpt_2023, bao_fast-detectgpt_2024, hans_spotting_2024} or trained classifiers \cite{solaiman_release_2019, emi_technical_2024, verma_ghostbuster_2024}. 
Prior work compares detectors' performance in different domains and regimes~\cite{dugan_raid_2024,li_mage_2024,zhang_llm-as--coauthor_2024}. 
Detectors can be vulnerable to paraphrasing attacks \cite{krishna_paraphrasing_2023, sadasivan_can_2024} and style changes~\cite{doughman_exploring_2025}. Studies have also examined humanization attacks to bypass detection \cite{wang_humanizing_2024, lu_large_2024, shi_red_2024, wang_raft_2024}, while \citet{zhou_humanizing_2024} and \citet{masrour_damage_2025} boost detector robustness.

\paragraph{Analyzing differences between human-written and AI-generated text: } 
 Many frameworks group errors in AI-generated text 
 similarly to our paper
 \cite{gehrmann_gltr_2019, dou_is_2022}.
 \citet{ma_ai_2023} discover gaps in 
 depth and content quality between AI-generated and human-written scientific text,
while \citet{shaib_detection_2024} notes the repetitiveness of AI-generated text by introducing syntactic templates. 
\citet{ji_detecting_2024}
categorize human nonexpert detection explanations.

%% file: sections/6-conclusion.tex
\section{Conclusion}
\label{sec:conclusion}

Our paper demonstrates that a population of ``expert'' annotators---those who frequently use LLMs for writing-related tasks---are highly accurate and robust detectors of AI-generated text without any additional training. The majority vote of five such experts performs near perfectly on a dataset of 300 articles, outperforming all automatic detectors except the commercial Pangram model (which the experts match). 
Analysis of explanations provided by our expert annotators reveals that they pick up on not just vocabulary and sentence structure-related clues but also more complex properties like originality
and tone. We observe that 
experts
focus
on different aspects of the text, and we conjecture that with explicit training, human 
annotators can be made even more robust to advances in LLMs as well as evasion tactics (e.g., humanization). 
Future work can also explore human annotators working alongside automatic detectors 
like Pangram to improve detection accuracy and explainability. 
We find it apt to end with a comment from one of our experts about a particularly formulaic conclusion generated by \gpt:

\fancyepigraph{This time I went to the end of the piece and said: ``Hello, AI.''  There it was in all its glory: the ``testament'' serving ``as a beacon of hope and inspiration'' and ``demonstrating'' to us humans ``that anything is possible.'' Sometimes I feel sorry for AI—it must have a dreary time trying to satisfy its human interlocutor's desire to ``showcase'' advocacy, social change, inclusivity, gender equality, equity, and representation in an essay of fewer than a thousand words.}{\textsc{Annotator 5}}

%% file: sections/7-limitations.tex
\section*{Limitations}

Our study is limited to articles in American English, chosen for their consistent formatting and high quality (i.e., professionally written and proofread). We also did not investigate factual accuracy, as it did not appear to be a significant cue for our annotators, who covered a broad range of topics.
While we selected articles from reputable sources, there remains a possibility that some included AI-generated edits beyond our scope of detection.

%% file: sections/8-ethical_considerations.tex
\section*{Ethical Considerations}
This study was reviewed by the UMass 
Institutional Review Board 
(IRB \#5927) 
and deemed exempt. All annotators were briefed on the purpose of the project and provided informed consent prior to participating. Those who wished to be acknowledged by name explicitly agreed to do so in their consent forms.  We ensured fair compensation for annotators in recognition of their time and expertise. We acknowledge the potential risks of misinformation and hallucinated content, especially when AI outputs are presented as human-written. Our goal is to examine these issues and inform best practices, rather than endorse or facilitate deceptive uses of AI.

%% file: sections/9-ack.tex
\section*{Acknowledgments}

We would like to extend our gratitude to our Upwork expert annotators for their dedicated efforts, and link their Upwork accounts for future work requiring human expertise on AI text detection: Benjamin Stewart,\footnote{Benjamin's Upwork: \url{https://www.upwork.com/freelancers/~01cfc42298090772b8?mp_source=share}} Elizabeth Bergman,\footnote{Elizabeth's Upwork: \url{https://www.upwork.com/freelancers/~01e0d3569ef95d391d?mp_source=share}} Lynette Stewart,\footnote{Lynette's Upwork: \url{https://www.upwork.com/freelancers/~01cc5594cf2405ea5e?mp_source=share}} Roberta King,\footnote{Roberta's Upwork: \url{https://www.upwork.com/fl/robertaking?mp_source=share}} and Shannon West.\footnote{Shannon's Upwork: \url{https://www.upwork.com/freelancers/~01daaa89e049285294?mp_source=share}} We thank Pangram and GPTZero for providing credits to access their APIs. We are also grateful to the members of the UMass NLP lab for their insightful input. This project was partially supported by awards IIS-2202506 and IIS-2312949 from the National Science Foundation (NSF) as well as an award from Open Philanthropy. 

%% file: sections/appendix.tex
%%%%%%%%%%%%%%%%%%%%%%%%%%%%%%%%%%%%%%%%%%%%%%%%%%%%%%%%%%%%
%%%%%%%%%%%%%%%  HUMAN ANNOTATIONS %%%%%%%%%%%%%%%%%%%%%%%%%%%%%%%%%
%%%%%%%%%%%%%%%%%%%%%%%%%%%%%%%%%%%%%%%%%%%%%%%%%%%%%%%%%%%%
\section{Human Evaluation}
\label{app_sec:human_evaluation}

In this section of the appendix we provide additional details about our experts the data collection pipeline.

\paragraph{Annotators: } 
The annotations for the experiment 1 were done by 5 annotators recruited on Upwork. All annotators are native English speakers from the US or South Africa. All annotators hold university degrees, worked in varying professions, and had varying levels of familiarity with AI assistants like ChatGPT. One had never used AI, 2 had little experience, and 2 used AI every day. Our five expert annotators are native English speakers hailing from the US, UK, and South Africa. Most work as editors, writers, and proofreaders and have extensively used AI assistants. See \autoref{tab:survey_data} for more information about annotators.

\begin{table*}[t]
\centering
\resizebox{\linewidth}{!}{%
\begin{tabular}{@{}llllll@{}}
\toprule
\textbf{Annotator} & \textbf{Education Level} & \textbf{English Dialect} & \textbf{LLM Usage}      & \textbf{AI Models Used}  & \textbf{Occupation}  \\ 
\midrule
\multicolumn{6}{l}{\textbf{Non-Expert}} \\ 
\midrule
 - & Professional Degree or Doctorate  & American English  & Few times  & ChatGPT & Writer  \\
 - & Bachelor's Degree  & American English & Daily & ChatGPT, Gemini, Llama & Finance/Operations  \\
 - & Some college, no degree & American English  & Few times  & None   & Transcription, editing, data entry  \\
 - & Bachelor's Degree   & American English   & Never & None & English tutor, copywriter, author         
 \\ \midrule
\multicolumn{6}{l}{\textbf{Expert}} \\ 
\midrule
Annotator 1& Bachelor's Degree    & South African English  & Daily & ChatGPT, Microsoft Copilot & Freelance editing, writing, proofreading  \\
Annotator 2 & Bachelor's Degree & South African English & Weekly   & ChatGPT & Editing and Proofreading    \\
Annotator 3 & Master's Degree & British English & Daily  & ChatGPT  & Copyeditor and Proofreader  \\
Annotator 4  & Bachelor's Degree  & American English  & Weekly & ChatGPT, Llama, Huggingface models & Freelancer Content Writer   \\
Annotator 5 & Master's Degree & South African English  & Weekly& ChatGPT, Claude& Language teacher 
\\ \bottomrule
\end{tabular}
}
\caption{Survey of Annotators, specifically their backgrounds relating to LLM usage and field of work. Note that expert Annotator \#1 was one of the original 5 annotators (along with the 4 non-experts) and remained an annotator for all expert trials.}
\label{tab:survey_data}
\end{table*}

\paragraph{Collecting Human Annotations}

\begin{figure*}[t]
    \centering
    \begin{subfigure}[b]{0.45\textwidth}
        \includegraphics[width=\linewidth]{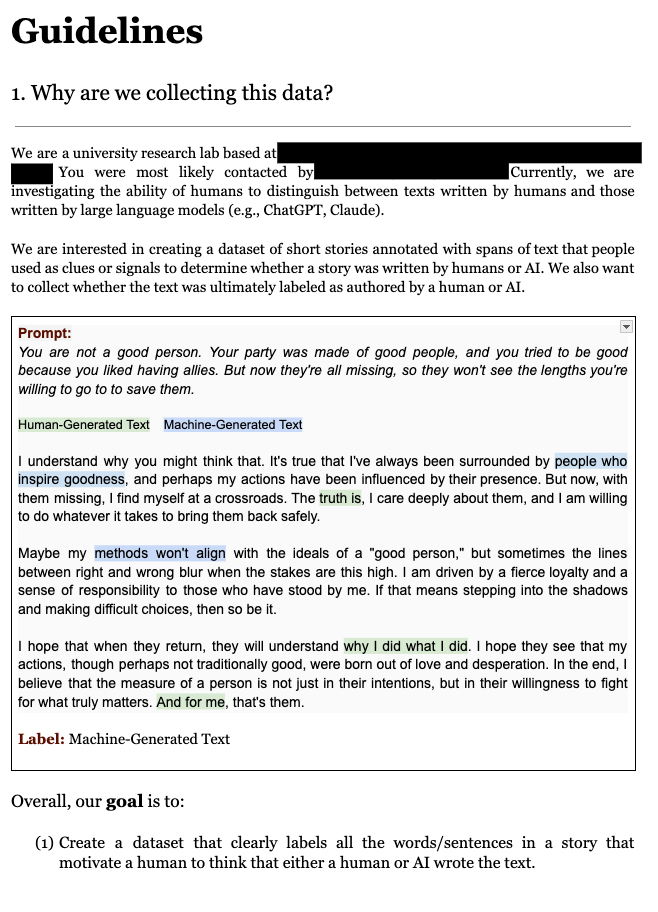}
        \caption{Guidelines: Page 1}
    \end{subfigure}
    \hfill
    \begin{subfigure}[b]{0.45\textwidth}
        \includegraphics[width=\linewidth]{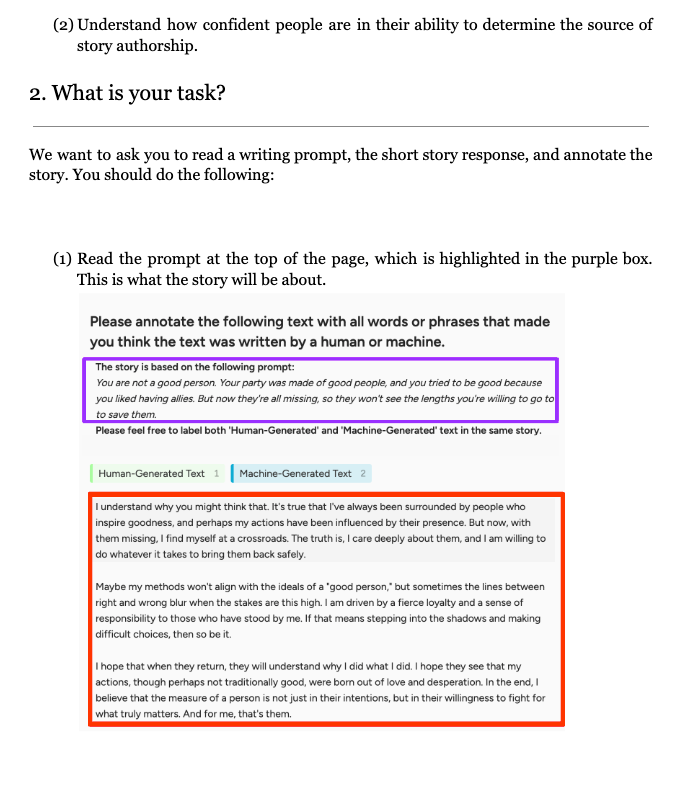}
        \caption{Guidelines: Page 2}
    \end{subfigure}
    \begin{subfigure}[b]{0.45\textwidth}
        \includegraphics[width=\linewidth]{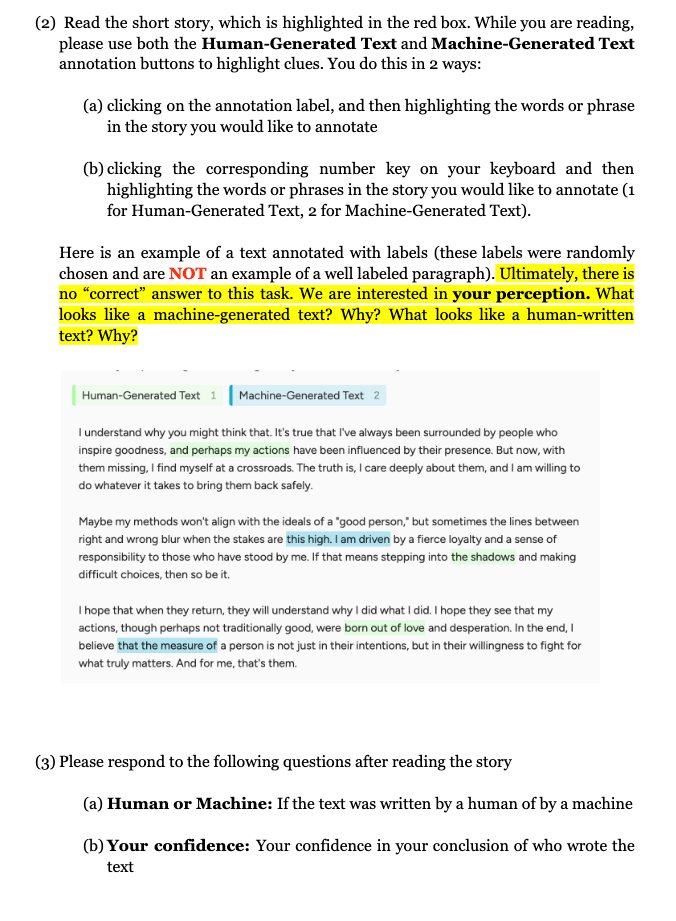}
        \caption{Guidelines: Page 3}
    \end{subfigure}
    \begin{subfigure}[b]{0.45\textwidth}
        \includegraphics[width=\linewidth]{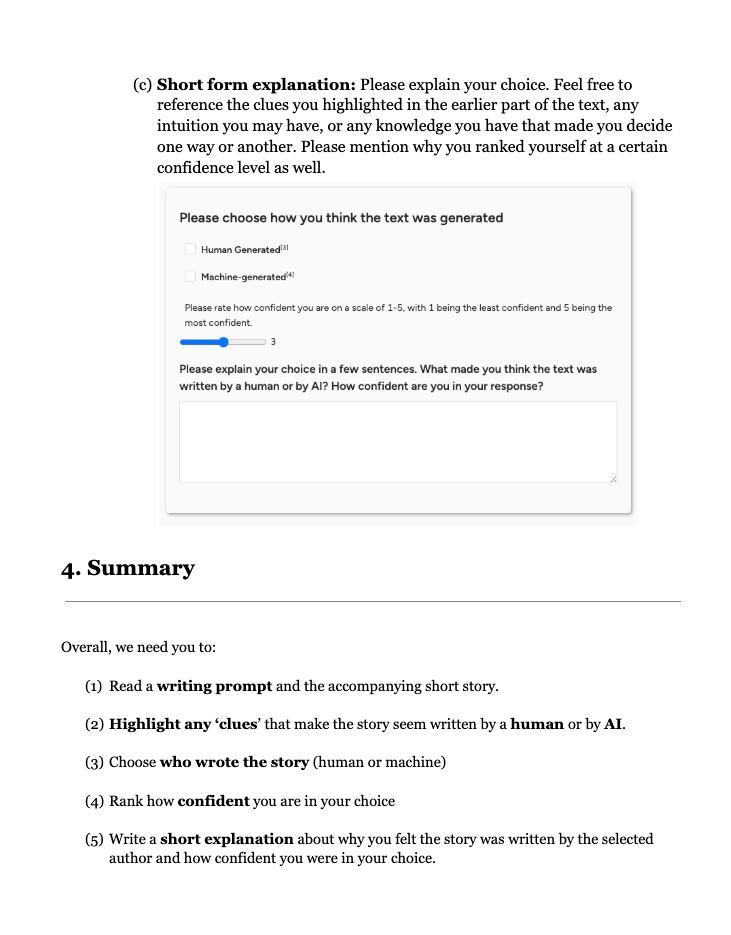}
        \caption{Guidelines: Page 4}
    \end{subfigure}
    \caption{Guidelines provided to the annotators for the annotation task. The annotators were also provided additional examples and guidance during the data collection process.}
    \label{fig:guidelines_for_annotators}
\end{figure*}

\begin{figure}[tbp]
  \includegraphics[width=.85\linewidth]{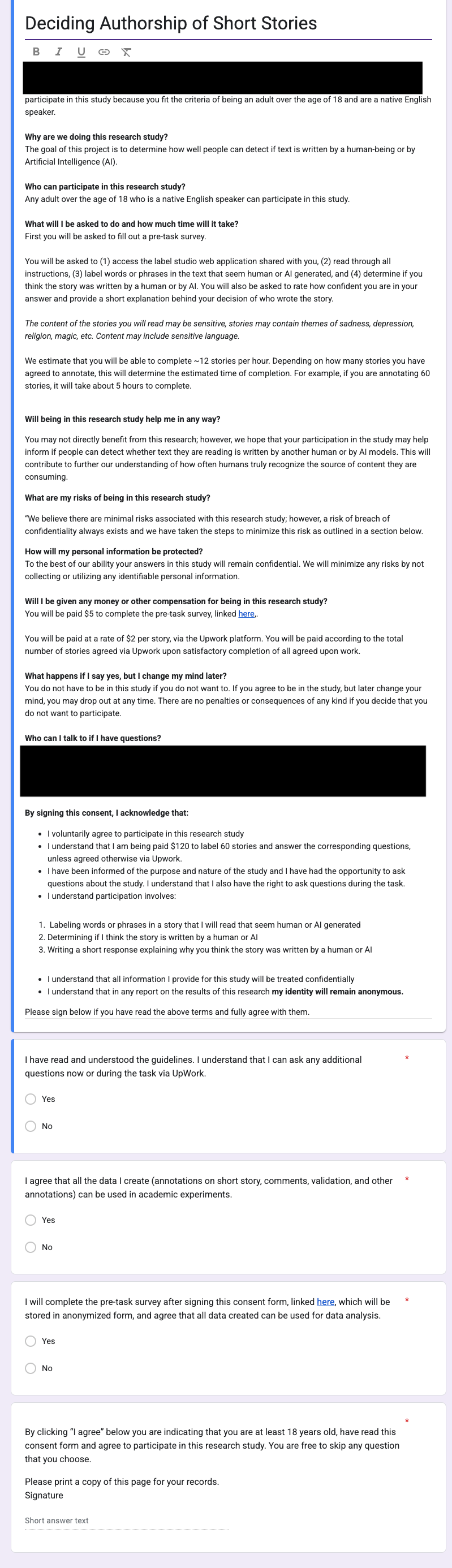} 
  \caption{Consent form which the annotators were asked to sign via \texttt{GoogleForms} before collecting the data.}
  \label{fig:consent_form}
\end{figure}

\begin{figure*}[tbp]
  \includegraphics[width=.85\linewidth]{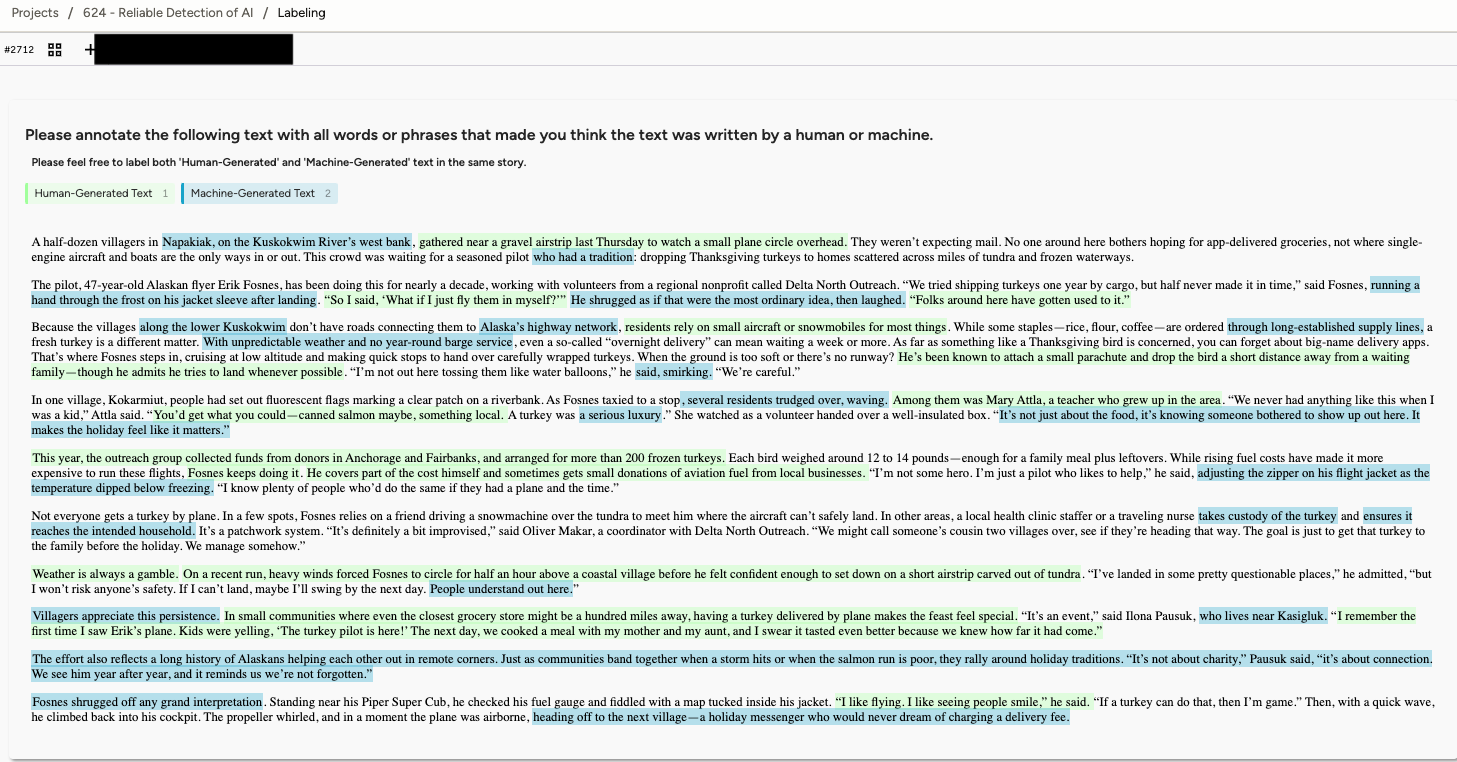}  
  \includegraphics[width=.85\linewidth]{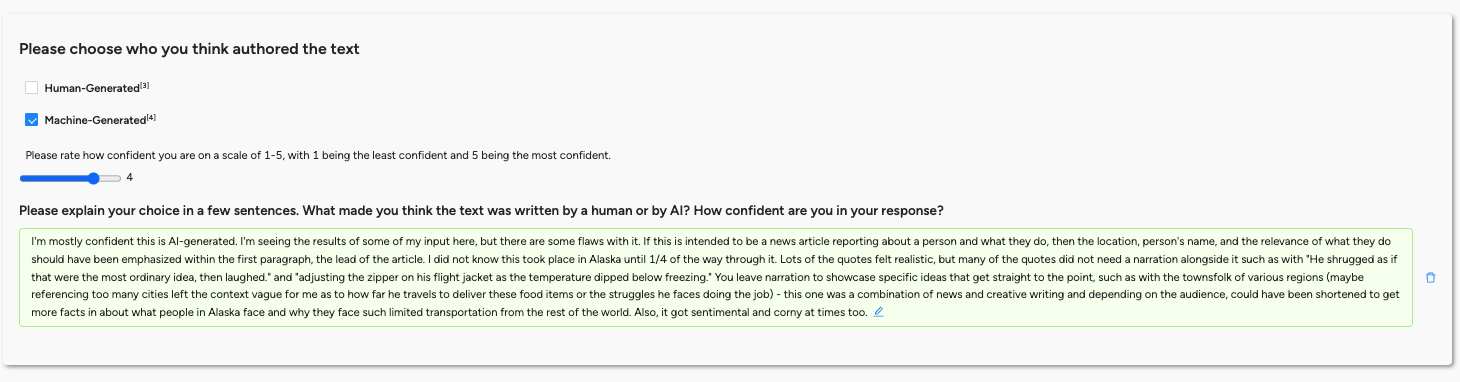}  
  \caption{Interface for annotators, with an example annotation from Annotator \#4 with a humanized article from \S\ref{subsec:experiment5}. This is the same article displayed in \autoref{fig:main}. An annotator can highlight texts, make their decision, put confidence, and write an explanation. This AI-generated article was based off of \href{https://apnews.com/article/alaska-turkeys-dropped-from-airplanes-9865b07e98826a77dd3570c679600f1a}{\textit{In Alaska, a pilot drops turkeys to rural homes for Thanksgiving}}, written by Mark Thiessen \& Becky Bohrer, and was originally published by Associated Press on Nov. 28, 2024. }
  \label{fig:interface}
\end{figure*}

All annotators were required to read the guidelines (\autoref{fig:guidelines_for_annotators}) and sign a consent form (\autoref{fig:consent_form}) prior to the labeling task. Collecting all labels usually required additional communication with the annotators, resulting in about 20 hours of work from the author involved in this process. We estimate that the annotators were able to read and label between 8 and 12 articles per hour based on self-reported time and records in the spreadsheets. \autoref{fig:interface} shows the interface annotators use to complete the annotation process. They read and highlight an article, then complete the annotation by providing their decision, confidence score, and explanation. Since completing annotations for 60 article batches takes a long time (estimated 6-8 hours of work), we have implemented an interface that made it possible for the annotators to save their work and come back at any time, allowing annotators to allocate their time as they saw best. Annotators were given 1 week to complete batches of 60 articles, with more time allowed when needed. Note that unlike \autoref{fig:main}, annotators did \emph{not} see the titles of the article when completing annotations. This decision was made to prevent obvious linkage of AI-generated and human-written text pairs. 

\paragraph{Study Design}: We employ a within-subjects design for our experiments, where each annotator judges both the human-written and AI-generated articles. This reduces variability from individual differences and requires fewer annotators \cite{Allen2017-withinsubject}. To further minimize bias, annotators are unaware of the pairing, and the article order is randomized. 

\subsection{Finding Expert Annotators}
\label{app_subsec:finding_expert_annotators}

In our pilot study (\S\ref{app_sec:pilot_study}), only the annotator who uses LLMs daily to edit AI-generated content achieved high accuracy. Prior work has noted on average, annotators cannot reliably detect AI-generated content \cite{brown_language_2020,clark_all_2021}, but vary widely in ability \cite{dugan_real_2023}; in this paper, we aim to investigate the \emph{upper bound} of human performance. To find this limit, 
 and understand the lexical clues humans use to recognize AI-generated content, 
 we recruit four additional annotators with similar backgrounds to the high-performing annotator. 

The annotations for understanding the limitations of humans to detect AI-generated texts were done by 5 annotators who passed performance requirements based on the article experiment of \autoref{subsec:experiment1}. Only one annotator from the original experiment met these requirements. We recruited 10 more native English speakers on Upwork to take a 5 question sample of the original article task, questions which at most 2 out of 5 of the original annotators got correct. These 10 recruited annotators all had experience editing LLM generated content, frequently used LLMs, and had a professional background in editing or writing. Those who got at least 4 out of 5 correct (80\%) given the rest of the 60 articles, and annotators who got at least 90\% correct (54/60) were considered to be experts and recruited for the rest of the labeling rounds. 5 out of 10 annotators passed the 5 question trial; out of those 5, 4 passed the 60 question set.

While some individuals unfamiliar with LLMs may prove to be reliable detectors, we focus on the reliability of people who frequently use LLMs for writing tasks (e.g. English teachers, editors, publishers), as our pilot study suggests their expertise enables them to consistently identify systematic patterns characteristic of AI-generated text.

%%%%%%%%%%%%%%%%%%%%%%%%%%%%%%%%%%%%%%%%%%%%%%%%%%%%%%%%%%%%
%%%%%%%%%%%%%%%  DATASET %%%%%%%%%%%%%%%%%%%%%%%%%%%%%%%%%
%%%%%%%%%%%%%%%%%%%%%%%%%%%%%%%%%%%%%%%%%%%%%%%%%%%%%%%%%%%%

\section{Dataset}
\label{app:dataset_info}

In this section of the appendix we provide more details on our article corpus ({\ref{app_subsec:corpus}), how our AI articles were generated (\ref{app_subsec:ai_text_generation}).

\subsection{Article Corpus}
\label{app_subsec:corpus}
Here we include more details about the articles collected for this study.
\autoref{tab:article_info} lists all publications of articles included in the corpus,\footnote{All publications in the corpus were purchased by the researchers.}  with section distribution presented in \autoref{fig:section_distribution}. \autoref{app_tab:summary_stats_dataset} provides the statistics for articles by publication.

\begin{table*}[!t]
\centering
\resizebox{\textwidth}{!}{
\begin{tabular}{c c c}
\toprule
\textbf{Publication} & \textbf{Example Sections} & \textbf{Pub. Date Range} \\
\midrule
\textit{Associated Press} & Science, Oddities, Animals & May 15, 2024 - Dec 5, 2024 \\
\textit{Discover Magazine} & Mind, The Sciences, Environment, Planet Earth & July 10, 2024 - Nov 16, 2024 \\
\textit{National Geographic} & Animals, Environment, Science, History \& Culture, Travel & Feb 8, 2023 - Nov 19th, 2024  \\
\textit{New York Times} & US News, Science, Travel, Arts & July 19, 2024 - Dec 6, 2024  \\
\textit{Readers Digest} & Knowledge, Holidays & March 16, 2023 - Nov 15, 2024 \\
\textit{Scientific American} & Mind \& Brain, Social Sciences, Technology & May 3, 2024 - Nov 22, 2024, 2024  \\
\textit{Smithsonian Magazine} & Smart News, Mind \& Body, History, Innovation, Travel, Science & Oct 7, 2022 - Dec 8, 2024  \\
\textit{Wall Street Journal} & Science, Personal Technology, Workplace & July 7, 2023 - Oct 29, 2024  \\
\bottomrule
\end{tabular}
}

\caption{List of publications included in \name. The section is provided as listed as the section of the publication website where the article was published. All articles were taken from publications that wrote using American English.}

\label{tab:article_info}
\end{table*}

\begin{figure}[tbp]
  \includegraphics[width=1\linewidth]{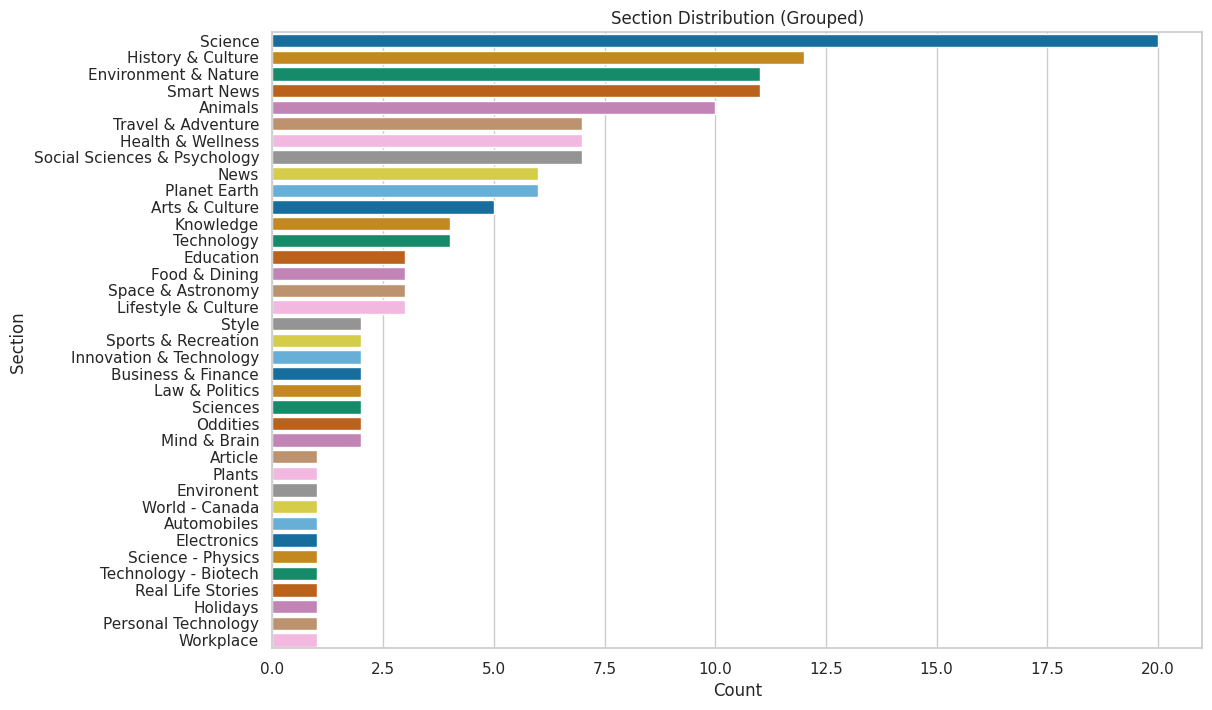} 
  \caption{Section distribution of articles across all trials.}
  \label{fig:section_distribution}
\end{figure}

\begin{table}[t]
\centering
\setlength{\tabcolsep}{3pt} 
\renewcommand{\arraystretch}{1.3} 
\footnotesize
\resizebox{\columnwidth}{!}{%
\begin{tabular}{c p{3cm} p{2.2cm} p{2.1cm}}
\hline
\textbf{Exp \#} & \textbf{Model} & \textbf{Human-written (\textit{n=30})} & \textbf{AI-generated (\textit{n=30})} \\ \hline
1 & \gpt  & 652.5\textsubscript{170.9}  & 625.1\textsubscript{126.9} \\ 
2 & \claude & 793.1\textsubscript{170.3} & 710.8\textsubscript{133.5} \\ 
3 & \gpt\ paraphrased & 777.8\textsubscript{175.9} & 654.9\textsubscript{110.6} \\ 
4 & \gptpro & 728.4\textsubscript{155.1}  & 813.8\textsubscript{233.6} \\ 
5 & \gptpro\ humanized & 739.0\textsubscript{128.3} & 815.6\textsubscript{171.2} \\ \hline
\end{tabular}%
}
\caption{Mean and standard deviation (subcripted) of article length in words across experiments, computed by splitting on whitespaces.}
\label{tab:article_lengths}
\end{table}

\begin{table*}[t]
\centering
\footnotesize
\resizebox{\textwidth}{!}{%
\begin{tabular}{l c c c c c c c c}
\hline
\addlinespace
& \multicolumn{8}{c}{\textbf{Articles}} \\ 
\cmidrule(lr){2-9}
 & {\renewcommand{\arraystretch}{1}\begin{tabular}[c]{@{}c@{}}\textbf{Associated Press}\\ \textit{(n=15)}\end{tabular}}
 & {\renewcommand{\arraystretch}{1}\begin{tabular}[c]{@{}c@{}}\textbf{Discover}\\ \textit{(n=20)}\end{tabular}}
 & {\renewcommand{\arraystretch}{1}\begin{tabular}[c]{@{}c@{}}\textbf{National Geographic}\\ \textit{(n=25)}\end{tabular}}
 & {\renewcommand{\arraystretch}{1}\begin{tabular}[c]{@{}c@{}}\textbf{New York Times}\\ \textit{(n=20)}\end{tabular}}
 & {\renewcommand{\arraystretch}{1}\begin{tabular}[c]{@{}c@{}}\textbf{Readers Digest}\\ \textit{(n=15)}\end{tabular}}
 & {\renewcommand{\arraystretch}{1}\begin{tabular}[c]{@{}c@{}}\textbf{Scientific American}\\ \textit{(n=15)}\end{tabular}}
 & {\renewcommand{\arraystretch}{1}\begin{tabular}[c]{@{}c@{}}\textbf{Smithsonian Magazine}\\ \textit{(n=25)}\end{tabular}} % Example new column
 & {\renewcommand{\arraystretch}{1}\begin{tabular}[c]{@{}c@{}}\textbf{Wall Street Journal}\\ \textit{(n=15)}\end{tabular}} % Another new column
\\
\hline
\addlinespace
\multicolumn{2}{c}{\textsc{Tokens} (\texttt{tiktoken})} & & &  & & \\
\cmidrule(r){1-2} 
Mean & 860.0 & 949.5 & 1070.7 & 1009.1 & 865.6 & 874.2 & 922.0 & 927.5 \\
St. Dev & 240.7 & 229.9 & 203.2 & 138.4 & 171.9 & 347.1 & 210.7 & 202.5 \\
Max & 1236.0 & 1463.0 & 1497.0 & 1254.0 & 1318.0 & 1760.0 & 1454.0 & 1336.0 \\
Min & 442.0 & 561.0 & 710.0 & 645.0 & 634.0 & 455.0 & 532.0 & 608.0 \\
\midrule
\multicolumn{2}{c}{\textsc{Words} (\texttt{whitespace})} & & & & & \\
\cmidrule(r){1-2}
Mean & 676.4 & 734.6 & 809.7 & 784.1 & 675.3 & 684.2 & 708.1 & 720.5 \\
St. Dev& 180.8 & 174.5 & 149.9 & 106.3 & 136.6 & 265.9 & 160.0 & 153.9 \\
Max & 1026.0 & 1117.0 & 1102.0 & 951.0 & 1008.0 & 1352.0 & 1095.0 & 1019.0 \\
Min & 342.0 & 415.0 & 536.0 & 513.0 & 501.0 & 362.0 & 388.0 & 491.0 \\
\hline
\end{tabular}
}
\caption{Number of tokens and words across articles by source.}

\label{app_tab:summary_stats_dataset}
\end{table*}

\subsection{AI Text Generation}
\label{app_subsec:ai_text_generation}
We generate the articles by prompting the models with the articles title, subtitle, approximate length, publication, and section. For stories, we prompt models with the title of the reddit thread. All closed-source models were prompted using the provider's API, \footnote{The estimated cost for generating all data with each model is as follows: \gpt{} \$1.85USD, \gptpro{} \$3.81 USD, \claude{} \$0.51 USD}. All models were prompted
and articles with the prompt presented in \autoref{tab:prompt_template_article}.
Instructions for rounding to an approximate length were included to ensure that articles on the same topic would be of similar length. Statistics on the distribution of lengths by trial are presented in \autoref{fig:word_counts_by_experiment}. An example article generation prompt is below:

\begin{quote}
\footnotesize
\texttt{You are given the following title and subtitle of a general article from the Science section of New York Times and asked to write a corresponding article of around 750 words. Include quotations from relevant experts and make sure the article is concise and easily understandable to a lay audience. \\\\ 
Title: The Science That Makes Baseball Mud ‘Magical’ \\\\
Subtitle: Scientists dug up the real dirt on the substance applied to all the baseballs used in the major leagues. 
\\ \\
Article: }
\end{quote}

\begin{table}[t]
    \setlength{\tabcolsep}{4pt}
    \centering
    \resizebox{0.9\columnwidth}{!}{%
    \begin{tabular}{c p{9cm}}
    \toprule
         & \multicolumn{1}{c}{\bf Article Generation Prompt} \\
    \midrule
     \noalign{\vskip 1mm}
     & \texttt{You are given the following title and subtitle of a general article from the \textbf{YOUR SECTION} section and asked to write a corresponding article of around \textbf{YOUR WORD COUNT} words. Include quotations from relevant experts and make sure the article is concise and easily understandable to a lay audience. }\\
     \noalign{\vskip 2mm} \\
     & \texttt{Title: \textbf{YOUR TITLE}}\\
      & \texttt{Subtitle: \textbf{YOUR SUBTITLE}}\\
      \noalign{\vskip 2mm}
    \bottomrule
    \end{tabular}
    }
    \caption{Prompt Template for Experiment 2 Paraphrasing} 
    \label{tab:prompt_template_article}
\end{table}

\begin{figure}[htbp]
\centering
\includegraphics[width=1\linewidth]{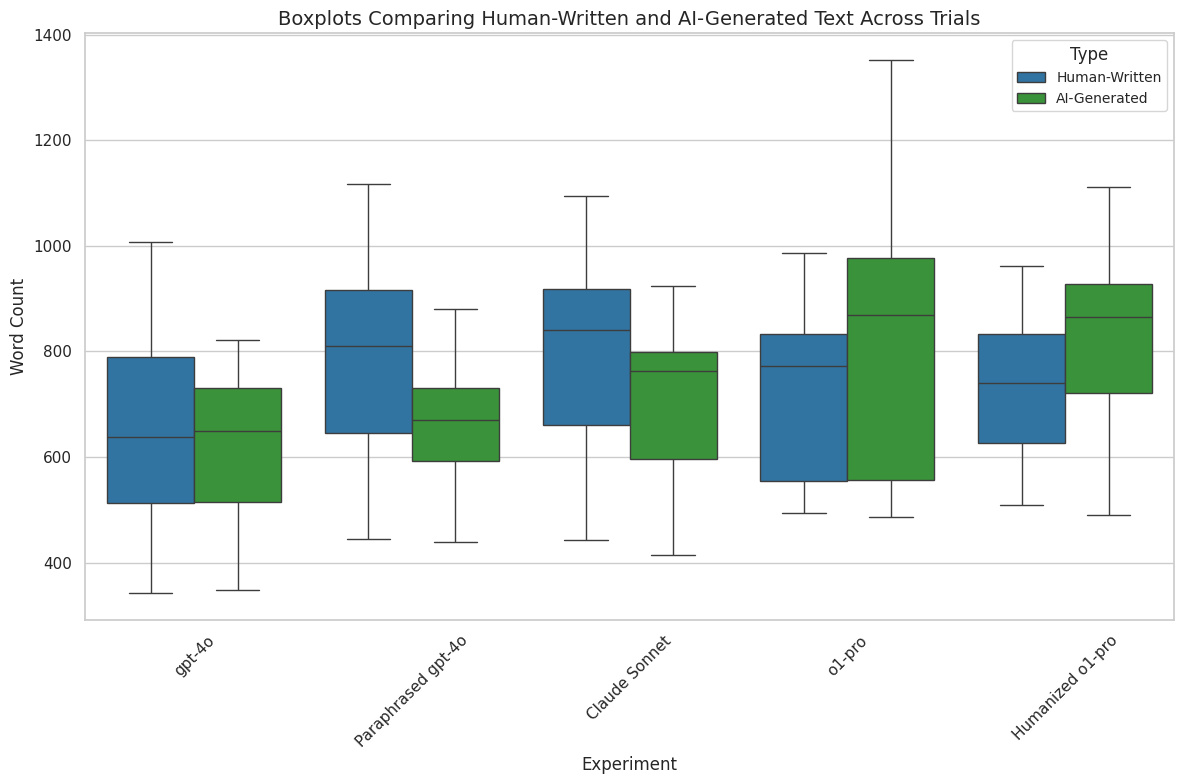}
\caption{Word Count Distribution Per Experiment}.
\label{fig:word_counts_by_experiment}
\end{figure}

\paragraph{Paraphrasing Attack}

 We use the sentence-level paraphrasing approach outlined in PostMark \cite{chang_postmark_2024}, changing the exact language in the prompts slightly for our use case. This approach paraphrases the text on a sentence by sentence level. For the initial sentence, only that sentence is given to be paraphrased. For all following sentences, the portion of the text already paraphrased is passed to the LLM, intending to improve the overall flow of the paraphrased article. Additionally, since our experts easily identified AI-generated named entities (e.g., Sarah Thompson) in the previous experiments, we extract named entities from the human-written article and instruct \gpt\ to include these names in its generated article. We use GPT-4o as the paraphraser model with a temperature set to 0. The prompt for the initial sentence can be found in \autoref{tab:prompt_template_paraphrase_init} and the prompt for all following sentences can be found in \autoref{tab:prompt_template_paraphrase_sent}.  
 
\begin{table}[t]
    \setlength{\tabcolsep}{4pt}
    \centering
    \resizebox{0.9\columnwidth}{!}{%
    \begin{tabular}{c p{9cm}}
    \toprule
         & \multicolumn{1}{c}{\bf Paraphrase Prompt (Initial Sentence Only)} \\
    \midrule
     \noalign{\vskip 1mm}
     & \texttt{Paraphrase the given sentence. Only return the paraphrased sentence in your response. Make it seem like a human wrote the article and that it is from the {\textbf{YOUR SECTION}} section of {\textbf{YOUR PUBLICATION}}. }\\
     \noalign{\vskip 2mm} \\
      & \texttt{Sentence to paraphrase: \textbf{YOUR SENTENCE}}\\
      \noalign{\vskip 2mm}
     & \texttt{Your paraphrase of the sentence: } \\
     \noalign{\vskip 2mm}
    \bottomrule
    \end{tabular}
    }
    \caption{Prompt Template for Experiment 2 Paraphrasing} 
    \label{tab:prompt_template_paraphrase_init}
\end{table}

\begin{table}[t]
    \setlength{\tabcolsep}{4pt}
    \centering
    \resizebox{0.9\columnwidth}{!}{%
    \begin{tabular}{c p{9cm}}
    \toprule
         & \multicolumn{1}{c}{\bf Paraphrase Prompt} \\
    \midrule
     \noalign{\vskip 1mm}
     & \texttt{Paraphrase the given sentence. Only return the paraphrased sentence in your response. Make it seem like a human wrote the article and that it is from the {\textbf{YOUR SECTION}} section of {\textbf{YOUR PUBLICATION}}.}\\
     \noalign{\vskip 2mm} \\
     & \texttt{Previous context: \textbf{YOUR PARAPHRASED ARTICLE SO FAR}}\\
      & \texttt{Sentence to paraphrase: \textbf{YOUR SENTENCE}}\\
      \noalign{\vskip 2mm}
     & \texttt{Your paraphrase of the sentence: } \\
     \noalign{\vskip 2mm}
    \bottomrule
    \end{tabular}
    }
    \caption{Prompt Template for Experiment 2 Paraphrasing} 
    \label{tab:prompt_template_paraphrase_sent}
\end{table}

\paragraph{Humanization Efforts: }
\label{app_subsec:evader}

To humanize the articles in \S\ref{subsec:experiment5}, we tried many prompt-based approaches before settling on a prompting framework that was capable of evading the first author's own AI-detection skills. Initially, we tried the following approaches that humanized already generated texts:
\begin{enumerate}
    \item \textbf{Generate, then Humanize}: Firstly, we tried to replicate what we believe most students or novice LLM users would do to humanize their AI-generated text. We asked \gptpro\ to first generate an article using the prompt template depicted in \autoref{tab:prompt_template_article}, then instructed the model to make it sound more human. Qualitative analysis from the first author found this method to result in an effect similar to paraphrasing. 
    \item \textbf{Step-by-Step}: To add a fine-grained approach to the humanization efforts, we next tried a step-by-step approach to humanization. We generated a base article, then iteratively prompted the model to humanize the article, focusing on a different element at each step. For example, we first asked it to make the article more creative, then alter the tone. Final steps included editing for grammar and replacing typical AI vocab. While this was better, the first author observed that the numerous calls to LLMs was adding \emph{more} characteristics of AI, so we abonded this method.
    \item \textbf{Two-step Humanization}: Next, we tried taking a 2-step approach to humanization. We gave the humanizing LLM the AI detection guide found in \autoref{tab:detection-guide}, and asked the LLM to make a list of everything identifiable as AI-generated in the article, and suggested edits. Then, giving the LLM the list of suggested edits, we prompted the LLM to edit the article to make it sound more human-written. Out of the humanization efforts, this yielded the best results.
\end{enumerate}

\paragraph{Evader Details: } While approaches to humanization were improving, we found many of the largest identifiable traits of AI were due to the base generation. We decided to switch our approach from humanizing already existing AI-generated text to having an LLM generate the article in a humanized fashion all in one step. Our final prompt includes the set of instructions, examples of reference articles, detection guide, and the initial article prompt. The evader prompt template can be found in \autoref{tab:evader_prompt}. The examples used were human-written and AI-generated articles from Experiments 1,2,3, and 4 from the same source. For example, when evading an article from New York Times, we provided all human-written New York Times articles and AI-generated articles based off those article titles.

\begin{table*}[t]
\centering
\footnotesize
\begin{tabular}{p{.95\linewidth}}
\toprule
\texttt{Guide to Distinguishing AI-Generated Text from Human Writing Template}\\
\midrule
\texttt{\#\# Vocabulary / Word Choice Patterns}\\
\texttt{- Certain words crop up unusually frequently throughout AI-generated text compared to human writing}\\
    \texttt{- Words like ‘delve’ and ‘tapestry’ are overused in AI-generated text but infrequently used in human writing ...}\\ \\
\texttt{\#\# Grammar}\\
\texttt{- Human writing generally less strictly adheres to English grammar rules and punctuation than AI-generated text}\\
\texttt{- AI-generated text uses a very formal writing style unless explicitly told not to ...}\\ \\
\texttt{\#\# Sentence Structure}\\
\texttt{- AI-generated sentences often follow the complex sentence structure, with multiple dependent and independent clauses, while human writing contains more of a mixture of simple, complex, and compound sentences}\\
    \texttt{- example AI-generated sentence: “When it comes to celebrating Halloween, this holiday is a testament to the importance of empathy and community.” ... }\\ \\
\texttt{\#\# Formatting }\\
\texttt{- When AI makes lists, it typically uses the format of creating a bold header per bullet point, followed by a colon and then description of that list item. } \\
\texttt{- If a book title is referenced in a text, AI-generated text will always italicize the title, while human writing does not always follow this convention.} \\
\texttt{- Some pieces of text, especially articles and essays, contain headers and sub-headers. The headers written by AI are quite repetitive ...}\\ \\
\texttt{\#\# Tone }\\
\texttt{- The tone of AI-written text is flowery and formal, and its sentences are frequently structured as a reflective, onlooking statement, regardless of topic.  }\\
\texttt{- AI tends to be inherently positive, attempting to emotionally uplift the reader, especially towards the conclusion.  }\\
\texttt{- AI prioritizes efficiency, sometimes sacrificing clarity or depth in its messaging ... }\\ \\
\texttt{\#\# Introductions}\\
\texttt{- AI-written introductions often contain a strong scene-opener with a description of a specific time or place, such as "On a drab November morning..." or "On December 8, 1660, a London audience gathered ..." ...}\\ \\
\texttt{\#\# Conclusions}\\
\texttt{- AI-generated text always ends with a neat conclusion, instead of just ending the article naturally. }\\
\texttt{- AI-generated conclusions are often overly long and summarize everything that has already been written in an article ...}\\ \\
\texttt{\#\# Content }\\
\texttt{- Unless specifically prompted, AI will avoid controversial topics at all costs.}  
\texttt{- AI will avoid any type of swear word, including mild ones like ‘darn’, or any other offensive vocabulary ...  }\\ \\
\texttt{\#\# Contextual Accuracy and Factuality}\\
\texttt{- Human writing in the domain of non-fiction is factually accurate and contains many specific factual claims. }\\
\texttt{- In human writing, people, places, brands, and other named objects can be verified or are highly plausible ...}\\ \\
\texttt{ \#\# Creativity \& Originality  }\\
\texttt{- AI-generated text is much less creative than that of humans, lacking originality and sticking to an ‘obvious’ way to answer a prompt. }\\
\texttt{- Humans incorporate twists, unexpected insights, and twists that AI hasn’t seemed to master quite yet ...}\\ \\
\bottomrule
\end{tabular}
\caption{A truncated version of the AI Text Detection Guide.}
\label{tab:detection-guide}
\end{table*}

% AI WORDS TABLE
\begin{table*}[t]
    \setlength{\tabcolsep}{4pt} 
    \centering
    \begingroup
        \renewcommand{\arraystretch}{1.3} 
        \resizebox{0.9\linewidth}{!}{%
            \begin{tabular}{@{}p{0.95\linewidth}@{}}
                \toprule
                \textbf{AI Vocabulary Included in Detection Guide} \\
                \midrule
                    \textbf{Nouns}: \texttt{aspect, challenges, climate, community, component, development, dreams, environment, exploration, grand scheme, health, hidden, importance, landscape, life, manifold, multifaceted, nuance, possibilities, professional, quest, realm, revolution, roadmap, role, significance, tapestry, testament, toolkit, whimsy} \\
                    \textbf{Verbs}: \texttt{capturing, change, consider, delve/dive into, elevate, embrace, empower, enact, enhance, engage, ensure, evoking, evolving, explore, fostering, guiding, harness, highlights, improve, integrate, intricate, jeopardizing, journey, navigating, navigate, notes, offering, partaking, resonate, revolutionize, shape, seamlessly, support, tailor, transcend, underscores, understanding}\\
                    \textbf{Adjectives}: \texttt{authentic, complex, comprehensive, crafted, creative, critical, crucial, curated, deeper, diverse, elegant, essential, groundbreaking, key, meaningful, paramount, pivotal, powerful, profound, quirky, robust, seamless, significant, straightforward, structured, sustainable, transformative, valuable, vast, vibrant, vivid, whimsical}\\
                    \textbf{Adverbs}: \texttt{additionally, aptly, creatively, moreover, successfully}\\
                    \textbf{Phrases}: \texttt{as we [verb] the topic, cautionary tale, connect with, has shaped the, in a world of/where, in conclusion, in summary, it’s crucial to, it’s important to note, it’s not about \_\_\_ it’s about \_\_\_, manage topic issues/problems, not only \_\_\_ but also, packs a punch/brings a punch, paving the way, personal growth, quality of life, remember that, simple yet \_\_\_, step-by-step, such as, the effects of, the rise of, their understanding of, they identified patterns, to form the, to mitigate the risk, weaving, when it comes to topic} \\
                \bottomrule
            \end{tabular}
        }
    \endgroup
    \caption{All `AI' Vocabulary our expert annotators noted, as listed in the Detector Guide. See the full detection guide prompt in \autoref{tab:detection-guide}.} 
    \label{tab:ai_vocab}
\end{table*}

\begin{table*}[t]
\centering
\footnotesize
\begin{tabular}{p{.95\linewidth}}
\toprule
\texttt{Evader Template}\\
\midrule
\texttt{You are a journalist writing an article for {}. Your goal is to write an article given a title and subtitle, but make it look like it was written by a human. To assist you, we provide the following detection guide to determine if something is written by a human or by AI.} \\
\\
\texttt{\textbf{AI Detection Guide}} \\
\\
\texttt{You are given some examples of human and AI written articles to help you.}\\
\texttt{<start examples>}\\
\texttt{\textbf{Example Articles}}\\
\texttt{<end examples>}\\
\\
\texttt{You are also given the following guidelines to make your article seem like a human wrote it.}\\ \\
\texttt{<start of guidelines>}\\
\texttt{- Follow the detection guideline to ensure writing does not sound AI-generated.}\\
\texttt{- Do not use any of the "Overused AI Words/Phrases", "Overused AI Metaphors", "AI Grammar Patterns", AI sentence structures, AI tones, common AI names, or AI content described in the guide.}  \\
\texttt{- Include words/phrases/grammar patterns/tone/content that are consistent with human writing as described in the guide.} \\
\texttt{- Use human-written examples and explanations as references for formatting and content.} \\
\texttt{- Use the example explanations to see what humans identify as 'AI' or 'Human'}\\
\texttt{- Avoid formulaic introductions that elaborate on scene-setting or time, immediately quote experts, or bring in too much historical context.}\\
\texttt{- Avoid formatting the first setting by mentioning the setting, then a comma, then some detail or context. Avoid mentioning atmospheric details or time (of day, month, year, etc ...) in the first sentence as well.}\\
\texttt{- Avoid generic, vague, and forward-looking conclusions, and do not summarize the article in the conclusion to make articles sound more human-like. Concluding sentences don't have to wrap everything up nicely.}\\
\texttt{- Avoid using generic statements to start paragraphs, such as 'For now' or 'In the end'}\\
\texttt{- Add references to darker topics when appropriate.}\\
\texttt{- Add specific references to places, people, brands, items, facts, and metrics when appropriate.}\\
\texttt{- Always opt for specifics over broad details. For example, mention correct statistics instead of broad claims and mention company names instead of saying the type of company.}\\
\texttt{- Make sure that none of the claims made in your article are factually incorrect or implausible.}\\
\texttt{- Use language aligned with the publication you are representing as a journalist}\\
\texttt{- If your article would benefit from being organized into sections (e.g., scientific or historical content), please use section headers to do so. There are 1/3 odds an article has headers.}\\
\texttt{<end of guidelines>}\\ \\
\texttt{Now, by following the guidelines above, examples, and the information in the detection guide, please write an article without any AI signatures that also includes signatures of human writing. No reader should doubt that your article could have been published by the given source. Readers with access to the information in the detection guide should not be able to detect the article was written by AI.}\\ \\
\texttt{\textbf{Article Generation Prompt}}\\
\texttt{Article:} \\ 
\bottomrule
\end{tabular}
\caption{Prompt used for evader. The first insert into the prompt is filled by \autoref{tab:detection-guide}. The second insert is filled by examples of human and machine-generated articles.}
\label{tab:evader_prompt}
\end{table*}

\begin{table*}[t]
\centering
\scriptsize
\begin{tabular}{p{.95\linewidth}}
\toprule
\small \texttt{New Telescope Could Potentially Identify Planet X}\\
\footnotesize \texttt{Are there hidden planets in our solar system? New technologies, like the powerful Rubin Observatory, brings us closer to answers. }\\ \\
% \midrule
\texttt{For decades, astronomers have speculated that there might be another large planet lurking in the outer reaches of our solar system—an elusive “Planet X.” Often referred to as “Planet Nine” by modern researchers, this hypothetical world is believed to be so distant that it has escaped detection by existing telescopes, leaving only subtle gravitational clues in its wake. Now, a new generation of telescopes, particularly the upcoming Vera C. Rubin Observatory, promises a more powerful set of eyes on the night sky. With these tools, scientists hope to either pinpoint this mysterious planet’s location or finally put the idea to rest.} \\
\\
\texttt{\textbf{Why the Search Matters}} \\
\texttt{The concept of a hidden planet is not new. In the 19th century, the discovery of Neptune followed suspicions that Uranus’s odd orbital path hinted at an unseen gravitational pull. Today, attention has turned to the distant suburbs of our solar system, where far-off objects in the Kuiper Belt—frozen remnants of planetary formation—appear to cluster in strange ways. Some astronomers argue that only a large, unseen planet’s gravity could explain these odd orbits.}\\ \\
\texttt{“The evidence is subtle, but it’s there,” said Dr. Konstantin Batygin, a planetary astrophysicist at the California Institute of Technology (Caltech) who, along with colleague Dr. Mike Brown, first proposed the existence of a Planet Nine in 2016. “We’re seeing several distant Kuiper Belt objects all tilted and clustered in a peculiar manner, and a ninth planet several times Earth’s mass, orbiting far beyond Neptune, could be the simplest explanation.”}\\ \\
\texttt{Not everyone is convinced. Skeptics point to the small number of known distant objects and argue that the clustering could be a statistical fluke. Others think unseen observational biases might make it look like these objects are behaving strangely. Either way, the mystery remains unsolved—at least for now.}\\ \\
\texttt{\textbf{Enter the Vera C. Rubin Observatory}}\\ 
\texttt{Scheduled to begin its full operations in the near future, the Vera C. Rubin Observatory (formerly known as the Large Synoptic Survey Telescope, or LSST) in Chile represents a leap forward in astronomical capability. With a massive 8.4-meter mirror and a state-of-the-art digital camera, it will repeatedly scan the entire southern sky over the course of a decade, providing a dynamic, time-lapse-like portrait of celestial motions.}\\ \\

\texttt{“The Rubin Observatory is really a game-changer,” said Dr. Meg Schwamb, an astronomer at Queen’s University Belfast who studies the outer solar system. “Instead of looking at a small patch of sky, we’ll be looking at pretty much everything visible from Chile, over and over again. This repeated coverage means we can detect faint, distant objects that move slowly across the sky—exactly the kind of signature we’d expect from a far-flung planet.”}\\ \\

\texttt{By systematically imaging the sky every few nights, the Rubin Observatory’s Legacy Survey of Space and Time (LSST) will reveal thousands—or even tens of thousands—of new objects in the outer solar system. Among them could be Planet Nine, if it exists. Even if the telescope doesn’t directly spot the planet, it might detect more distant objects whose orbits can be mapped with unprecedented precision, allowing astronomers to figure out once and for all if a hidden giant is lurking out there.}\\ \\
\texttt{\textbf{Technological Edge}}  \\ 
\texttt{One reason Planet Nine (or any Planet X) has been so hard to pin down is that if it exists, it’s incredibly faint and slow-moving, possibly hundreds of times farther from the Sun than Earth is. Traditional telescopes rely on painstaking surveys that cover only small portions of the sky at a time. In contrast, Rubin’s wide field of view—about 40 times the size of the full Moon—means it will cover the visible sky every few days.} \\ \\
\texttt{“This isn’t just another telescope—it’s a new way of doing astronomy,” said Dr. Lynne Jones, an LSST researcher at the University of Washington. “We’re moving from static snapshots to continuous movies. If there’s something out there, no matter how faint, as long as it’s moving, we have a good chance of picking it up over time.”} \\ \\
\texttt{\textbf{Ruling In or Ruling Out}}\\
\texttt{If the Rubin Observatory does find Planet Nine, the discovery would reshape our understanding of the solar system’s architecture. A distant giant could be a leftover core of a gas giant that got tossed out in the early days of planetary formation, or it might have formed in situ, far from the Sun’s warmth. Such a find could help scientists piece together the chaotic early era when planets were jostling for position and smaller bodies were flung into distant orbits.}\\ \\
\texttt{On the other hand, if Rubin’s comprehensive survey concludes after several years without any sign of Planet Nine—or if newly discovered distant objects don’t line up as predicted by the planet hypothesis—then the case for a hidden giant will weaken. Instead, astronomers might refine models of how solar system objects distribute themselves naturally. Understanding these patterns could still teach us valuable lessons about how gravity and cosmic debris shape the outskirts of our celestial neighborhood.}\\ \\
\texttt{\textbf{Beyond Planet X}}\\
\texttt{The quest for Planet Nine is just one highlight of what the Rubin Observatory offers. It will also track near-Earth asteroids that could pose future hazards, study dark matter and dark energy by observing the distribution of galaxies, and contribute to countless other fields of astronomy.}\\ \\
\texttt{Still, for many sky-watchers, the idea of a hidden planet holds a particular allure. The notion that our solar system might still hold major surprises underscores how much we have yet to learn about our own cosmic backyard.}\\ \\
\texttt{“As much as astronomy has advanced, we’re still explorers,” said Batygin. “The Rubin Observatory gives us the tools to either find this planet or lay the mystery to rest. In the next decade, I expect we’ll know a lot more about what’s really out there—and what isn’t.”}\\ \\
\texttt{Until then, the search continues, driven by powerful new telescopes and the human desire to uncover the unseen.
}\\
\bottomrule
\end{tabular}
\caption{The one article the majority of annotators did not detect correctly, generated from \gptpro\ as part of Experiment 4 (see \S\ref{subsec:experiment4}). \href{https://www.discovermagazine.com/the-sciences/new-telescope-could-potentially-identify-planet-x}{"New Telescope Could Potentially Identify Planet X"} was originally written by Emilie Le Beau Lucchesi in Discover Magazine on Nov. 6th, 2024.}
\label{tab:incorrect_expert_article}
\end{table*}

\paragraph{Additional Existing Humanizers}
\label{app_para:existing_humanizers}
Recently, the general population is much more interested in evading automatic detectors, \footnote{Relevant discussions on Reddit: \href{https://www.reddit.com/r/WritingWithAI/comments/1b624ap/ai_humanizer_recommendations/}{"AI Humanizer Recommendations?" (2024)}, \href{https://www.reddit.com/r/ArtificialInteligence/comments/1c0jsa5/how_to_humanize_aigenerated_texts/}{"How to humanize ai-generated texts?" (2024)}, \href{https://www.reddit.com/r/ChatGPTPro/comments/1bjf18c/how_to_humanize_the_ai_generated_content/}{"How to humanize the AI generated content?" (2024)}} many commercial humanization services exist to fill this demand.Some humanization services include \href{https://undetectable.ai/detector-humanizer}{Undetectable AI}, \href{https://www.myessaywriter.ai/}{MyEssayWriter.ai} and \href{https://stealthwriter.ai/}{Stealth Writer}.

\section{Pilot Study}
\label{app_sec:pilot_study}

In this section, we provide details on the pilot study, where five annotators with varying frequencies of familiarity with LLMs detect if texts are human-written or AI-generated. The pilot study follows the same annotation process described in \S\ref{sec:data_methodology}. In the study, annotators completed two rounds of testing, one using the same articles as identified in \S\ref{subsec:experiment1}, another using stories from the subreddit \href{https://www.reddit.com/r/WritingPrompts/}{r/WritingPrompts}. 

\paragraph{Story Corpus: }
\label{app_subsec:story_generation}
Here we include more details about the stories collected for this study. We collect 30 stories from \href{https://www.reddit.com/r/WritingPrompts/}{r/WritingPrompts}. We generate corresponding AI-generated stories with the prompt in \autoref{tab:prompt_template_stories}. 

\begin{table}[t]
    \setlength{\tabcolsep}{4pt}
    \centering
    \resizebox{0.9\columnwidth}{!}{%
    \begin{tabular}{c p{9cm}}
    \toprule
         & \multicolumn{1}{c}{\bf Story Generation Prompt} \\
    \midrule
     \noalign{\vskip 1mm}
     & \texttt{Please write a story around \textbf{WORD COUNT} words responding to the following prompt: {\textbf{STORY PROMPT}}.}\\
     \noalign{\vskip 2mm} 
    \bottomrule
    \end{tabular}
    }
    \caption{Prompt Template for Story Generation, where STORY PROMPT is the writing prompt from r/WritingPrompts that the human-written story was written about and WORD COUNT is the length of the story to generate.} 
    \label{tab:prompt_template_stories}
\end{table}

\paragraph{Story Results :}
Overall, nonexperts faired better on stories than articles, with an average TPR of 69.3\% (including performance of the expert annotator). The nonexperts had a TPR of 62.5\% not including our expert annotator, as shown in \autoref{tab:story_results}. We found that many of the differences spotted were superficial due to the very informal nature of reddit, compared to our prompt which did not instruct the LLM to generate stories of that nature. In reddit, people use acronyms, for example shortening `you' to `u', type in ALL CAPS, or generally write in a very R-rated manner. While these are real differences between AI-generated and human-written texts, we wanted to pursue clues humans could use in settings where a person was genuinely attempting to pass of AI-generated writing as human-writtten, such as students writing an essay or someone publishing a news article. Future work may explore human detection abilities on fictional work, where the human-written references are more edited than the reddit thread stories.

\begin{table}[t]
\centering
\resizebox{0.9\columnwidth}{!}{%
\label{tab:annotator_performance}
\begin{tabular}{@{}lcc@{}}
\toprule
\textbf{Annotator} & \textbf{TPR (\%)} & \textbf{FPR (\%)} \\ \midrule
Nonexpert Annotator 1   & 73.3 & 76.7 \\
 Nonexpert Annotator 2  & 43.3 & 80.0 \\
Nonexpert Annotator 3   & 66.7 & 80.0 \\
Expert Annotator 1 & 96.7 & 93.3 \\
Nonexpert Annotator 4 & 66.7 & 76.7 \\ \midrule
\textbf{Average} & 69.3 & 81.3 \\ \bottomrule
\end{tabular}
}
\caption{Story performance of initial 5 annotators, which included 4 nonexpert annotators and expert annotator \# 1, who was used in all article experiments.}
\label{tab:story_results}
\end{table}

%%%%%%%%%%%%%%%%%%%%%%%%%%%%%%%%%%%%%%%%%%%%%%%%%%%%%%%%%%%%
%%%%%%%%%%%%%%%  COMMENT ANALYSIS %%%%%%%%%%%%%%%%%%%%%%%%%%%%%%%%%
%%%%%%%%%%%%%%%%%%%%%%%%%%%%%%%%%%%%%%%%%%%%%%%%%%%%%%%%%%%%
\section{Comment Analysis}
\label{app_sec:comment_analysis}

In this section, we outline the framework for analyzing expert explanations and provide additional analysis of explanations.

\subsection{Categorization of Comments}
 To categorize explanation comments, we first define the explanation categories found in \autoref{tab:explanation_category_definitions}. The first two authors individually annotated a sample of 25 expert explanations. The sample is a stratified sample of one comment per expert per each of the five experiments. Authors then came to an agreement on final human labels, refining categories as needed. We then prompt \gptnov\ with the prompt found in \autoref{tab:comment_analysis_prompt} to categorize the sample explanations. Once the prompt was able to classify the majority of the sample explanations in alignment with the authors labels, we prompt \gptnov\ to categories the explanations of all 1500 explanations. Each explanation can contain labels for multiple categories, since most explanations touch on multiple reasons of why a text is AI-generated. The total cost of classifying comments using \gpt\ was \textbf{\$6.02 USD}.

\begin{table*}[!t]
\centering
\scriptsize
\scalebox{0.88}{
\begin{tabular}{p{0.1\textwidth}p{0.06\textwidth}p{0.34\textwidth}p{0.53\textwidth}}
\toprule
\multicolumn{1}{c}{\bf\textsc{Category}} & \multicolumn{1}{c}{\bf\textsc{Freq}} & \multicolumn{1}{c}{\bf\textsc{Definition}} & \multicolumn{1}{c}{\bf\textsc{Example Explanations}} \\
\midrule
\textsc{Vocabulary} & 53.1\% & LLMs use specific words and phrases more often than human writers, which often results in repetitive, unnatural, or overly complex wording. & \vspace{-2mm} \textbf{\color{teal}{Human}}: \textit{"Furthermore, I very much doubt AI would have used adventurous adjectives like `chunky', `musky' or `thin' to describe food. Nor would it have used verbs like `blitzing' or `bolstering'."} \newline \textbf{\color{purple}{AI (\textsc{o1-Humanized})}}: \emph{"Odd word choices: wheat that `stores' a lineage; genes that are `honed.'"}\\
\midrule
\textsc{Sentence} \textsc{Structure}  & 35.9\% & AI-generated sentences follow predictable patterns (e.g., high frequency of “not only … but also …”, or consistently listing three items), while human-written sentences vary more in terms of length. & \vspace{-2mm} \textbf{\color{teal}{Human}}: \textit{"Short choppy sentences and paragraphs."} \newline \textbf{\color{purple}{AI (\textsc{o1-Pro})}}: \textit{"One pattern I've been noticing with AI, and I think I've stated this before, is the comparison of `it's not just this, it's this' and I'm seeing it here, along with listings of specifically three ideas."}\\

\midrule
\textsc{Grammar \& Punctuation}  & 24.8\% & AI-generated text is usually grammatically perfect (also avoiding dashes and ellipses), while human-written text often contains minor errors. & \vspace{-2mm} \textbf{\color{teal}{Human}}: \textit{"There's a lot of variety in the article's grammar use, with dashes, brackets, quotes intermixed with sentences, and short spurts of comma sections throughout."} \newline \textbf{\color{purple}{AI (\textsc{GPT-4o-Para})}}:\textit{"there's nothing off about the grammar or syntax in this piece..."}\\

\midrule
\textsc{Originality} & 23.7\% & AI-generated writing is generally straightforward, “safe,” and lacking in surprises or humor, leaving annotators bored or disengaged. & \vspace{-2mm} \textbf{\color{teal}{Human}}: \textit{"it's offset by some great analogies and creative phrasing that works well to convey the topic, such as with "amateur sleuths", "catnip for a certain type of Reddit user." }\newline \textbf{\color{purple}{AI (\textsc{o1-Pro})}}: \textit{"What happens when AI tries to be creative? Penguins "stand on their own flippers"."}\\

\midrule
\textsc{Quotes}  & 22.3\% & AI-generated quotes sound overly formal, lack the varied nuances of real conversation, and often mirror the article’s main text too closely in style. & \vspace{-2mm}  \textbf{\color{teal}{Human}}: \textit{" The quotes being short snippets also makes me think they're real, as the writer had to find a way to fit them into the text, rather than them just perfectly stating either side's views."} \newline \textbf{\color{purple}{AI (\textsc{GPT-4o})}}: \textit{"The quotes also feel fake, every expert speaks the same way and it's too homogenous with the text."}\\

\midrule
\textsc{Clarity} & 19.5\% & AI-generated text often lacks concise flow by over-explaining or including irrelevant details, effectively “telling” rather than “showing”. & \vspace{-2mm} \textbf{\color{teal}Human}: \textit{"Words like "meander" are used, but are used sparingly to create better flow of ideas, and its writing style is simplified in the best way possible."} \newline \textbf{\color{purple}AI (\textsc{Claude-3.5-Sonnet})}: \textit{"The sentences are condensed to provide the best possible precision with its word choice, but the article lacks flow and clarity."}\\

\midrule
\textsc{Formatting} & 15.0\% & AI-generated formatting is overly consistent (e.g., fully capitalized headings, bolded lists, paragraphs of similar length). & \vspace{-2mm} \textbf{\color{teal}{Human}}: \textit{""the formatting itself is varied, with topic headers focusing on lowercase lettering, longer explanations, and the bullet points not going into a title: one to two sentence explanation format ... " }\newline \textbf{\color{purple}{AI (\textsc{o1-Pro})}}: \textit{"... it has a structured article format of commonly used headers..."}\\
\midrule
\textsc{Conclusions} & 13.1\% & AI-generated conclusions tend to be repetitive and overly optimistic summaries, while human-written conclusions end more abruptly and less tidily. & \vspace{-2mm}  \textbf{\color{teal}{Human}}: \textit{"Both the introduction and the conclusion were humorous and unique."} \newline \textbf{\color{purple}{AI (\textsc{Claude-3.5-Sonnet})}}: \textit{Most of the conclusion leads to that summarizing, flowery tone...}\\

\midrule
\textsc{Formality} & 12.3\% & AI-generated text (especially without humanization) rarely contains filler words, contractions, slang, or abbreviations, favoring fully spelled-out terms and a polished tone. & \vspace{-2mm} \textbf{\color{teal}{Human}}: \textit{"It includes filler words like `just', `very' and `really'. Colloquial language like `sucked'."} \newline \textbf{\color{purple}{AI (\textsc{Claude-3.5-Sonnet})}}: \textit{"Phrases like `maintains that extending habeas corpus rights to animals' and `fundamentally alter the ability of accredited zoos to conduct their vital conservation work.' are so strictly formal, so tight and dense with wordage."}\\

\midrule
\textsc{Names \& Titles}  & 11.7\% & LLMs frequently generate the same names regardless of the prompt (e.g., “Emily Carter,” “Sarah Thompson”). Furthermore, if an article contains multiple people, they tend to refer to all of them with the same title (e.g., “Dr.”) without offering unique details. Unique or context-specific names (including real brands or products) are associated with human writing. & \vspace{-2mm} \textbf{\color{teal}Human}: \textit{"A couple of the experts also have quite unique names, and none of them are referred to as Dr. X (bonus points for none of them being named Emily)."} \newline \textbf{\color{purple}AI (\textsc{GPT-4o})}: \textit{"However, the introduction of Dr. Sarah Thompson and Dr. Emily Carter (who by now has more than a lifetime's worth of qualifications), means that it has to be AI text."}\\

\midrule
\textsc{Tone} & 9.3\% & The tone of AI-generated text is consistently neutral or positive, lacking depth or emotional variety compared to human writing. & \vspace{-2mm} \textbf{\color{teal}{Human}}: \textit{"The article seems to display some bias against TikTok while trying to remain impartial or impartial-seeming. The same goes for its political slant which although subtle seems to be there."} \newline \textbf{\color{purple}{AI (\textsc{o1-human})}}: \textit{"...the majority of the writing is filled with the same language it uses to describe everything - inspirational, stunning, essential, and resonating - using formal words, an inherent positivity bias, and a reflective, romantic tone that doesn't give details on why this topic matters, why we still don't know who she is, and whether or not there was any kind of controversy around the idea of women being on stage..."}\\

\midrule
\textsc{Introductions} & 7.3\% & AI-generated introductions are usually generic or focus on scenic details without providing key background, while human introductions have more compelling hooks and relevant context. & \vspace{-2mm} \textbf{\color{teal}{Human}}: \textit{"The introduction is unique as it starts with the subject of the article watching a movie rather than instantly explaining the entire point of the article."} \newline \textbf{\color{purple}{AI (\textsc{Claude-3.5-Sonnet})}}: \textit{"The article has a very generic introduction and conclusion, especially the introduction, which essentially tells you what the entire article is about."}\\

\midrule
\textsc{Factuality}  & 7.2\% & AI-generated text contains factual inconsistencies or hallucinations more often than human writing. & \vspace{-2mm}  \textbf{\color{teal}{Human}}: \textit{"The article is very factual, with details about the dates, materials used, and sales portions."} \newline \textbf{\color{purple}{AI (GPT-4o-Para)}}: \textit{"Incorrect information delivered with customary AI confidence: the `quad axel' is not another name for the backflip."} \\

\midrule
\textsc{Topics}   & 3.1\% & Unlike human authors, LLMs generally avoid darker or more mature topics (e.g., violence, graphic descriptions). & \vspace{-2mm}  \textbf{\color{teal}{Human}}: \textit{"Phrases throughout the article, including "burned to ashes and scattered at sea to prevent the crowd from venerating them as relics" and "to trample on a brass likeness of Jesus or the Virgin Mary—a blasphemous act." actively show what happened, the horrors and tragedies that occurred during that time."} \newline \textbf{\color{purple}{AI (\textsc{GPT-4o})}}: \textit{"It spends very little time talking about the horrors of the disease, and instead focuses on future research, hopeful quotes, and potential cures, even referring to it as "embarking on a new chapter"."}\\

\midrule
\textsc{Other}  & 2.6\%   & Other clues that do not fall into any of the above categories, often based on the annotator's intuition or overall impression. & \vspace{-2mm}  \textbf{\color{teal}{Human}}: \textit{"There are no clues here apart from the highlighted sentence which seems to have a human `ring' to it."} \newline \textbf{\color{purple}{AI (\textsc{o1-Humanized})}}: \textit{"The article feels slightly artificial but I can\'t really find any clear clues for it."}\\
\bottomrule
\end{tabular}}
\caption{Full taxonomy of clues used by experts to explain their detection decisions. For each category, we report the frequency of explanations that mention that category (regardless of if the annotator was correct) and provide examples of explanations for both human-written and AI-generated articles. While vocabulary and sentence structure form the most frequent clues, more complex phenomena like originality, clarity, formality, and factuality are also distinguishing features.}
\label{tab:explanation_category_definitions_full}
\end{table*}

\begin{table*}[t]
\centering
\footnotesize
\begin{tabular}{p{.95\linewidth}}
\toprule
\texttt{Comment Categorization Prompt}\\
\midrule
\texttt{We hired an annotator to determine whether an article is AI-generated or human-written. Alongside their “machine-generated” or “human-generated” label, they provided an **explanation** detailing the specific clues that led them to their decision.} \\
\\
\texttt{Your task is to **review** the annotator’s explanation and **categorize** each clue they mention. For each clue, identify: \newline
1. The **category** it falls under (e.g., Vocabulary, Grammar, etc.).\newline
2. The **label** (AI-generated or Human-written) the annotator associates with that clue.\newline
3. The exact **quote** from the annotator’s explanation that shows this clue.} \\
\\
\texttt{Below is a list of categories and **example** indicators the annotator might reference. **Note:** These examples do **not** cover all possibilities. If a comment fits two categories, choose the best one.}\\ \\
\texttt{\textbf{\{DEFINITIONS\_OF\_ALL\_CATEGORIES\}}}\\ \\
\texttt{Instructions:\newline
- Read the annotator’s explanation.\newline
- Identify any category that applies. \newline
- Indicate whether the annotator says it points to AI-generated or human-written.\newline
- Provide the exact quote from the annotator's explanation that led you to this conclusion.}\\ \\
\texttt{**If the annotator cites both AI and human clues within the same category**, please create **two separate entries** (one for AI-generated, one for human-written).  \newline
**If no category is applicable**, categorize it under “other.”}\\
\\
\texttt{**Answer Format** (use this structure for each separate category/label pair):}\\ \\
\texttt{<category>YOUR CATEGORY HERE</category>\newline
<label>YOUR LABEL HERE (AI-generated or Human-written)</label>\newline
<quote>RELEVANT QUOTE FROM EXPLANATION</quote>}\\ \\
\texttt{<category>YOUR CATEGORY HERE</category>\newline
<label>YOUR LABEL HERE (AI-generated or Human-written)</label>\newline
<quote>RELEVANT QUOTE FROM EXPLANATION</quote>}\\ \\
\texttt{(Repeat as needed, using a new block for each category/label pair.)}    \\
\texttt{Example:\newline
<category>Vocabulary</category>\newline
<label>AI-generated</label>\newline
<quote>"The article mentioned the word ‘crucial’ and used a lot of unusual synonyms."</quote>} \\ \\
\texttt{The annotators explanation is as follows:} \\ \\
\texttt{<explanation>\textbf{EXPLANATION}</explanation>}\\
\bottomrule
\end{tabular}
\caption{Truncated prompt used for comment analysis. The first insert is filled by a list of definitions of the categories found in \autoref{tab:explanation_category_definitions} and the second insert is filled by an annotator explanation.}
\label{tab:comment_analysis_prompt}
\end{table*}

\subsection{What do expert annotators see that nonexperts don't?}
\label{app_subsec:nonexpert_comments}

\begin{figure*}[t!]
\centering
\includegraphics[width=0.9\textwidth]{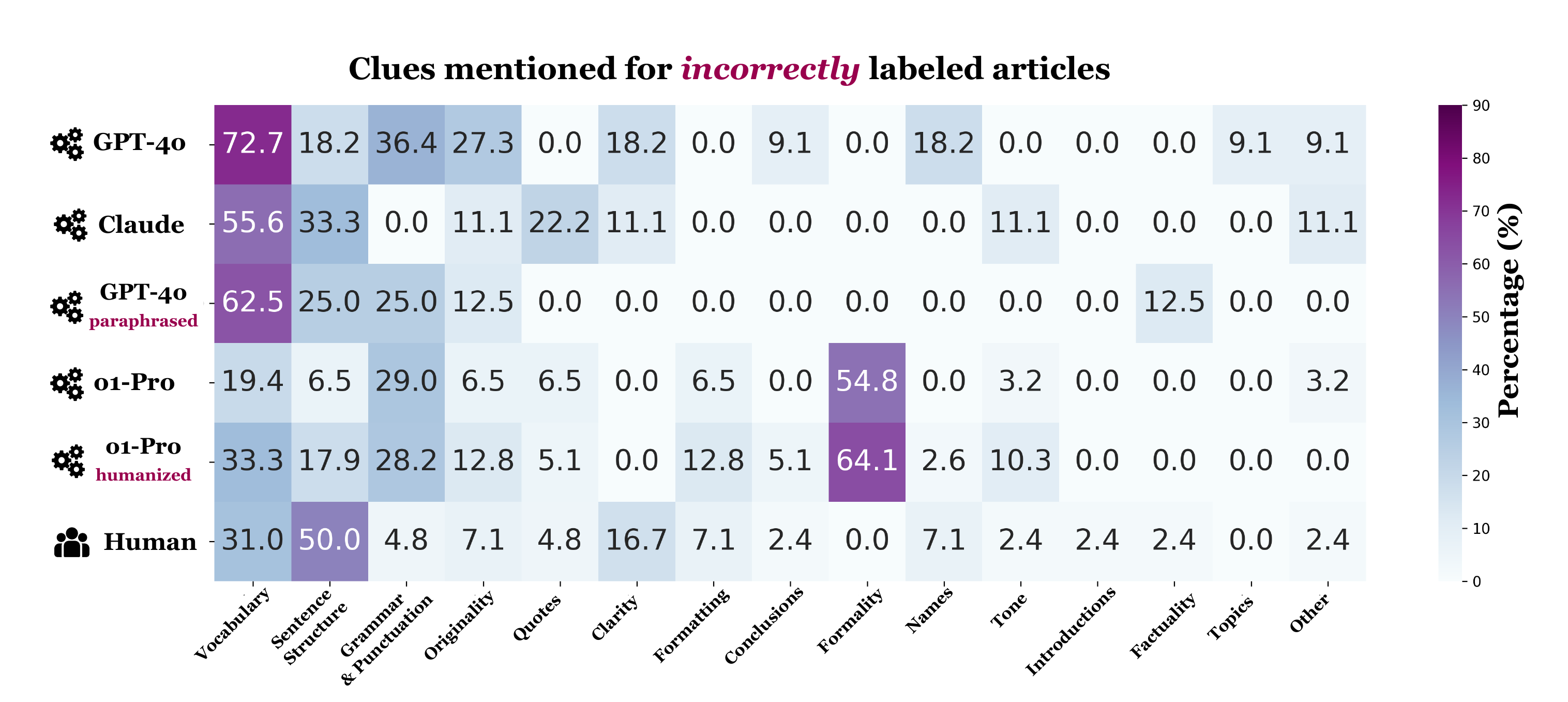}
\caption{Same as \autoref{fig:heatmaps}, except only computed over explanations when experts were incorrect. Formality is a big source of misdirection for \gptpro\ articles, while fixating on sentence structure can lead experts to false positives. 
Details of each category can be found in \autoref{tab:explanation_category_definitions}.}
\label{fig:heatmaps_incorrect}
\end{figure*}

\subsection{Individual Experts Commentary}
\label{app_subsec:individual_commentary}

Each expert had clues they favored using throughout all experiments. Annotator 1,whose category mention frequencies can be found in \autoref{fig:heatmap_ann1}, \autoref{fig:heatmap_ann2} depicts Annotator 2 comments, \autoref{fig:heatmap_ann3} shows Annotator 3 comments, \autoref{fig:heatmap_ann4} shows Annotator 4 comments and \autoref{fig:heatmap_ann5} has commentary frequencies from Annotator 5. The individual heatmaps highlight the range of clues used by annotators, who individually had no reference of clues other experts used. We again note that our experts are \emph{untrained} at this detection task, and they could likely improve their individual performance if provided with feedback on their errors.

\begin{figure*}[htbp]
\centering
\includegraphics[width=\textwidth]{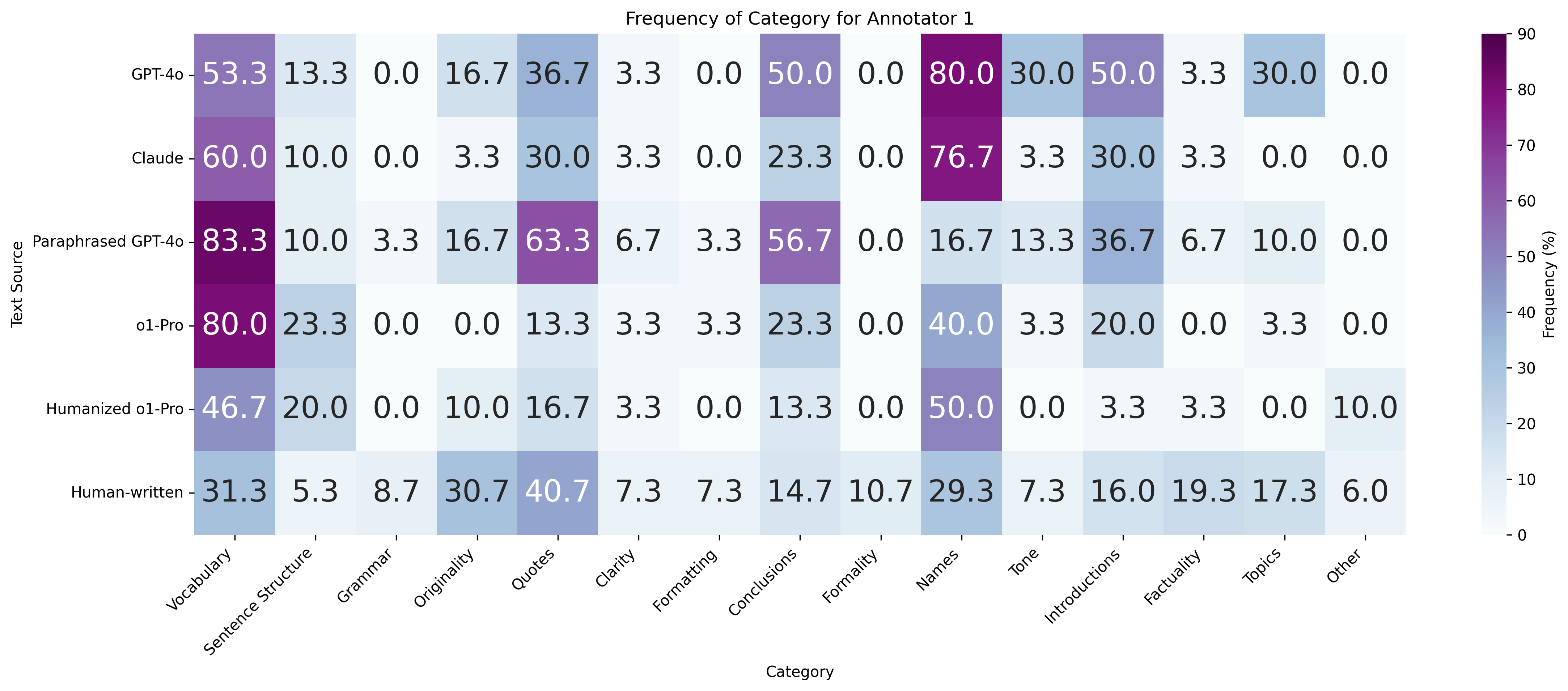}
\caption{Annotator 1 Frequency of Categories Mentioned in Explanations}
\label{fig:heatmap_ann1}
\end{figure*}

\begin{figure*}[htbp]
\centering
\includegraphics[width=\textwidth]{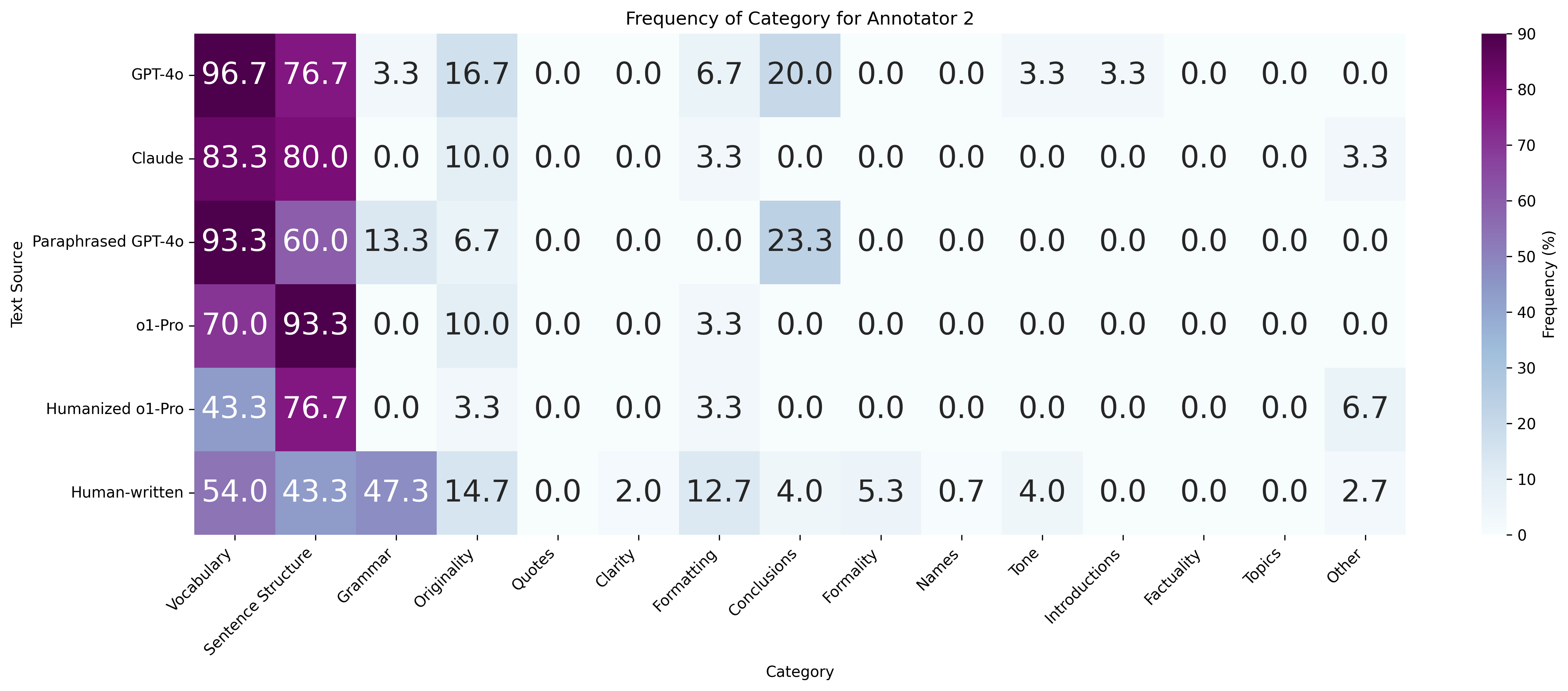}
\caption{Annotator 2 Frequency of Categories Mentioned in Explanations}
\label{fig:heatmap_ann2}
\end{figure*}

\begin{figure*}[htbp]
\centering
\includegraphics[width=\textwidth]{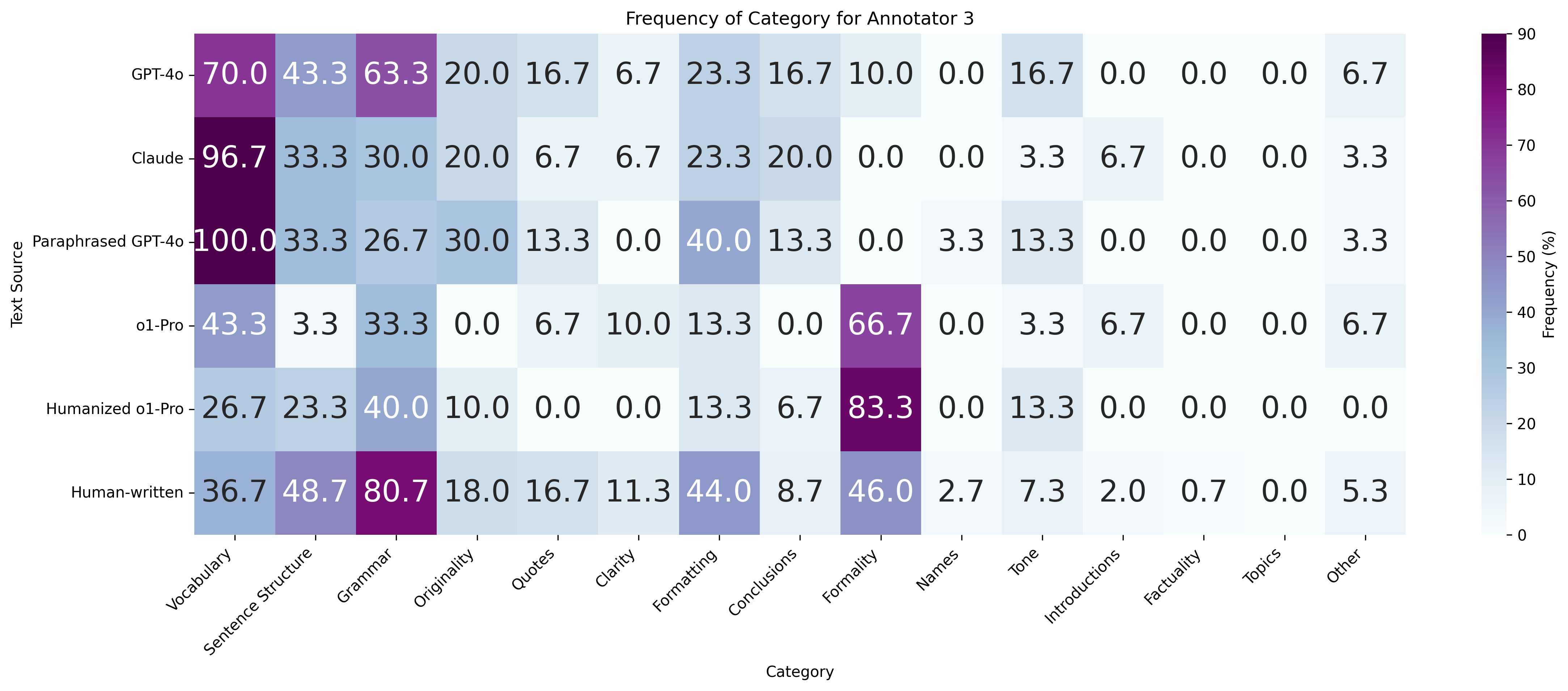}
\caption{Annotator 3 Frequency of Categories Mentioned in Explanations}
\label{fig:heatmap_ann3}
\end{figure*}

\begin{figure*}[htbp]
\centering
\includegraphics[width=\textwidth]{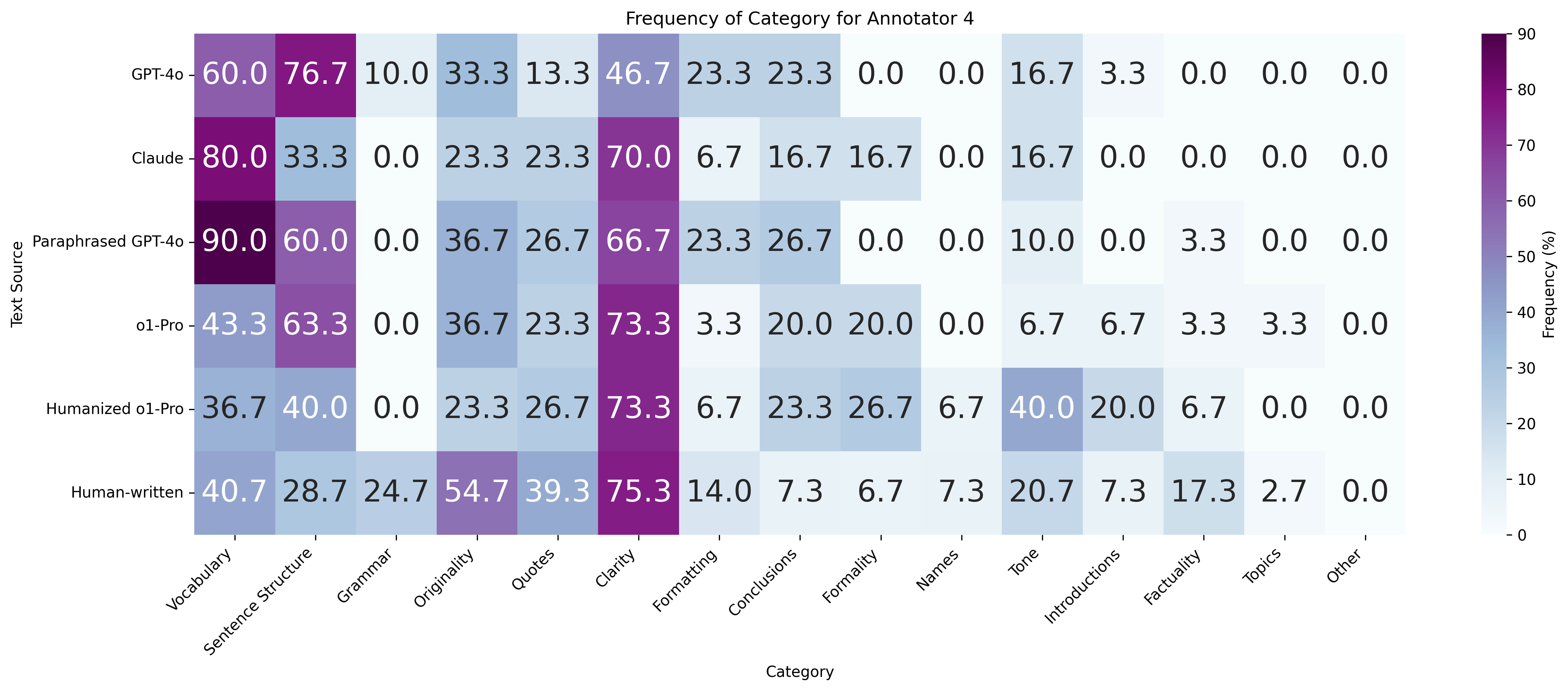}
\caption{Annotator 4 Frequency of Categories Mentioned in Explanations}
\label{fig:heatmap_ann4}
\end{figure*}

\begin{figure*}[htbp]
\centering
\includegraphics[width=\textwidth]{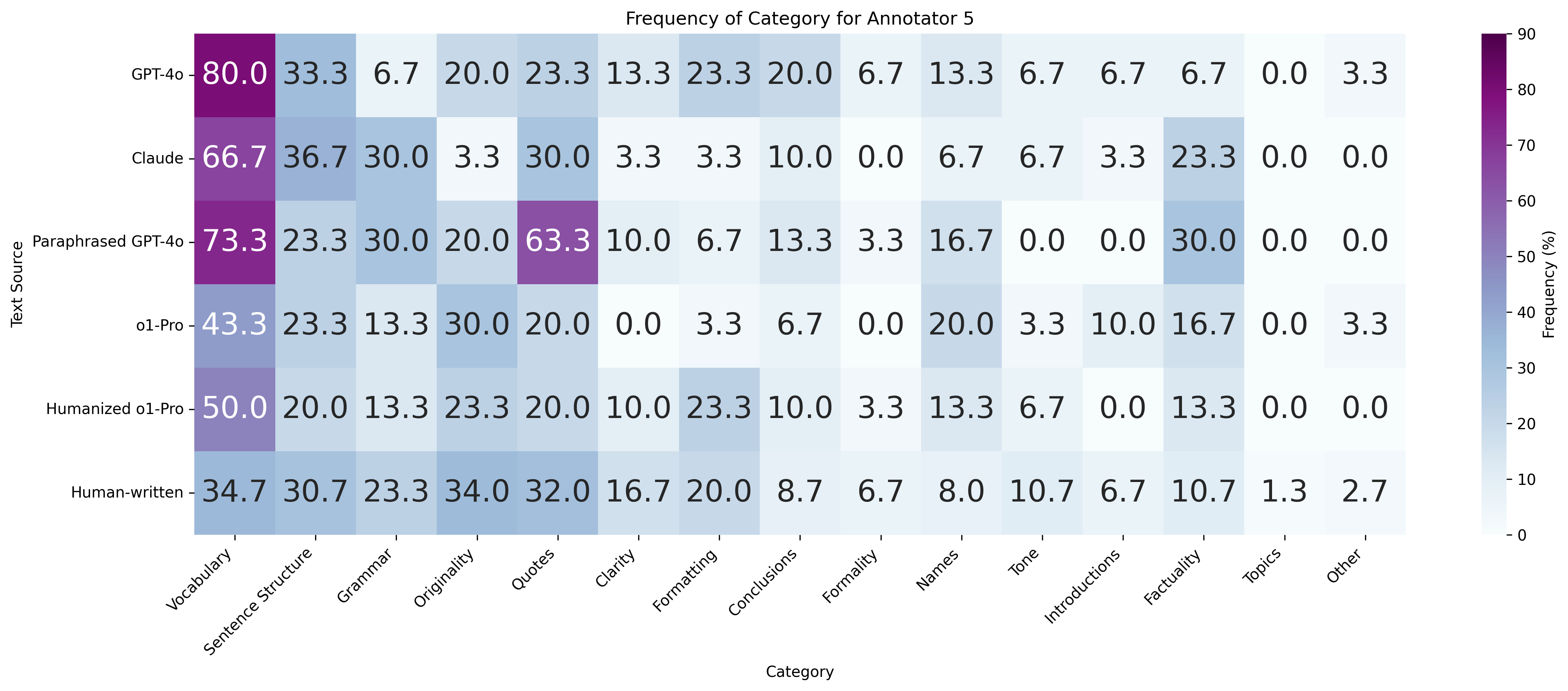}
\caption{Annotator 5 Frequency of Categories Mentioned in Explanations}
\label{fig:heatmap_ann5}
\end{figure*}

\paragraph{Annotators don't always focus on the same clues:}
Our experts focus on different properties of the text to arrive at their decisions. Annotator 1 is the only one to pick up on ``AI names'':
in fact, 63.3\% of \gpt\ and 70\% of \claude\ articles include either the name Emily or Sarah.\footnote{\gptpro\ favors names of real people instead of fictional names when generating articles.} Annotator 2 and 3 emphasize linguistic features, such as humans not always following ``proper'' writing conventions or the tendency of LLMs to always list examples in groupings of three. Annotator 4 focuses on the flow of the articles, 
analyzing the specificity of detail and motivation behind an article, while Annotator 5 frequently mentions whether a quotation sounds natural.

\paragraph{Commentary on Paraphrasing Attack in Experiment 3: }
Annotators continue to pick up on many of the same clues within the paraphrased articles that were also apparent in Experiments 1 \& 2, such as high frequency of ``AI vocab'' (even after paraphrasing), formulaic sentence structures, and cheerful summary conclusions. Somewhat counterintuitively, explanations about paraphrased articles note AI vocab in 88\% of explanations, compared to only 69.8\% of non-paraphrased GPT-4o articles. Similarly, quotations are mentioned in 33.8\% of explanations about paraphrased articles, a much higher rate than other configurations: a close reading of explanations reveals that experts flagged quotes that were always in the same format and style (e.g., only placed at the end of each paragraph). 

\paragraph{One Annotator Finds o1 Content Hard to Detect}
Interestingly, Annotator 3's TPR drops considerably, as they prioritize signs of human writing over signs of AI writing when making their judgments. Annotator 3, the only one consistently fooled by \gptpro\ outputs, relied too much on signs of informality (e.g., contractions, slang usage, usage of \emph{just} and \emph{actually}) as a sign of human writing, with 66.7\% of their explanations relating to formality (\autoref{fig:heatmap_ann3}).

% \begin{figure}[htbp]
% \centering
% \includegraphics[width=1\linewidth]{figures/conf_score_dist/confidence.pdf}
% \caption{Expert confidence in their decisions drops when judging humanized articles generated by \gptpro.}
% \vspace{-5pt}
% \label{fig:confidence_score_distribution}
% \end{figure}

%%%%%%%%%%%%%%%%%%%%%%%%%%%%%%%%%%%%%%%%%%%%%%%%%%%%%%%%%%%%
%%%%%%%%%%%%%%%  Explainable Detection & Results %%%%%%%%%%%%%%%%%%%%%%%%%%%%%%%%%
%%%%%%%%%%%%%%%%%%%%%%%%%%%%%%%%%%%%%%%%%%%%%%%%%%%%%%%%%%%%
\section{Automatic Detection}
\label{app_sec:automatic_detection}

\subsection{Automatic Detectors Benchmarked}
\label{app_subsec:benchmark_models}
In this section, we detail the detectors used as automatic benchmarks in \autoref{tab:main_results}.

\begin{itemize}
    \item \textbf{Pangram} \cite{emi_technical_2024} is a closed-source commercial detector implemented via a Transformer classifier trained via an iterative process that uses hard negative mining and synthetic data to improve data efficiency. We run both their base model and a newer model (\textsc{Pangram Humanizers}) trained to distinguish humanized data \cite{masrour_damage_2025}. We directly use the labels (i.e., thresholds) produced by the Pangram API and do not award credit for the neutral label ``Possibly AI''.
    \item \textbf{GPTZero} \cite{tian2023gptzero} is also a closed-source commercial detector that runs a classifier sentence-by-sentence across the document. We  directly use the binary labels produced by the GPTZero API. 
    \item \textbf{Binoculars} \cite{hans_spotting_2024} is an open-source detector that relies on the cross perplexity computed by two different language models to perform detection. We run Binoculars with the two FPR thresholds recommended by its authors (``Accuracy'' and ``Low FPR'' modes).
    \item \textbf{Fast-DetectGPT} \cite{bao_fast-detectgpt_2024} is an open-source method that samples and scores many perturbations of the text to estimate conditional probability curvature. We threshold this method at a FPR of 0.05, set on our held-out dev set of 40 articles. 
    \item \textbf{RADAR} \cite{hu_radar_2023} train an open-source classifier adversarially against a paraphraser. We threshold this method at a FPR of 0.05, set on our held-out dev set of 40 articles. 
\end{itemize}

\paragraph{Thresholding for Automatic Detectors: }
\label{app_subsec:thresholds}

Some automatic detectors benchmarked in \autoref{tab:main_results} and \autoref{tab:additional results} did not provide suggested thresholds for model usage. In these scenarios, we test models on a held-out test set 40 human-written articles, finding a threshold for a 5\% FPR. The respective thresholds used were 0.6051510572 for RADAR, 0.96 for Fast-DetectGPT, and 0.8963184953 for e5-lora.

\subsection{Explainable Detection}

In this section we detail the prompt based detector used for experiments in \S\ref{sec:explainable_detector}. While humans experts are found to be robust detectors, an obvious drawback is that hiring humans is expensive and slow: on average, we paid \$2.82 per article including bonuses, and gave annotators a week to complete each experiment. We wondered if we could prompt an LLM to do the same.

\paragraph{Implementation Details}

The prompt-based detector is set up to mimic how humans think about AI-generated text detection rather rather than how current automatic detectors do.  
\begin{itemize}
    \item \textbf{Zero-Shot}: To find a baseline performance of detector models, we prompt the model using the template in \autoref{tab:prompt_template_detector_zero_shot} to return if the candidate text is Human-written or AI-generated.
     \item \textbf{Zero-Shot + COT}: We prompt the model using the template in \autoref{tab:prompt_template_zero_shot_cot} to return if the candidate text is Human-written or AI-generated and an explanation of \emph{why} they text is human-written. This ablation was conducted to understand the effect explanations may have on over LLM detection performance.
     \item \textbf{Zero-Shot + Guide}: In this experiment, we prompt the model using the template in \autoref{tab:prompt_template_zero_shot_guide} to return if the candidate text is Human-written or AI-generated. This experiment serves to understand the effect including the guide in the prompt has on LLM performance.
     \item \textbf{Zero-Shot + COT + Guide}: In this experiment, we prompt the model using the template in \autoref{tab:prompt_template_explainable_detector} to return if the candidate text is Human-written or AI-generated and an explanation of \emph{why} they text is human-written. THis set up is set up to fully mimic how a human detects, thinking through clues and providing explanations of what makes them think a text is either human-written or AI-generated.
\end{itemize}

We observe that expert annotators perform better when using a majority vote strategy. Future work in prompt-based detectors could employ majority voting as a strategy to increase detection accuracy and avoid false positives. 

\begin{table}[t]
    \setlength{\tabcolsep}{4pt}
    \centering
    \resizebox{0.9\columnwidth}{!}{%
    \begin{tabular}{c p{9cm}}
    \toprule
         & \multicolumn{1}{c}{\bf Zero Shot Detector Template} \\
    \midrule
     \noalign{\vskip 1mm}
     & \texttt{You are given a candidate text and an AI detection guide. Your task is to carefully read the candidate text and determine whether it was either written by a human or generated by AI.}\\
     \noalign{\vskip 2mm}
     & \texttt{Answer HUMAN-WRITTEN if the candidate text was likely written by a human.}\\
      & \texttt{Answer AI-GENERATED if the candidate text was likely generated by an AI.}\\
      \noalign{\vskip 2mm}
     & \texttt{\textless start of candidate text\textgreater \textbf{ Candidate Text }\textless/end of candidate text\textgreater} \\
     \noalign{\vskip 2mm}
    \noalign{\vskip 2mm}
     & \texttt{\textless question\textgreater Is the above candidate text HUMAN-WRITTEN or AI-GENERATED? \textless/question\textgreater} \\
     \noalign{\vskip 2mm}
     & \texttt{Provide your final answer in this format.} \\
     \noalign{\vskip 2mm}
     & \texttt{\textless answer\textgreater\ YOUR ANSWER \textless/answer\textgreater} \\
     \noalign{\vskip 1mm}
    \bottomrule
    \end{tabular}
    }
    \caption{Prompt Template for the zero-shot detector set up } 
    \label{tab:prompt_template_detector_zero_shot}
\end{table}

\begin{table}[t]
    \setlength{\tabcolsep}{4pt}
    \centering
    \resizebox{0.9\columnwidth}{!}{%
    \begin{tabular}{c p{9cm}}
    \toprule
         & \multicolumn{1}{c}{\bf Zero-shot + CoT template} \\
    \midrule
     \noalign{\vskip 1mm}
     & \texttt{You are given a candidate text. Your task is to carefully read the candidate text and determine whether it was either written by a human or generated by AI.}\\
     \noalign{\vskip 2mm}
     & \texttt{Answer HUMAN-WRITTEN if the candidate text was likely written by a human.}\\
      & \texttt{Answer AI-GENERATED if the candidate text was likely generated by an AI.}\\
     \noalign{\vskip 2mm}
     & \texttt{\textless start of candidate text\textgreater \textbf{ Candidate Text }\textless/end of candidate text\textgreater} \\
     \noalign{\vskip 2mm}
    \noalign{\vskip 2mm}
     & \texttt{\textless question\textgreater Is the above candidate text HUMAN-WRITTEN or AI-GENERATED? \textless/question\textgreater} \\
     \noalign{\vskip 2mm}
     & \texttt{First, concisely describe the features of the candidate text that exemplify either AI or human writing. Then, provide your final answer.} \\
     \noalign{\vskip 2mm}
     & \texttt{\textless description\textgreater\ YOUR DESCRIPTION \textless/description\textgreater} \\
     \noalign{\vskip 2mm}
     & \texttt{\textless answer\textgreater\ YOUR ANSWER \textless/answer\textgreater} \\
     \noalign{\vskip 1mm}
    \bottomrule
    \end{tabular}
    }
    \caption{Prompt Template for the Zero-shot + CoT detector setup} 
    \label{tab:prompt_template_zero_shot_cot}
\end{table}

\begin{table}[t]
    \setlength{\tabcolsep}{4pt}
    \centering
    \resizebox{0.9\columnwidth}{!}{%
    \begin{tabular}{c p{9cm}}
    \toprule
         & \multicolumn{1}{c}{\bf Zero-shot + Guide Template} \\
    \midrule
     \noalign{\vskip 1mm}
     & \texttt{You are given a candidate text and an AI detection guide. Your task is to carefully read the candidate text and determine whether it was either written by a human or generated by AI.}\\
     \noalign{\vskip 2mm}
     & \texttt{Answer HUMAN-WRITTEN if the candidate text was likely written by a human based on the information in the provided guide.}\\
      & \texttt{Answer AI-GENERATED if the candidate text was likely generated by an AI based on the information in the provided guide.}\\
      \noalign{\vskip 2mm}
       & \texttt{\textbf{DETECTION GUIDE}} \\
     \noalign{\vskip 2mm}
     & \texttt{\textless start of candidate text\textgreater \textbf{ Candidate Text }\textless/end of candidate text\textgreater} \\
     \noalign{\vskip 2mm}
    \noalign{\vskip 2mm}
     & \texttt{\textless question\textgreater Based on the detection guide, is the above candidate text HUMAN-WRITTEN or AI-GENERATED? \textless/question\textgreater} \\
     \noalign{\vskip 2mm}
     & \texttt{\textless answer\textgreater\ YOUR ANSWER \textless/answer\textgreater} \\
     \noalign{\vskip 1mm}
    \bottomrule
    \end{tabular}
    }
    \caption{Prompt Template for the Zero-shot + Guide detector setup} 
    \label{tab:prompt_template_zero_shot_guide}
\end{table}

\begin{table}[t]
    \setlength{\tabcolsep}{4pt}
    \centering
    \resizebox{0.9\columnwidth}{!}{%
    \begin{tabular}{c p{9cm}}
    \toprule
         & \multicolumn{1}{c}{\bf Zero-shot + CoT + Guide Template} \\
    \midrule
     \noalign{\vskip 1mm}
     & \texttt{You are given a candidate text and an AI detection guide. Your task is to carefully read the candidate text and determine whether it was either written by a human or generated by AI.}\\
     \noalign{\vskip 2mm}
     & \texttt{Answer HUMAN-WRITTEN if the candidate text was likely written by a human based on the information in the provided guide.}\\
      & \texttt{Answer AI-GENERATED if the candidate text was likely generated by an AI based on the information in the provided guide.}\\
      \noalign{\vskip 2mm}
       & \texttt{\textbf{DETECTION GUIDE}} \\
     \noalign{\vskip 2mm}
     & \texttt{\textless start of candidate text\textgreater \textbf{ Candidate Text }\textless/end of candidate text\textgreater} \\
     \noalign{\vskip 2mm}
    \noalign{\vskip 2mm}
     & \texttt{\textless question\textgreater Based on the detection guide, is the above candidate text HUMAN-WRITTEN or AI-GENERATED? \textless/question\textgreater} \\
     \noalign{\vskip 2mm}
     & \texttt{First, use the provided guide to concisely describe the features of the candidate text that exemplify either AI or human writing. Then, provide your final answer.} \\
     \noalign{\vskip 2mm}
     & \texttt{\textless description\textgreater\ YOUR DESCRIPTION \textless/description\textgreater} \\
     \noalign{\vskip 2mm}
     & \texttt{\textless answer\textgreater\ YOUR ANSWER \textless/answer\textgreater} \\
     \noalign{\vskip 1mm}
    \bottomrule
    \end{tabular}
    }
    \caption{Prompt Template for the Zero-shot + CoT + Guide detector setup} 
    \label{tab:prompt_template_explainable_detector}
\end{table}

\subsection{Results}

\paragraph{Detector Model}

We initially test \gptaug, \gptnov, \oone, and \claude\ as potential detector models. Based on early results, we continue with ablations of \o1\ due to its low FPR and \gptnov\ due to having the highest TPR. The cost of all \oone\ detection experiments on the 300 texts is \$100.60 USD and \$25.24 USD for \gptnov\ detection experiments.  

\paragraph{Additional Results: }
While testing both \gptaug\ and \gptnov\ as detectors, we find extremely varied performance, with \gptnov\ performing much better, particularly at detecting paraphrased and \oone\ articles. \claude\ failed to detect content generated by itself and \oone\. We also tested e5-lora, one of the top performing models on the Raid \cite{dugan_raid_2024} benchmark, which we found was unable to correctly classify any AI-generated articles at a FPR of 5\%. See all additional results in \autoref{tab:additional results}.

\begin{table*}[t!]
\centering
\begin{adjustbox}{max width=\textwidth}
\begin{tabular}{lllllll}
\toprule
\multirow{2}{*}{\textsc{Detection Method}} 
& \multicolumn{5}{c}{\textsc{Generation Method}} \\ 
\cmidrule(lr){2-6} 
& \makecell{\textsc{GPT-4o}\\ \footnotesize TPR\% (\scriptsize FPR\%)}
& \makecell{\textsc{Claude}\\ \footnotesize TPR\% (\scriptsize FPR\%)}
& \makecell{\textsc{GPT-4o PARA.}\\ \footnotesize TPR\% (\scriptsize FPR\%)}
& \makecell{\textsc{o1-Pro}\\ \footnotesize TPR\% (\scriptsize FPR\%)}
& \makecell{\textsc{o1-Pro HUMAN.}\\ \footnotesize TPR\% (\scriptsize FPR\%)}
& \makecell{\textsc{OVERALL}\\ \footnotesize TPR\% (\scriptsize FPR\%)} \\
\midrule

% ==========================
% (b) Automatic detectors
% ==========================
\multicolumn{6}{l}{\textbf{(B) Automatic detectors}}\vspace{0.1cm}\\
% \midrule
 \\
 \textsc{\faUnlock\ \href{https://huggingface.co/MayZhou/e5-small-lora-ai-generated-detector}{e5-Lora} (FPR=0.05)}
 & \tprfpr{0}{0}
 & \tprfpr{0}{0}
 & \tprfpr{0}{0}
 & \tprfpr{0}{0}
 & \tprfpr{0}{0}
 & \tprfpr{0}{0}
 \\
 \midrule
\multicolumn{6}{l}{\textbf{(C) Prompt-based detectors}}\vspace{0.2cm}\\
% \midrule
\multicolumn{6}{l}{\hspace{0.5cm}Detector LLM: \textbf{\gptaug}}\vspace{0.1cm}\\

  \textsc{\faCommentDots\ Zero-Shot \gptaug }
 & \tprfpr{100}{13.3} 
 & \tprfpr{100}{3.3}
 & \tprfpr{26.7}{0}
 & \tprfpr{3.3}{0}
 & \tprfpr{0}{0} \\

 \textsc{\faCogs\ Zero-shot + CoT + Guide (\gptaug)}
 & \tprfpr{96.7}{3.3}
 & \tprfpr{100}{10}
 & \tprfpr{70}{3.3}
 & \tprfpr{46.7}{6.7}
 & \tprfpr{0}{3.3} \\

\multicolumn{6}{l}{\hspace{0.5cm}Detector LLM: \textbf{\claude\footnotesize-2024-12-17}}\vspace{0.1cm}\\

 \textsc{\faCogs\ Zero-shot + CoT + Guide}
 & \tprfpr{86.7}{0}
 & \tprfpr{43.3}{0}
  & \tprfpr{90.0}{0}
 & \tprfpr{6.7}{0}
 & \tprfpr{0}{0} 
  & \tprfpr{53.3}{0.6} 
 \\

\bottomrule
\end{tabular}
\end{adjustbox}
\caption{%
Each cell displays \textbf{TPR (FPR)}, with TPR in normal text and FPR in smaller parentheses.
Colors indicate performance bins:
\textbf{TPR} is darkest teal (\textbf{\textcolor{teal100}{100}}) at best, medium teal (\textbf{\textcolor{teal80}{90--99}}), and burnt orange (\textbf{\textcolor{burntorange}{89--70}}). Scores 69 and below are in purple (\textbf{\textcolor{purple}{$<$70}}).
\textbf{FPR} is darkest teal (\textbf{\textcolor{teal100}{0}}) at best, medium teal (\textbf{\textcolor{teal80}{1--5}}), burnt orange (\textbf{\textcolor{burntorange}{6--10}}), and purple (\textbf{\textcolor{purple}{$>$10}}) at worst.
No percentage signs appear in the cells, but the numeric values represent percentages (e.g., “90” means 90\%). We further mark closed-source (\faLock) and open-weights (\faUnlock) detectors.
}

\label{tab:additional results}
\end{table*}

%% file: acl_latex.bbl
\begin{thebibliography}{49}
\providecommand{\natexlab}[1]{#1}

\bibitem[{Allen(2017)}]{Allen2017-withinsubject}
Mike Allen. 2017.
\newblock \href {https://us.sagepub.com/en-us/nam/the-sage-encyclopedia-of-communication-research-methods/book244974} {\emph{The {SAGE} encyclopedia of communication research methods}}.
\newblock SAGE Publications, Inc, 2455 Teller Road, Thousand Oaks California 91320.

\bibitem[{Anthropic(2023)}]{anthropicclaude}
Anthropic. 2023.
\newblock \href {https://www.anthropic.com/claude} {Claude: A language model by anthropic}.
\newblock Accessed: 2025-01-20.

\bibitem[{Anthropic(2024)}]{anthropic_claude_3_addendum}
Anthropic. 2024.
\newblock Claude 3 model card addendum.
\newblock \url{https://www-cdn.anthropic.com/fed9cc193a14b84131812372d8d5857f8f304c52/Model_Card_Claude_3_Addendum.pdf}.
\newblock Accessed: 2024-12-30.

\bibitem[{Bao et~al.(2023)Bao, Zhao, Teng, Yang, and Zhang}]{bao_fast-detectgpt_2024}
Guangsheng Bao, Yanbin Zhao, Zhiyang Teng, Linyi Yang, and Yue Zhang. 2023.
\newblock Fast-detectgpt: Efficient zero-shot detection of machine-generated text via conditional probability curvature.
\newblock In \emph{The Twelfth International Conference on Learning Representations}.

\bibitem[{Brown et~al.(2020)Brown, Mann, Ryder, Subbiah, Kaplan, Dhariwal, Neelakantan, Shyam, Sastry, Askell, Agarwal, Herbert-Voss, Krueger, Henighan, Child, Ramesh, Ziegler, Wu, Winter, Hesse, Chen, Sigler, Litwin, Gray, Chess, Clark, Berner, McCandlish, Radford, Sutskever, and Amodei}]{brown_language_2020}
Tom~B. Brown, Benjamin Mann, Nick Ryder, Melanie Subbiah, Jared Kaplan, Prafulla Dhariwal, Arvind Neelakantan, Pranav Shyam, Girish Sastry, Amanda Askell, Sandhini Agarwal, Ariel Herbert-Voss, Gretchen Krueger, Tom Henighan, Rewon Child, Aditya Ramesh, Daniel~M. Ziegler, Jeffrey Wu, Clemens Winter, Christopher Hesse, Mark Chen, Eric Sigler, Mateusz Litwin, Scott Gray, Benjamin Chess, Jack Clark, Christopher Berner, Sam McCandlish, Alec Radford, Ilya Sutskever, and Dario Amodei. 2020.
\newblock \href {https://doi.org/10.48550/arXiv.2005.14165} {Language {Models} are {Few}-{Shot} {Learners}}.
\newblock \emph{arXiv preprint}.
\newblock ArXiv:2005.14165 [cs].

\bibitem[{Chakrabarty et~al.(2024)Chakrabarty, Laban, Agarwal, Muresan, and Wu}]{tuhin_creative2024}
Tuhin Chakrabarty, Philippe Laban, Divyansh Agarwal, Smaranda Muresan, and Chien-Sheng Wu. 2024.
\newblock \href {https://doi.org/10.1145/3613904.3642731} {Art or artifice? large language models and the false promise of creativity}.
\newblock In \emph{Proceedings of the 2024 CHI Conference on Human Factors in Computing Systems}, CHI '24, New York, NY, USA. Association for Computing Machinery.

\bibitem[{Chang et~al.(2024)Chang, Krishna, Houmansadr, Wieting, and Iyyer}]{chang_postmark_2024}
Yapei Chang, Kalpesh Krishna, Amir Houmansadr, John~Frederick Wieting, and Mohit Iyyer. 2024.
\newblock \href {https://doi.org/10.18653/v1/2024.emnlp-main.506} {{P}ost{M}ark: A robust blackbox watermark for large language models}.
\newblock In \emph{Proceedings of the 2024 Conference on Empirical Methods in Natural Language Processing}, pages 8969--8987, Miami, Florida, USA. Association for Computational Linguistics.

\bibitem[{Clark et~al.(2021)Clark, August, Serrano, Haduong, Gururangan, and Smith}]{clark_all_2021}
Elizabeth Clark, Tal August, Sofia Serrano, Nikita Haduong, Suchin Gururangan, and Noah~A. Smith. 2021.
\newblock \href {https://doi.org/10.18653/v1/2021.acl-long.565} {All that`s {\textquoteleft}human' is not gold: Evaluating human evaluation of generated text}.
\newblock In \emph{Proceedings of the 59th Annual Meeting of the Association for Computational Linguistics and the 11th International Joint Conference on Natural Language Processing (Volume 1: Long Papers)}, pages 7282--7296, Online. Association for Computational Linguistics.

\bibitem[{Dou et~al.(2021)Dou, Forbes, Koncel-Kedziorski, Smith, and Choi}]{dou_is_2022}
Yao Dou, Maxwell Forbes, Rik Koncel-Kedziorski, Noah~A. Smith, and Yejin Choi. 2021.
\newblock \href {https://api.semanticscholar.org/CorpusID:247315430} {Is gpt-3 text indistinguishable from human text? scarecrow: A framework for scrutinizing machine text}.
\newblock In \emph{Annual Meeting of the Association for Computational Linguistics}.

\bibitem[{Doughman et~al.(2025)Doughman, Mohammed~Afzal, Toyin, Shehata, Nakov, and Talat}]{doughman_exploring_2025}
Jad Doughman, Osama Mohammed~Afzal, Hawau~Olamide Toyin, Shady Shehata, Preslav Nakov, and Zeerak Talat. 2025.
\newblock \href {https://aclanthology.org/2025.coling-main.288/} {Exploring the {Limitations} of {Detecting} {Machine}-{Generated} {Text}}.
\newblock In \emph{Proceedings of the 31st {International} {Conference} on {Computational} {Linguistics}}, pages 4274--4281, Abu Dhabi, UAE. Association for Computational Linguistics.

\bibitem[{Dugan et~al.(2024)Dugan, Hwang, Trhl{\'\i}k, Zhu, Ludan, Xu, Ippolito, and Callison-Burch}]{dugan_raid_2024}
Liam Dugan, Alyssa Hwang, Filip Trhl{\'\i}k, Andrew Zhu, Josh~Magnus Ludan, Hainiu Xu, Daphne Ippolito, and Chris Callison-Burch. 2024.
\newblock \href {https://aclanthology.org/2024.acl-long.674} {{RAID}: A shared benchmark for robust evaluation of machine-generated text detectors}.
\newblock In \emph{Proceedings of the 62nd Annual Meeting of the Association for Computational Linguistics (Volume 1: Long Papers)}, pages 12463--12492, Bangkok, Thailand. Association for Computational Linguistics.

\bibitem[{Dugan et~al.(2020)Dugan, Ippolito, Kirubarajan, and Callison-Burch}]{dugan_roft_2020}
Liam Dugan, Daphne Ippolito, Arun Kirubarajan, and Chris Callison-Burch. 2020.
\newblock \href {https://doi.org/10.18653/v1/2020.emnlp-demos.25} {{RoFT}: {A} {Tool} for {Evaluating} {Human} {Detection} of {Machine}-{Generated} {Text}}.
\newblock In \emph{Proceedings of the 2020 {Conference} on {Empirical} {Methods} in {Natural} {Language} {Processing}: {System} {Demonstrations}}, pages 189--196, Online. Association for Computational Linguistics.

\bibitem[{Dugan et~al.(2022)Dugan, Ippolito, Kirubarajan, Shi, and Callison-Burch}]{dugan_real_2023}
Liam Dugan, Daphne Ippolito, Arun Kirubarajan, Sherry Shi, and Chris Callison-Burch. 2022.
\newblock \href {https://api.semanticscholar.org/CorpusID:255125274} {Real or fake text?: Investigating human ability to detect boundaries between human-written and machine-generated text}.
\newblock In \emph{AAAI Conference on Artificial Intelligence}.

\bibitem[{Emi and Spero(2024)}]{emi_technical_2024}
Bradley Emi and Max Spero. 2024.
\newblock \href {https://doi.org/10.48550/arXiv.2402.14873} {Technical {Report} on the {Pangram} {AI}-{Generated} {Text} {Classifier}}.
\newblock \emph{arXiv preprint}.
\newblock ArXiv:2402.14873 [cs].

\bibitem[{Gameiro et~al.(2024)Gameiro, Kucharavy, and Dolamic}]{gameiro_llm_2024}
Henrique Da~Silva Gameiro, Andrei Kucharavy, and Ljiljana Dolamic. 2024.
\newblock \href {https://doi.org/10.48550/arXiv.2409.03291} {{LLM} {Detectors} {Still} {Fall} {Short} of {Real} {World}: {Case} of {LLM}-{Generated} {Short} {News}-{Like} {Posts}}.
\newblock \emph{arXiv preprint}.
\newblock ArXiv:2409.03291 [cs].

\bibitem[{Gardner et~al.(2020)Gardner, Artzi, Basmov, Berant, Bogin, Chen, Dasigi, Dua, Elazar, Gottumukkala, Gupta, Hajishirzi, Ilharco, Khashabi, Lin, Liu, Liu, Mulcaire, Ning, Singh, Smith, Subramanian, Tsarfaty, Wallace, Zhang, and Zhou}]{gardner-etal-2020-evaluating}
Matt Gardner, Yoav Artzi, Victoria Basmov, Jonathan Berant, Ben Bogin, Sihao Chen, Pradeep Dasigi, Dheeru Dua, Yanai Elazar, Ananth Gottumukkala, Nitish Gupta, Hannaneh Hajishirzi, Gabriel Ilharco, Daniel Khashabi, Kevin Lin, Jiangming Liu, Nelson~F. Liu, Phoebe Mulcaire, Qiang Ning, Sameer Singh, Noah~A. Smith, Sanjay Subramanian, Reut Tsarfaty, Eric Wallace, Ally Zhang, and Ben Zhou. 2020.
\newblock \href {https://doi.org/10.18653/v1/2020.findings-emnlp.117} {Evaluating models{'} local decision boundaries via contrast sets}.
\newblock In \emph{Findings of the Association for Computational Linguistics: EMNLP 2020}, pages 1307--1323, Online. Association for Computational Linguistics.

\bibitem[{Gehrmann et~al.(2019)Gehrmann, Strobelt, and Rush}]{gehrmann_gltr_2019}
Sebastian Gehrmann, Hendrik Strobelt, and Alexander Rush. 2019.
\newblock \href {https://doi.org/10.18653/v1/P19-3019} {{GLTR}: {Statistical} {Detection} and {Visualization} of {Generated} {Text}}.
\newblock In \emph{Proceedings of the 57th {Annual} {Meeting} of the {Association} for {Computational} {Linguistics}: {System} {Demonstrations}}, pages 111--116, Florence, Italy. Association for Computational Linguistics.

\bibitem[{Hans et~al.(2024)Hans, Schwarzschild, Cherepanova, Kazemi, Saha, Goldblum, Geiping, and Goldstein}]{hans_spotting_2024}
Abhimanyu Hans, Avi Schwarzschild, Valeriia Cherepanova, Hamid Kazemi, Aniruddha Saha, Micah Goldblum, Jonas Geiping, and Tom Goldstein. 2024.
\newblock \href {https://proceedings.mlr.press/v235/hans24a.html} {Spotting {LLM}s with binoculars: Zero-shot detection of machine-generated text}.
\newblock In \emph{Proceedings of the 41st International Conference on Machine Learning}, volume 235 of \emph{Proceedings of Machine Learning Research}, pages 17519--17537. PMLR.

\bibitem[{Hu et~al.(2023)Hu, Chen, and Ho}]{hu_radar_2023}
Xiaomeng Hu, Pin{-}Yu Chen, and Tsung{-}Yi Ho. 2023.
\newblock {RADAR:} robust ai-text detection via adversarial learning.
\newblock In \emph{Advances in Neural Information Processing Systems 36: Annual Conference on Neural Information Processing Systems 2023, NeurIPS 2023, New Orleans, LA, USA, December 10 - 16, 2023}.

\bibitem[{Ippolito et~al.(2020)Ippolito, Duckworth, Callison-Burch, and Eck}]{ippolito_automatic_2020}
Daphne Ippolito, Daniel Duckworth, Chris Callison-Burch, and Douglas Eck. 2020.
\newblock \href {https://doi.org/10.18653/v1/2020.acl-main.164} {Automatic detection of generated text is easiest when humans are fooled}.
\newblock In \emph{Proceedings of the 58th Annual Meeting of the Association for Computational Linguistics}, pages 1808--1822, Online. Association for Computational Linguistics.

\bibitem[{Ji et~al.(2024)Ji, Li, Li, Guo, Qiu, Huang, Chen, Jiang, and Lu}]{ji_detecting_2024}
Jiazhou Ji, Ruizhe Li, Shujun Li, Jie Guo, Weidong Qiu, Zheng Huang, Chiyu Chen, Xiaoyu Jiang, and Xinru Lu. 2024.
\newblock \href {http://arxiv.org/abs/2406.18259} {Detecting {Machine}-{Generated} {Texts}: {Not} {Just} "{AI} vs {Humans}" and {Explainability} is {Complicated}}.
\newblock \emph{arXiv preprint}.
\newblock ArXiv:2406.18259 [cs].

\bibitem[{Karpinska et~al.(2021)Karpinska, Akoury, and Iyyer}]{karpinska-etal-2021-perils}
Marzena Karpinska, Nader Akoury, and Mohit Iyyer. 2021.
\newblock \href {https://doi.org/10.18653/v1/2021.emnlp-main.97} {The perils of using {M}echanical {T}urk to evaluate open-ended text generation}.
\newblock In \emph{Proceedings of the 2021 Conference on Empirical Methods in Natural Language Processing}, pages 1265--1285, Online and Punta Cana, Dominican Republic. Association for Computational Linguistics.

\bibitem[{Karpinska et~al.(2022)Karpinska, Raj, Thai, Song, Gupta, and Iyyer}]{karpinska-etal-2022-demetr}
Marzena Karpinska, Nishant Raj, Katherine Thai, Yixiao Song, Ankita Gupta, and Mohit Iyyer. 2022.
\newblock \href {https://doi.org/10.18653/v1/2022.emnlp-main.649} {{DEMETR}: Diagnosing evaluation metrics for translation}.
\newblock In \emph{Proceedings of the 2022 Conference on Empirical Methods in Natural Language Processing}, pages 9540--9561, Abu Dhabi, United Arab Emirates. Association for Computational Linguistics.

\bibitem[{Karpinska et~al.(2024)Karpinska, Thai, Lo, Goyal, and Iyyer}]{karpinska-etal-2024-one}
Marzena Karpinska, Katherine Thai, Kyle Lo, Tanya Goyal, and Mohit Iyyer. 2024.
\newblock \href {https://doi.org/10.18653/v1/2024.emnlp-main.948} {One thousand and one pairs: A {\textquotedblleft}novel{\textquotedblright} challenge for long-context language models}.
\newblock In \emph{Proceedings of the 2024 Conference on Empirical Methods in Natural Language Processing}, pages 17048--17085, Miami, Florida, USA. Association for Computational Linguistics.

\bibitem[{Kirchenbauer et~al.(2023)Kirchenbauer, Geiping, Wen, Katz, Miers, and Goldstein}]{pmlr-v202-kirchenbauer23a}
John Kirchenbauer, Jonas Geiping, Yuxin Wen, Jonathan Katz, Ian Miers, and Tom Goldstein. 2023.
\newblock \href {https://proceedings.mlr.press/v202/kirchenbauer23a.html} {A watermark for large language models}.
\newblock In \emph{Proceedings of the 40th International Conference on Machine Learning}, volume 202 of \emph{Proceedings of Machine Learning Research}, pages 17061--17084. PMLR.

\bibitem[{Krishna et~al.(2023)Krishna, Song, Karpinska, Wieting, and Iyyer}]{krishna_paraphrasing_2023}
Kalpesh Krishna, Yixiao Song, Marzena Karpinska, John Wieting, and Mohit Iyyer. 2023.
\newblock \href {https://proceedings.neurips.cc/paper_files/paper/2023/file/575c450013d0e99e4b0ecf82bd1afaa4-Paper-Conference.pdf} {Paraphrasing evades detectors of ai-generated text, but retrieval is an effective defense}.
\newblock In \emph{Advances in Neural Information Processing Systems}, volume~36, pages 27469--27500. Curran Associates, Inc.

\bibitem[{Li et~al.(2024)Li, Li, Cui, Bi, Wang, Wang, Yang, Shi, and Zhang}]{li_mage_2024}
Yafu Li, Qintong Li, Leyang Cui, Wei Bi, Zhilin Wang, Longyue Wang, Linyi Yang, Shuming Shi, and Yue Zhang. 2024.
\newblock \href {https://doi.org/10.18653/v1/2024.acl-long.3} {{MAGE}: {Machine}-generated {Text} {Detection} in the {Wild}}.
\newblock In \emph{Proceedings of the 62nd {Annual} {Meeting} of the {Association} for {Computational} {Linguistics} ({Volume} 1: {Long} {Papers})}, pages 36--53, Bangkok, Thailand. Association for Computational Linguistics.

\bibitem[{Lipton(2018)}]{lipton2018mythos}
Zachary~C. Lipton. 2018.
\newblock \href {https://doi.org/10.1145/3236386.3241340} {The mythos of model interpretability: In machine learning, the concept of interpretability is both important and slippery.}
\newblock \emph{Queue}, 16(3):31–57.

\bibitem[{Lu et~al.(2023)Lu, Liu, He, and Tang}]{lu_large_2024}
Ning Lu, Shengcai Liu, Ruidan He, and Ke~Tang. 2023.
\newblock \href {https://api.semanticscholar.org/CorpusID:258762215} {Large language models can be guided to evade ai-generated text detection}.
\newblock \emph{Trans. Mach. Learn. Res.}, 2024.

\bibitem[{Ma et~al.(2023)Ma, Liu, Yi, Cheng, Huang, Lu, and Liu}]{ma_ai_2023}
Yongqiang Ma, Jiawei Liu, Fan Yi, Qikai Cheng, Yong Huang, Wei Lu, and Xiaozhong Liu. 2023.
\newblock \href {https://arxiv.org/abs/2301.10416} {Ai vs. human -- differentiation analysis of scientific content generation}.
\newblock \emph{Preprint}, arXiv:2301.10416.

\bibitem[{Masrour et~al.(2025)Masrour, Emi, and Spero}]{masrour_damage_2025}
Elyas Masrour, Bradley Emi, and Max Spero. 2025.
\newblock \href {https://doi.org/10.48550/arXiv.2501.03437} {{DAMAGE}: {Detecting} {Adversarially} {Modified} {AI} {Generated} {Text}}.
\newblock \emph{arXiv preprint}.
\newblock ArXiv:2501.03437 [cs].

\bibitem[{Mitchell et~al.(2023)Mitchell, Lee, Khazatsky, Manning, and Finn}]{mitchell_detectgpt_2023}
Eric Mitchell, Yoonho Lee, Alexander Khazatsky, Christopher~D. Manning, and Chelsea Finn. 2023.
\newblock \href {https://api.semanticscholar.org/CorpusID:256274849} {Detectgpt: Zero-shot machine-generated text detection using probability curvature}.
\newblock In \emph{International Conference on Machine Learning}.

\bibitem[{OpenAI(2024)}]{openai_openai_2024}
OpenAI. 2024.
\newblock \href {https://doi.org/10.48550/arXiv.2412.16720} {{OpenAI} o1 {System} {Card}}.
\newblock \emph{arXiv preprint}.
\newblock ArXiv:2412.16720 [cs].

\bibitem[{Porter and Machery(2024)}]{porter_ai-generated_2024}
B~Porter and Edouard Machery. 2024.
\newblock \href {https://api.semanticscholar.org/CorpusID:274057422} {Ai-generated poetry is indistinguishable from human-written poetry and is rated more favorably}.
\newblock \emph{Scientific Reports}, 14.

\bibitem[{Puccetti et~al.(2024)Puccetti, Rogers, Alzetta, Dell'Orletta, and Esuli}]{puccetti_ai_2024}
Giovanni Puccetti, Anna Rogers, Chiara Alzetta, Felice Dell'Orletta, and Andrea Esuli. 2024.
\newblock \href {https://doi.org/10.18653/v1/2024.acl-long.817} {{AI} "{News}" {Content} {Farms} {Are} {Easy} to {Make} and {Hard} to {Detect}: {A} {Case} {Study} in {Italian}}.
\newblock In \emph{Proceedings of the 62nd {Annual} {Meeting} of the {Association} for {Computational} {Linguistics} ({Volume} 1: {Long} {Papers})}, pages 15312--15338.
\newblock ArXiv:2406.12128 [cs].

\bibitem[{Sadasivan et~al.(2024)Sadasivan, Kumar, Balasubramanian, Wang, and Feizi}]{sadasivan_can_2024}
Vinu~Sankar Sadasivan, Aounon Kumar, Sriram Balasubramanian, Wenxiao Wang, and Soheil Feizi. 2024.
\newblock \href {https://doi.org/10.48550/arXiv.2303.11156} {Can {AI}-{Generated} {Text} be {Reliably} {Detected}?}
\newblock \emph{arXiv preprint}.
\newblock ArXiv:2303.11156 [cs].

\bibitem[{Shaib et~al.(2024)Shaib, Elazar, Li, and Wallace}]{shaib_detection_2024}
Chantal Shaib, Yanai Elazar, Junyi~Jessy Li, and Byron~C. Wallace. 2024.
\newblock \href {https://api.semanticscholar.org/CorpusID:270869797} {Detection and measurement of syntactic templates in generated text}.
\newblock In \emph{Conference on Empirical Methods in Natural Language Processing}.

\bibitem[{Shi et~al.(2023)Shi, Wang, Yin, Chen, Chang, and Hsieh}]{shi_red_2024}
Zhouxing Shi, Yihan Wang, Fan Yin, Xiangning Chen, Kai-Wei Chang, and Cho-Jui Hsieh. 2023.
\newblock \href {https://api.semanticscholar.org/CorpusID:258987266} {Red teaming language model detectors with language models}.
\newblock \emph{Transactions of the Association for Computational Linguistics}, 12:174--189.

\bibitem[{Solaiman et~al.(2019)Solaiman, Brundage, Clark, Askell, Herbert-Voss, Wu, Radford, and Wang}]{solaiman_release_2019}
Irene Solaiman, Miles Brundage, Jack Clark, Amanda Askell, Ariel Herbert-Voss, Jeff Wu, Alec Radford, and Jasmine Wang. 2019.
\newblock \href {https://www.semanticscholar.org/paper/Release-Strategies-and-the-Social-Impacts-of-Models-Solaiman-Brundage/c7462e0ee928f095a7fc40b91f1e7557d283ae8e} {Release {Strategies} and the {Social} {Impacts} of {Language} {Models}}.
\newblock \emph{ArXiv}.

\bibitem[{Tian and Cui(2023)}]{tian2023gptzero}
Edward Tian and Alexander Cui. 2023.
\newblock \href {https://gptzero.me} {Gptzero: Towards detection of ai-generated text using zero-shot and supervised methods"}.

\bibitem[{Verma et~al.(2023)Verma, Fleisig, Tomlin, and Klein}]{verma_ghostbuster_2024}
Vivek~Kumar Verma, Eve Fleisig, Nicholas Tomlin, and Dan Klein. 2023.
\newblock \href {https://api.semanticscholar.org/CorpusID:258865787} {Ghostbuster: Detecting text ghostwritten by large language models}.
\newblock In \emph{North American Chapter of the Association for Computational Linguistics}.

\bibitem[{Wang et~al.(2024{\natexlab{a}})Wang, Li, Yang, and Mao}]{wang_raft_2024}
James Wang, Ran Li, Junfeng Yang, and Chengzhi Mao. 2024{\natexlab{a}}.
\newblock \href {https://doi.org/10.48550/arXiv.2410.03658} {{RAFT}: {Realistic} {Attacks} to {Fool} {Text} {Detectors}}.
\newblock \emph{arXiv preprint}.
\newblock ArXiv:2410.03658 [cs].

\bibitem[{Wang et~al.(2024{\natexlab{b}})Wang, Chen, Liu, Chen, Chen, Zhang, and Cheng}]{wang_humanizing_2024}
Tianchun Wang, Yuanzhou Chen, Zichuan Liu, Zhanwen Chen, Haifeng Chen, Xiang Zhang, and Wei Cheng. 2024{\natexlab{b}}.
\newblock \href {https://doi.org/10.48550/ARXIV.2410.19230} {Humanizing the {Machine}: {Proxy} {Attacks} to {Mislead} {LLM} {Detectors}}.
\newblock \emph{ArXiv}.
\newblock Publisher: arXiv Version Number: 1.

\bibitem[{Wang et~al.(2025)Wang, Xing, Mansurov, Puccetti, Xie, Ta, Geng, Su, Abassy, Ahmed, Elozeiri, Laiyk, Goloburda, Mahmoud, Tomar, Aziz, Koike, Kaneko, Shelmanov, Artemova, Mikhailov, Tsvigun, Aji, Habash, Gurevych, and Nakov}]{wang2025humanliketextlikedhumans}
Yuxia Wang, Rui Xing, Jonibek Mansurov, Giovanni Puccetti, Zhuohan Xie, Minh~Ngoc Ta, Jiahui Geng, Jinyan Su, Mervat Abassy, Saad El~Dine Ahmed, Kareem Elozeiri, Nurkhan Laiyk, Maiya Goloburda, Tarek Mahmoud, Raj~Vardhan Tomar, Alexander Aziz, Ryuto Koike, Masahiro Kaneko, Artem Shelmanov, Ekaterina Artemova, Vladislav Mikhailov, Akim Tsvigun, Alham~Fikri Aji, Nizar Habash, Iryna Gurevych, and Preslav Nakov. 2025.
\newblock \href {https://arxiv.org/abs/2502.11614} {Is human-like text liked by humans? multilingual human detection and preference against ai}.
\newblock \emph{Preprint}, arXiv:2502.11614.

\bibitem[{Warstadt et~al.(2020)Warstadt, Parrish, Liu, Mohananey, Peng, Wang, and Bowman}]{warstadt-etal-2020-blimp-benchmark}
Alex Warstadt, Alicia Parrish, Haokun Liu, Anhad Mohananey, Wei Peng, Sheng-Fu Wang, and Samuel~R. Bowman. 2020.
\newblock \href {https://doi.org/10.1162/tacl_a_00321} {{BL}i{MP}: The benchmark of linguistic minimal pairs for {E}nglish}.
\newblock \emph{Transactions of the Association for Computational Linguistics}, 8:377--392.

\bibitem[{Zhang et~al.(2024)Zhang, Gao, Chen, Huang, Huang, Sun, Zhang, Li, Fu, Wan, and Sun}]{zhang_llm-as--coauthor_2024}
Qihui Zhang, Chujie Gao, Dongping Chen, Yue Huang, Yixin Huang, Zhenyang Sun, Shilin Zhang, Weiye Li, Zhengyan Fu, Yao Wan, and Lichao Sun. 2024.
\newblock \href {https://api.semanticscholar.org/CorpusID:266933659} {Llm-as-a-coauthor: Can mixed human-written and machine-generated text be detected?}
\newblock In \emph{NAACL-HLT}.

\bibitem[{Zhong et~al.(2024)Zhong, Liu, Pan, Zhang, Zhou, Liang, Wu, Lyu, Shu, Yu, Cao, Jiang, Chen, Li, Chen, Hu, Liu, Zhao, Xu, Dai, Zhao, Zhang, Zhao, Yang, Chen, Wang, Ruan, Wang, Zhao, Zhang, Ren, Qin, Chen, Li, Zidan, Jahin, Chen, Xia, Holmes, Zhuang, Wang, Xu, Xia, Yu, Tang, Yang, Sun, Yang, Lu, Wang, Chai, Li, Lu, Sun, Zhang, Ge, Hu, Zhang, Zhou, Zhang, Zhang, Liu, Jiang, Kong, Xiang, Ren, Liu, Jiang, Bao, Zhang, Li, Li, Liu, Shen, Sikora, Zhai, Zhu, and Liu}]{zhong_evaluation_2024}
Tianyang Zhong, Zhengliang Liu, Yi~Pan, Yutong Zhang, Yifan Zhou, Shizhe Liang, Zihao Wu, Yanjun Lyu, Peng Shu, Xiaowei Yu, Chao Cao, Hanqi Jiang, Hanxu Chen, Yiwei Li, Junhao Chen, Huawen Hu, Yihen Liu, Huaqin Zhao, Shaochen Xu, Haixing Dai, Lin Zhao, Ruidong Zhang, Wei Zhao, Zhenyuan Yang, Jingyuan Chen, Peilong Wang, Wei Ruan, Hui Wang, Huan Zhao, Jing Zhang, Yiming Ren, Shihuan Qin, Tong Chen, Jiaxi Li, Arif~Hassan Zidan, Afrar Jahin, Minheng Chen, Sichen Xia, Jason Holmes, Yan Zhuang, Jiaqi Wang, Bochen Xu, Weiran Xia, Jichao Yu, Kaibo Tang, Yaxuan Yang, Bolun Sun, Tao Yang, Guoyu Lu, Xianqiao Wang, Lilong Chai, He~Li, Jin Lu, Lichao Sun, Xin Zhang, Bao Ge, Xintao Hu, Lian Zhang, Hua Zhou, Lu~Zhang, Shu Zhang, Ninghao Liu, Bei Jiang, Linglong Kong, Zhen Xiang, Yudan Ren, Jun Liu, Xi~Jiang, Yu~Bao, Wei Zhang, Xiang Li, Gang Li, Wei Liu, Dinggang Shen, Andrea Sikora, Xiaoming Zhai, Dajiang Zhu, and Tianming Liu. 2024.
\newblock \href {https://doi.org/10.48550/arXiv.2409.18486} {Evaluation of {OpenAI} o1: {Opportunities} and {Challenges} of {AGI}}.
\newblock \emph{arXiv preprint}.
\newblock ArXiv:2409.18486 [cs].

\bibitem[{Zhou et~al.(2024)Zhou, He, and Sun}]{zhou_humanizing_2024}
Ying Zhou, Ben He, and Le~Sun. 2024.
\newblock \href {https://api.semanticscholar.org/CorpusID:268856865} {Humanizing machine-generated content: Evading ai-text detection through adversarial attack}.
\newblock In \emph{International Conference on Language Resources and Evaluation}.

\bibitem[{Zhu et~al.(2024)Zhu, Zhang, and Wang}]{zhu_embracing_2024}
Tiffany Zhu, Kexun Zhang, and William~Yang Wang. 2024.
\newblock \href {https://doi.org/10.48550/arXiv.2411.18708} {Embracing {AI} in {Education}: {Understanding} the {Surge} in {Large} {Language} {Model} {Use} by {Secondary} {Students}}.
\newblock \emph{arXiv preprint}.
\newblock ArXiv:2411.18708 [cs] version: 1.

\end{thebibliography}
